%% file: style.tex
\documentclass[11pt,oneside]{book}

\usepackage[english]{babel}
\usepackage[utf8]{inputenc}
\usepackage[T1]{fontenc}

\usepackage[a4paper]{geometry}
\usepackage{mathptmx}
\usepackage{courier}
\usepackage{anyfontsize}

\usepackage{amsmath}
\usepackage{amssymb}
\usepackage{amsfonts}

\usepackage[table,xcdraw]{xcolor}
\definecolor{gray97}{gray}{.97}
\definecolor{gray75}{gray}{.75}
\definecolor{gray45}{gray}{.45}

\usepackage{graphicx}
\graphicspath{ {images/} }
\usepackage{float}
\usepackage{subcaption}
\usepackage{tikz}
\usetikzlibrary{shapes.geometric, arrows}
\usepackage{pgfgantt}

\usepackage{longtable}
\usepackage{booktabs}
\usepackage{multirow}
\usepackage{lscape}

\usepackage{enumerate}
\usepackage{pifont}
\usepackage{lettrine}
\usepackage{setspace}
\usepackage{verbatim}
\usepackage{soul}      
\usepackage{placeins}  
\usepackage{siunitx}
\usepackage[toc,page]{appendix}

\usepackage{acro}
\DeclareAcronym{ML}{
  short = ML,
  long  = Machine Learning,
  sort  = ML,
}
\DeclareAcronym{MU}{
  short = MU,
  long  = Machine Unlearning,
  sort  = MU,
}
\DeclareAcronym{AI}{
  short = AI,
  long  = Artificial Intelligence,
  sort  = AI,
}
\DeclareAcronym{CNN}{
  short = CNN,
  long  = Convolutional Neural Network,
  sort  = CNN,
  plural = s,
}
\DeclareAcronym{GAN}{
  short = GAN,
  long  = Generative Adversarial Network,
  plural = s,
}
\DeclareAcronym{cGAN}{
  short = cGAN,
  long  = Conditional Generative Adversarial Network,
  plural = s,
}
\DeclareAcronym{GDPR}{
  short = GDPR,
  long  = General Data Protection Regulation,
}
\DeclareAcronym{FIM}{
  short = FIM,
  long  = Fisher Information Matrix,
}
\DeclareAcronym{EWC}{
  short = EWC,
  long  = Elastic Weight Consolidation,
}
\DeclareAcronym{uce}{
  short = UCE,
  long  = Unified Concept Editing,
}
\DeclareAcronym{spare}{
  short = SPARE,
  long  = Self-distillation for PARameter-Efficient Removal,
}
\DeclareAcronym{munba}{
  short = MUNBa,
  long  = Machine Unlearning via Nash Bargaining,
}
\DeclareAcronym{PEFT}{
  short = PEFT,
  long  = Parameter Efficient Fine-Tuning,
}
\DeclareAcronym{PEM}{
  short = PEM,
  long  = Parameter Efficient Module,
}
\DeclareAcronym{OMP}{
  short = OMP,
  long  = One-shot Magnitude Pruning,
}
\DeclareAcronym{SFD}{
  short = SFD,
  long  = Score Forgetting Distillation,
}
\DeclareAcronym{DDPM}{
  short = DDPM,
  long  = Denoising Diffusion Probabilistic Models,
}
\DeclareAcronym{FID}{
  short = FID,
  long  = Fréchet Inception Distance,
}
\DeclareAcronym{RT}{
  short = RT,
  long  = Result Template,
  plural = s,
}

\usepackage{listings}
\lstnewenvironment{listing}[1][]
{\lstset{#1}\pagebreak[0]}{\pagebreak[0]}
\lstdefinestyle{consola}
{basicstyle=\scriptsize\bf\ttfamily,
backgroundcolor=\color{gray75},
}
\lstdefinestyle{C}
{language=C,
}

\usepackage{orcidlink}

\usepackage{hyperref}
\usepackage[accsupp]{axessibility} 
\usepackage{cleveref}  
\usepackage{xurl}  

\makeatletter

\newcommand{\inst}[1]{\textsuperscript{#1}} 
\makeatother

\AtBeginDocument{

}

\begin{document}

\frontmatter

\begin{titlepage}
\centering
\vspace*{1cm}
{\LARGE\bfseries I-CARE: Analysis of interference-related phenomena in a controllable, diverse and representative unlearning setting for text-to-image models\par}
\vspace{1.5cm}
{\large
Leonardo S. Benitez Pereira\inst{1}\,\orcidlink{0000-0001-9429-7308}\quad
Marcos Escudero\inst{1}\,\orcidlink{0000-0002-9156-3428}\quad
Luis Herranz Arribas\inst{2}\,\orcidlink{0000-0002-7022-3395}\par}
\vspace{0.6cm}
{\normalsize
\inst{1}\,Universidad Aut\'onoma de Madrid, Spain\\
\inst{2}\,Universidad Polit\'ecnica de Madrid, Spain\par}
\vspace{1.5cm}
\begin{minipage}{0.86\textwidth}
\begin{singlespace}
\noindent\textbf{Abstract.} Machine unlearning studies the removal of knowledge from an AI model, making the system forget a concept it previously learned. Despite rapid progress in generative machine unlearning, the unintended degradation of semantically related concepts that should have been retained (henceforth, interference) remains poorly characterized and inconsistently evaluated. This paper introduces I-CARE, a methodology that formalizes interference as a first-class object of study in generative unlearning. Rather than proposing a new benchmark or unlearning algorithm, I-CARE provides formal definitions for tasks, metrics, and templates for reporting results, enabling the systematic and reproducible study of interference across unlearning settings. While our methodology is designed to remain valid as models and unlearning algorithms evolve, decoupling long-term scientific insight from transient empirical results, we present a feasibility demonstration with state-of-the-art algorithms and frequently used datasets. The results demonstrate that I-CARE enables meaningful analysis of interference patterns across multiple unlearning settings, establishing the practical applicability of the framework. The software implementation of the methodology is provided in an open-source framework, together with a web-based graphical interface that enables exploration of the outcomes of this study without requiring direct interaction with the codebase or specialized data analysis tools.

\vspace{0.4cm}
\end{singlespace}
\end{minipage}
\vfill
{\large \today\par}
\end{titlepage}

\chapter*{Acknowledgments}
\addcontentsline{toc}{chapter}{Acknowledgments}
We are grateful for the support of Carolina R. Kelsch and Natnael Mola in the development of both Vision-Unlearning and Forgety, as they were active contributors during the early development of both. Soham Vaze improved the Vision-Unlearning documentation, tutorials, and the implementation of the UCE method. Moreover, we thank the contributions of Juan C. SanMiguel Avedillo and Fabien Baldacci as supervisors of former projects that were instrumental in the development of I-CARE.

This work is part of the Fair, Explainable and Sustainable Transfer and Adaptation of computer vision models (FESTA) (PID2024-156186OB-I00) project, funded by the Ministerio de Ciencia, Innovaci\'on y Universidades of the Spanish Government.

\begin{singlespace}
\tableofcontents
\end{singlespace}

\mainmatter
\input{content}

\end{document}

%% file: content.tex
\chapter{Introduction} \label{sec:intro}

\ac{MU} is the area of \ac{ML} that studies the removal of data from an \ac{AI} system, making the system unlearn a concept or forget what it learned from a specific data point \cite{shaik2024}.
This area has gained importance over the past few years with the advancement of regulations that provide users with the right to have more control over the data they generate, such as the
General Data Protection Regulation
in the EU \cite{gdpr2016}. With that in force, companies need to be able to remove data from specific users from their trained models. Besides that, \ac{MU} can be used to mitigate biases \cite{zhang2024} and improve model interpretability \cite{choi2025mu_interpretation}, thus contributing to broader efforts toward safe and responsible AI.

Existing unlearning methods for image generation struggle to balance the removal of the target concept with the preservation of unrelated capabilities \cite{shaik2024, Cai2025Slug}, as illustrated in \cref{tab:interference_example}. 
This echoes challenges in fields close to \ac{MU}, such as 
continual learning (in which the term catastrophic forgetting is used to describe the loss of knowledge of a previously learned task when information relevant to the current task is incorporated \cite{kirkpatrick2016}) 
and transfer learning (where negative transfer refers to source-domain knowledge undesirably reducing learning performance in the target domain).
In \ac{MU}, this problem is sometimes referred to as interference, collateral forgetting, or concept adjacency \cite{Thakral2025fade}. To the best of our knowledge, this phenomenon has never been systematically investigated, and no benchmark allows its precise quantification.

\begin{table}[tb]
    \caption{Example of interference using method \textit{\ac{spare}} in 3 different tasks (forgetting people, dog breeds, and scenes). One representative image was selected from before and after unlearning (generated for the same seed), respectively for the entity to be forgotten, one entity to be retained but that suffered interference (respectively Tony Blair, Affenpinscher, and waterfall cataract), and one entity to be retained that was correctly preserved (Arnold Schwarzenegger, Australian Cattle Dog, and basketball court indoor).}
    \label{tab:interference_example}
    
    \centering
    \setlength{\tabcolsep}{1pt}          
    \renewcommand{\arraystretch}{1}
    \setlength{\extrarowheight}{0pt}
    
    \begin{tabular}{
    @{}
    >{\centering\arraybackslash}m{2cm}
    >{\centering\arraybackslash}m{1.4cm}
    m{1.7cm}m{1.7cm}
    m{1.7cm}m{1.7cm}
    m{1.7cm}m{1.7cm}
    @{}
    }
    \hline
    Forget Concept &
    Overwrite Concept &
    \multicolumn{2}{c}{Forget} &
    \multicolumn{2}{c}{Interfered} &
    \multicolumn{2}{c}{Retain} \\
    & &
    \multicolumn{1}{c}{Before} & \multicolumn{1}{c}{After} &
    \multicolumn{1}{c}{Before} & \multicolumn{1}{c}{After} &
    \multicolumn{1}{c}{Before} & \multicolumn{1}{c}{After} \\
    \hline
    
    George W. Bush &
    Kid &
    \includegraphics[width=\linewidth]{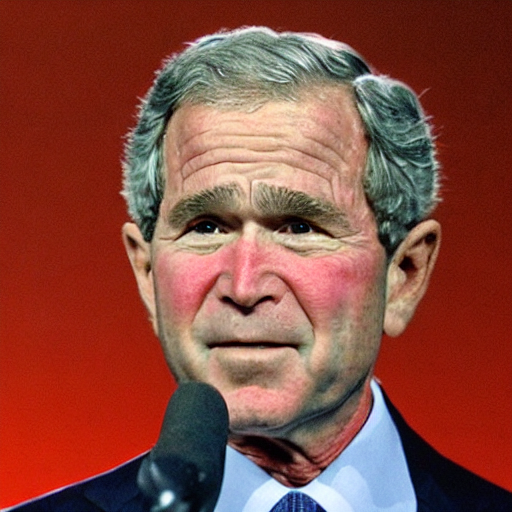} &
    \includegraphics[width=\linewidth]{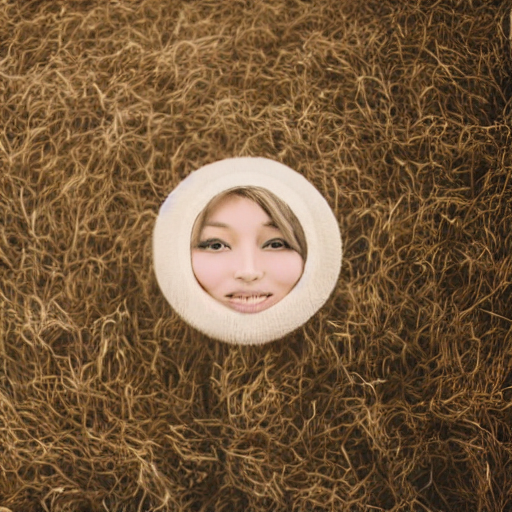} &
    \includegraphics[width=\linewidth]{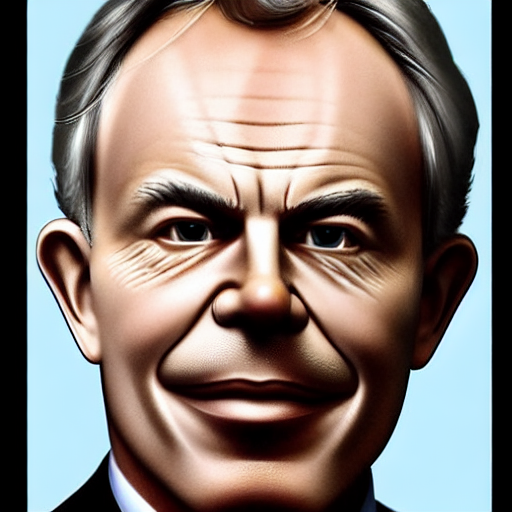} &
    \includegraphics[width=\linewidth]{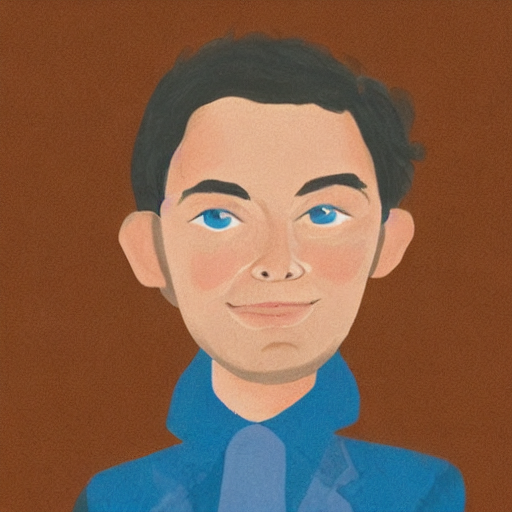} &
    \includegraphics[width=\linewidth]{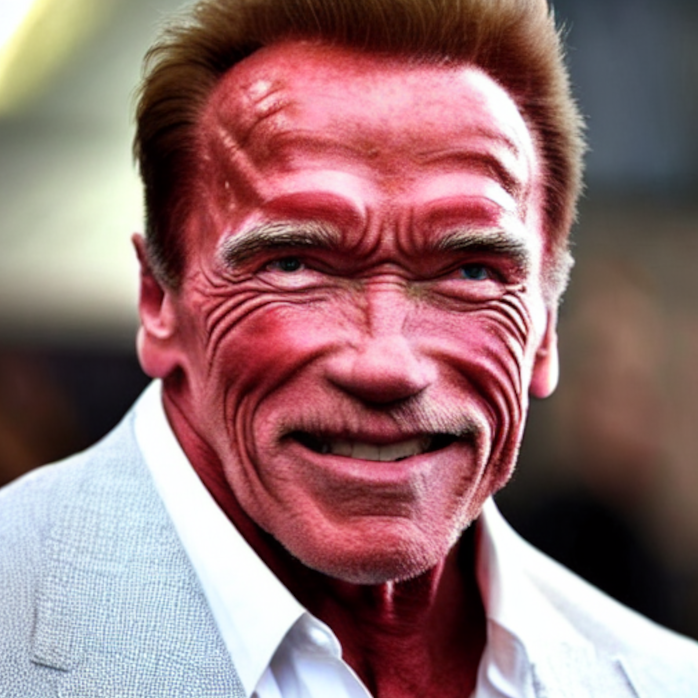} &
    \includegraphics[width=\linewidth]{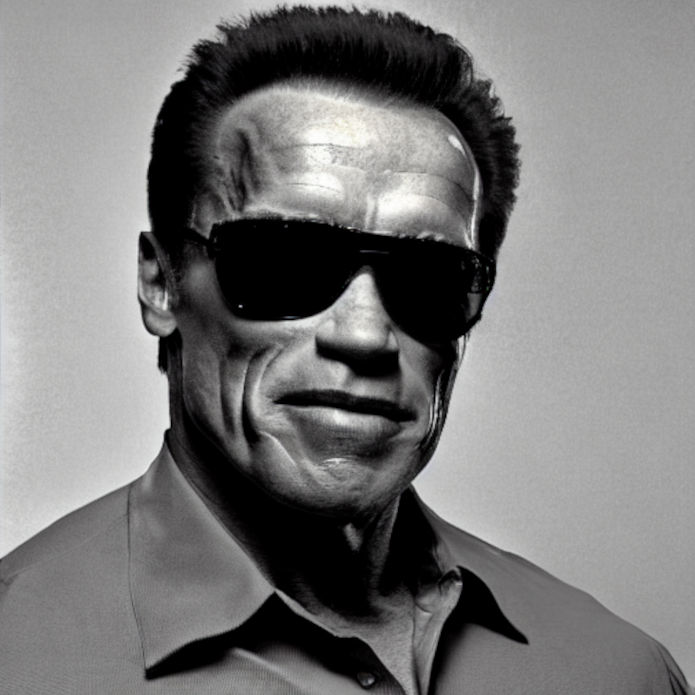} \\
    
    Griffon Bruxellois &
    Cat &
    \includegraphics[width=\linewidth]{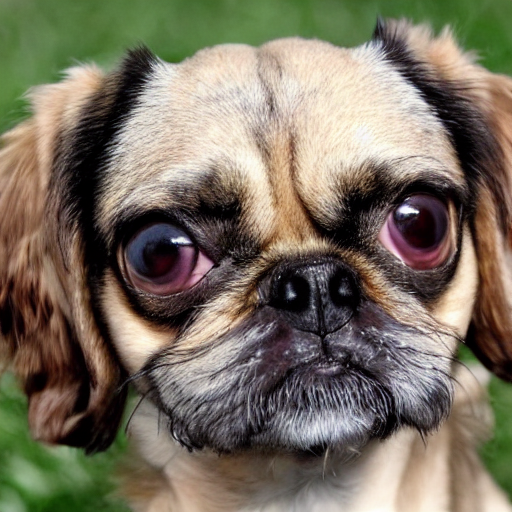} &
    \includegraphics[width=\linewidth]{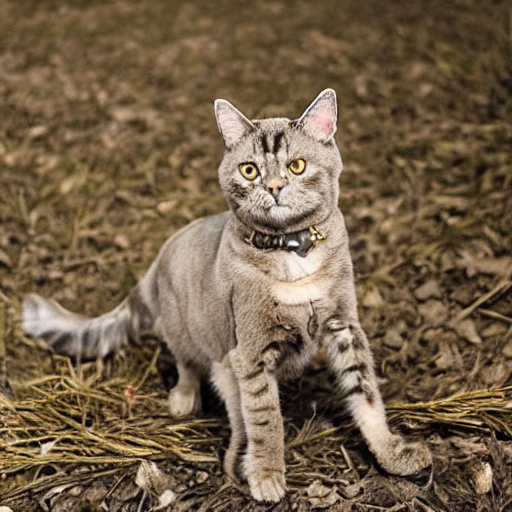} &
    \includegraphics[width=\linewidth]{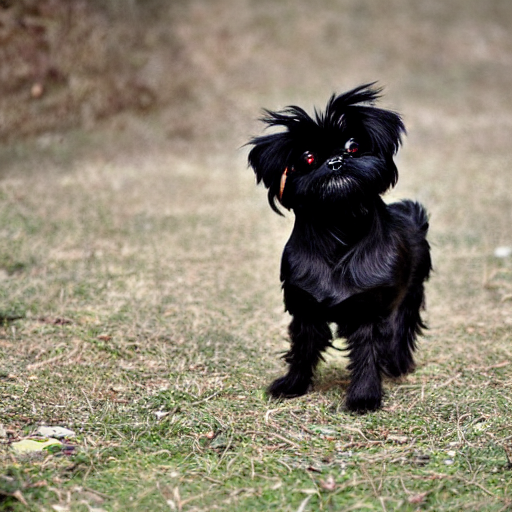} &
    \includegraphics[width=\linewidth]{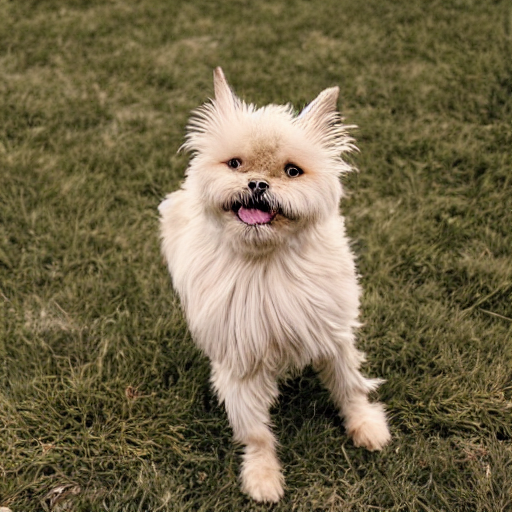} &
    \includegraphics[width=\linewidth]{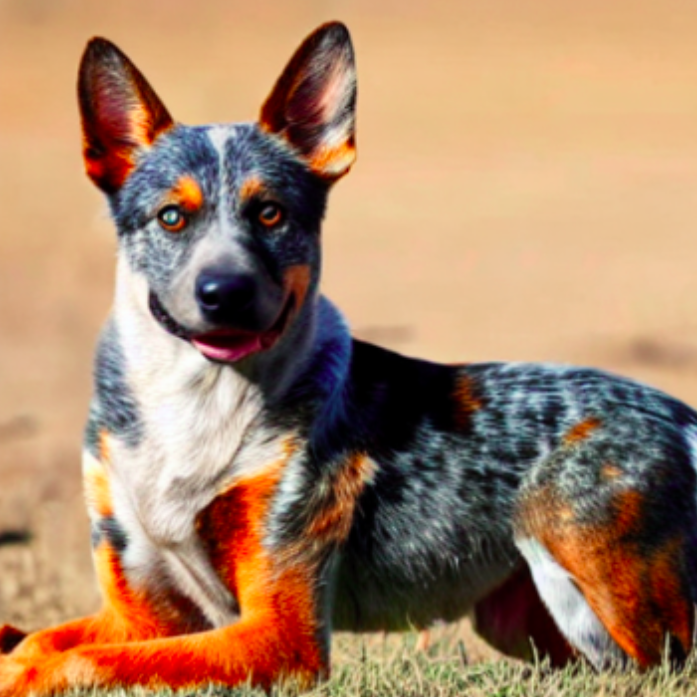} &
    \includegraphics[width=\linewidth]{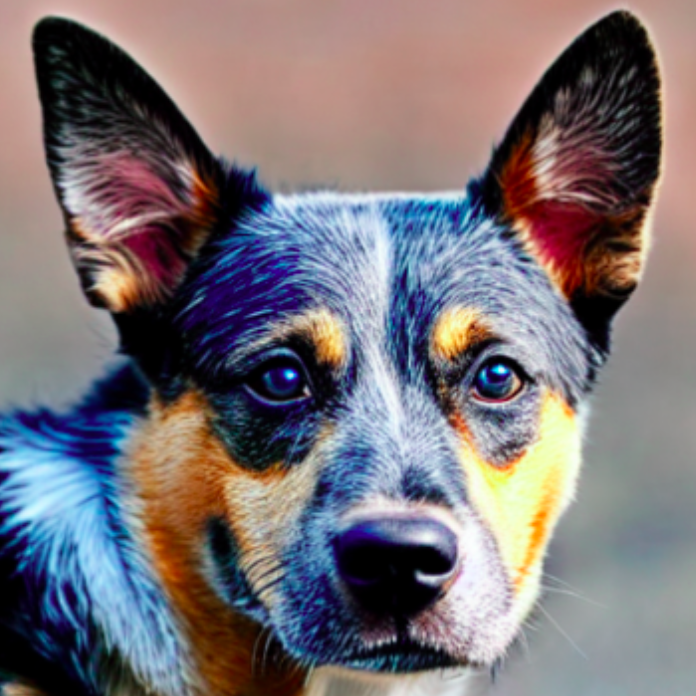} \\
    
    Waterfall Cascade &
    Moon &
    \includegraphics[width=\linewidth]{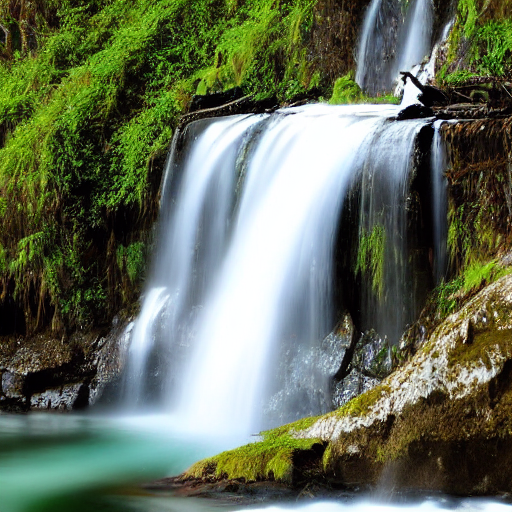} &
    \includegraphics[width=\linewidth]{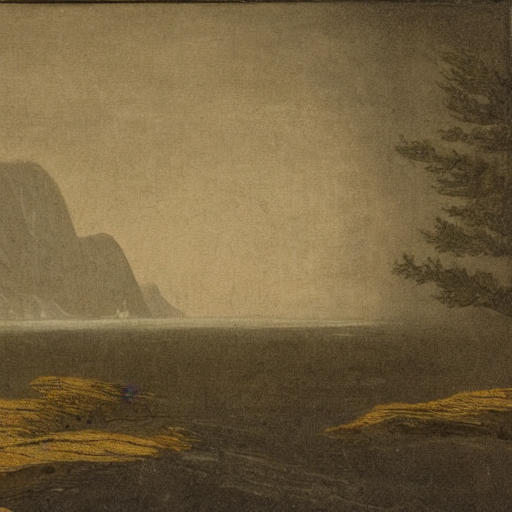} &
    \includegraphics[width=\linewidth]{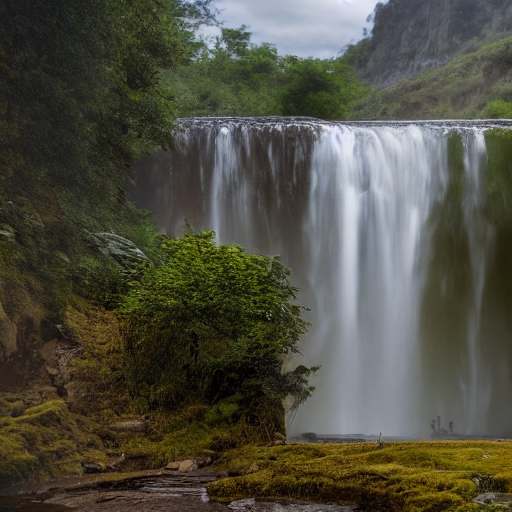} &
    \includegraphics[width=\linewidth]{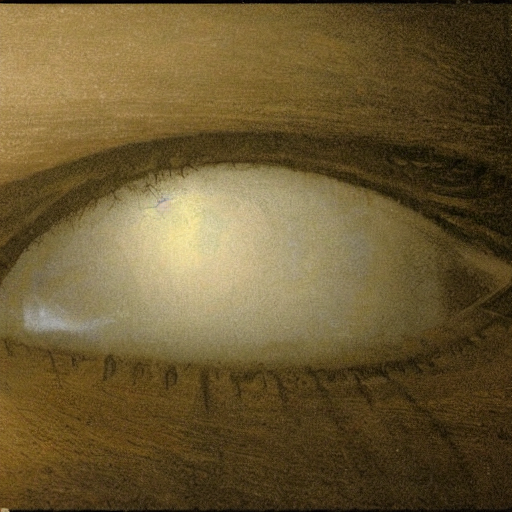} &
    \includegraphics[width=\linewidth]{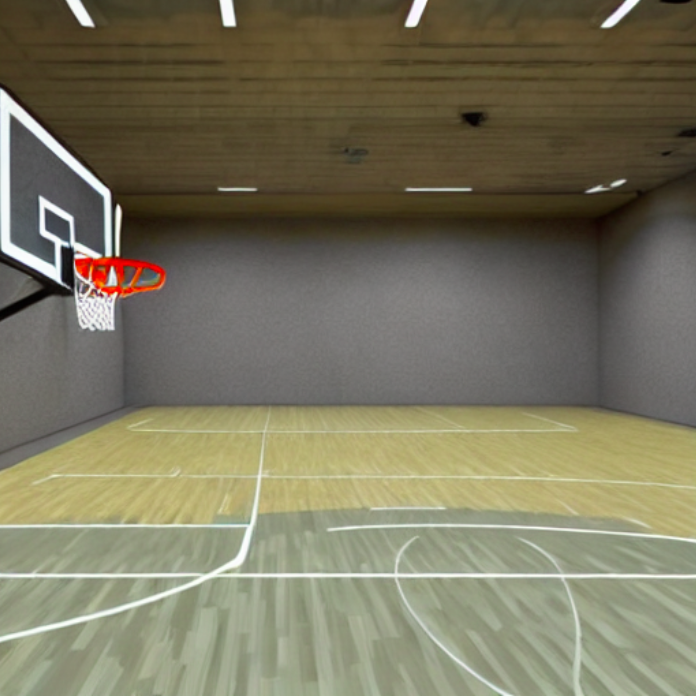} &
    \includegraphics[width=\linewidth]{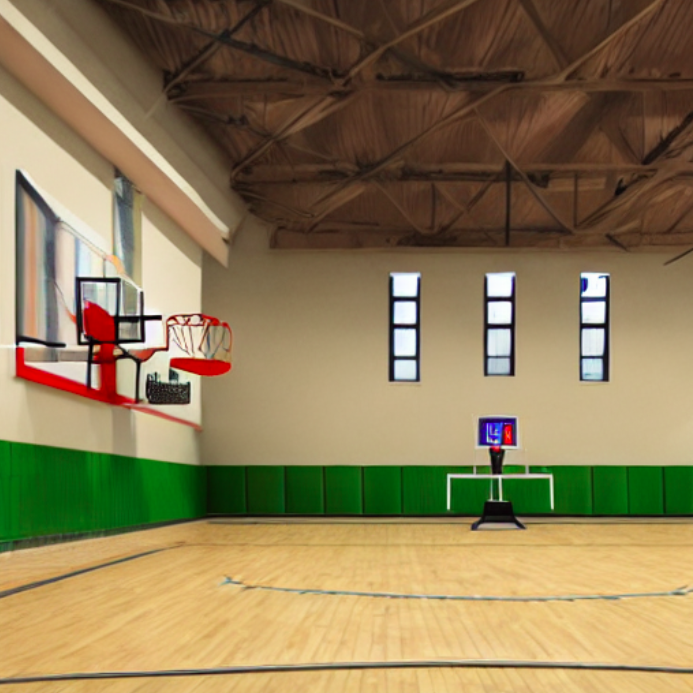} \\
    \hline
    \end{tabular}
\end{table}

This work proposes a methodology for analyzing interference-related phenomena in a controllable, diverse, and representative \ac{MU} setting. More specifically, our novel contributions are threefold: first, a conceptual scheme 
named Interference in Concept Adaptation and geneRative Erasure (I-CARE), that enables the controllable investigation of interference dynamics. Second, a feasibility demonstration showing that the scheme works and reveals non-random interference patterns across several tasks and unlearning methods. Third, an open-source framework for applying the I-CARE methodology 
in novel domains, as well as a visual UI (named Forgety) for non-technical researchers to explore results obtained by I-CARE-compatible studies, thus lowering the barrier to studying the broader implications of interference-related phenomena.

As the state of the art advances, the specific results obtained by the second contribution will eventually become outdated. 
Similarly, the contribution of Forgety goes beyond the software itself, as it exemplifies that downstream applications can uniformly interact with the results produced by several benchmarks if they follow a common structure.
While Forgety provides a visual UI, other possible downstream applications could be Model Context Protocol servers, Structured Query Language interfaces to query the computed results, reinforcement learning training environments, among others.

Nevertheless, the conceptual scheme was designed to withstand such changes, allowing it to be used repeatedly with new base models and unlearning methods. 
Ultimately, the goal of the I-CARE methodology is to accelerate scientific research in several frontiers, including but not restricted to: 
determining whether patterns of interference are consistent across protected attributes; 
inducing rules that approximate the occurrence of interference between new concepts, without the need to execute an expensive evaluation benchmark; 
comparing different unlearning methods by how much interference is observed when they are used with the same base model;
comparing different base models by how much interference is observed when the same unlearning method is applied to them.

\clearpage

\chapter{Literature Review}
\section{Diffusion Models}

Diffusion models have become the dominant architecture for text-to-image generation, achieving high sample quality and strong semantic alignment with natural-language prompts \cite{ho2020denoising,rombach2022stableDiffusion}.  
They generate images by iteratively denoising a latent variable, conditioned on text embeddings produced by a large language encoder, typically via cross-attention layers.  
Because semantic information is distributed across many denoising steps and model blocks, modifying the behavior of a diffusion model requires careful control: even small parameter updates can propagate through the iterative sampling process and alter unrelated visual concepts \cite{fan2023salun}.

Stable Diffusion (SD) \cite{rombach2022stableDiffusion} follows the latent-diffusion design, where denoising operates in a compressed latent space rather than pixel space, enabling training and fine-tuning at a manageable computational cost.  
Its architecture consists of a UNet backbone for iterative denoising, a frozen Clip text encoder for conditioning, and a variational autoencoder (VAE) for mapping between images and latent representations.

\section{Machine Unlearning}


The unlearning task can be formalized as the modification of a \ac{ML} model in order to remove the influence and knowledge learned from part of the dataset, denoted as the forget set $(D_f)$, while retaining knowledge obtained from training on the remaining data, denoted as the retain set $(D_r)$ \cite{cao2015}. 
Optionally, an overwrite set $(D_o)$ may be specified, defining which concepts $D_f$ should map to after the unlearning process \cite{kumari2023}.


We introduce this notation and use it uniformly throughout this paper, even though unlearning methods vary slightly in how these concepts are defined. The most common variations represent these concepts in terms of a dataset (a set of images), a set of prompts used to generate images with a generative model (also known as Data-Free Unlearning \cite{Gandikota2023UCE}), or metadata (for example, a textual description of $D_f$ or some of its properties \cite{Huo2025}).

\section{Interference and related phenomena}

While it is well established that \ac{MU} may harm the generative capabilities of concepts that should be retained \cite{shaik2024, Cai2025Slug}, some concepts are more severely affected during unlearning \cite{Thakral2025fade}. Such unintended degradation of semantically related concepts that should have been retained, referred to in this paper as interference, has several practical consequences for unlearning processes, and its mitigation depends on how we can describe and predict its occurrence. Previous works on the topic did not systematically analyze the problem. For instance, the work of \cite{Thakral2025fade} uses at most four target classes when evaluating interference. These were selected based on the assumption that interference occurs only when the semantic similarity between entities is high. Furthermore, similarity is measured with a pretrained embedding model, which has its own biases.

Similar problems exist in the field of continual learning \cite{li2016lwf,kirkpatrick2016,wang2024mineganpp,imanov2026}. The work of \cite{kirkpatrick2016} defines catastrophic forgetting as the abrupt loss of previously learned task performance that occurs when a network trained sequentially on tasks updates weights to optimize a new task, thereby overwriting parameters important for earlier tasks. This phenomenon was measured by training networks sequentially without replay: constructing multiple tasks via fixed random permutations of the MNIST inputs (for supervised experiments), training on sequences of Atari 2600 games (for reinforcement-learning experiments), among others; after each training segment, the model was evaluated on the test set of every previously seen task to quantify retained performance and average performance across tasks \cite{kirkpatrick2016}.
The work of \cite{imanov2026} predicts catastrophic forgetting by measuring gradient directions with respect to loss functions for different tasks, concluding that their cosine similarity exhibits a strong negative correlation with retention performance.

Another field in which similar problems are observed is transfer learning \cite{Zhang2020survey,ouyang2024TGDP}. In this literature, negative transfer is described as the degradation of target-task performance caused by leveraging knowledge from a source domain that is mismatched, noisy, or otherwise unhelpful. The survey by \cite{Zhang2020survey} characterizes negative transfer operationally through empirical comparison: a transfer-based method is evaluated against a target-only baseline trained with the same algorithmic choices, and negative transfer is identified whenever the inclusion of source knowledge leads to worse generalization on held-out target data. 
When adapting large pretrained generative models to small target datasets, Ouyang et al. \cite{ouyang2024TGDP} emphasize that the dominant issue is not improved transfer but degradation driven by data scarcity, stating that ``training a generative model directly or fine-tuning a pre-trained generative model on limited data from the target domain often results in significant performance degradation due to overfitting and memorization'' \cite{ouyang2024TGDP}. 
Consequently, evaluation in these settings extends beyond aggregate fidelity metrics to include analyses of sample diversity, nearest-neighbor overlap with training data, and utility on downstream tasks, in order to distinguish genuine transfer from memorization-driven artifacts.

\section{Benchmark Methodologies}

A distinct strand of work asks not how to run an evaluation but how to design one whose results are valid. The Common Task Framework (CTF), a term introduced by \cite{liberman2010} for an evaluation practice that matured into a general methodology \cite{donoho2017}, is a structure that lets results from a shared evaluation carry meaning across studies. A CTF-compliant benchmark has three ingredients: a public training dataset, a set of competitors whose common task is to infer a prediction rule from it, and a referee that scores every submitted rule against a test set withheld from the competitors and reports the result automatically. The CTF is thus not a benchmark but a template: a given benchmark, whether an image-classification challenge or a question-answering leaderboard, instantiates that template, and scores obtained under different instantiations are comparable only because they share it.

The structure of a CTF says nothing about the artifacts it circulates. Those artifacts, namely the training and held-out data, the submitted models, and the recorded scores, are reusable across studies only if other researchers and their software can locate, retrieve, and interpret them. The FAIR principles \cite{wilkinson2016} state these requirements as four properties of a digital research object: that it be findable, accessible, interoperable, and reusable. FAIR prescribes properties rather than implementations; it fixes what an artifact must afford, including persistent identification, retrieval through standard protocols, machine-readable metadata, and reuse under explicit terms, without dictating how those affordances are provided. Formulated first for data, the principles were later extended to research software, which becomes relevant once an evaluation is itself distributed as code. FAIR constrains the form of evaluation artifacts but not their meaning: it cannot establish whether the quantity a benchmark reports corresponds to the capability the benchmark claims to measure.

\section{Software-oriented Science}

Computational and \ac{ML} research carries a documented reproducibility problem: reported results often cannot be reproduced because the code or data are missing, and even when both are released, sensitivity to training and configuration conditions makes the original numbers hard to recover \cite{hutson2018}. A method is typically conveyed in prose while the code and data needed to rerun it go unshared in any executable form, despite long-standing calls for both to accompany the publication \cite{stodden2016}.

Software engineering faced comparable coordination failures and resolved them with mechanisms rather than good intentions. Usual remediations involve techinques such as version control, continuous integration, declarative infrastructure, and containerization make coordination a property of the system, so that the system, and not the vigilance of an individual, verifies that results still hold. \ac{ML} systems in production accumulate technical debt from data dependencies and configuration sprawl \cite{sculley2015}, and research code exhibits the same weaknesses: surveys of the repositories attached to published papers find testing, dependency pinning, and containerization largely absent \cite{wolter2025}, and analyses of irreproducibility trace it to experiment-design and configuration choices as much as to missing code or data \cite{gundersen2022}.

The response developed across a growing body of work is to treat the research pipeline as software and hold it to the same engineering standards, on the premise that in computational science the method is the code and a result reproduces only to the extent that its code does \cite{wilson2014, sandve2013, chen2025aria}. In practice this means version control, automated testing, and re-executable workflows for the code behind reported results, in place of analysis scripts discarded once the paper is written. Measured adoption of these practices remains low even though they would directly improve reproducibility \cite{wolter2025}, and existing guidance recommends that code and data be released in a form that regenerates the results rather than merely describing them \cite{stodden2016}.

\clearpage
\chapter{Proposed methodology}
The goal of the I-CARE methodology is to formalize the following phenomenon as a first-class object of study: the unintended degradation of semantically related concepts that should have been retained during unlearning.
The pioneering work of \cite{Thakral2025fade} refers to this phenomenon as ``concept adjacency'', under the implicit assumption that it is caused by concepts lying close to each other in the model's representation space; since I-CARE is not restricted by that assumption, and thus allows the analysis of other sources of degradation (such as biased fine-tuning data or hyperparameter misconfiguration), we do not use this term.
The work of \cite{Thakral2025fade} also proposes the term ``collateral forgetting'', which implies that the phenomenon can only be characterized by its clearest observed consequence; since I-CARE provides a broader framework to analyze what happens during unlearning (such as changes in the strength of association between protected attributes), we also criticize the use of that term.
Instead, we adopt the term ``interference'', also used in \cite{imanov2026}; this term points to the phenomenon itself, instead of its causes or consequences; moreover, it generalizes beyond \ac{MU}, and it allows
didactic analogies with the electromagnetic domain (such as discussing ``isolation'' and ``transmission wires'').

This section establishes the definitions and rules that any valid study of interference-related phenomena must satisfy.
I-CARE is properly understood as a methodology for interference studies rather than a benchmark in the narrow sense: it specifies the formal grammar of valid interference research (the ontology of permitted objects and operations, the constraints on admissible instantiations, and the standard interface for reporting results) without fixing the particular entities, models, or methods to be studied. In short, a benchmark must define a set of entities to be separately unlearned; then, each unlearned model is used to generate images for all entities, allowing fine-grained analysis of the effects caused by the unlearning process.
I-CARE is an empirical and inductive method, providing solely a description of observed interferences. Analyzing the causes of interference is beyond our scope, even though our results can be used together with other studies to support causal conclusions.

We start by providing standardized naming for concepts used in I-CARE, alongside a mathematical notation used to complement the explanation. We then restrict the space of possible choices for each concept, ensuring that results obtained following the I-CARE conceptual scheme are meaningful and methodologically sound. We end this section by defining \acp{RT}, predefined ways of reporting results that can directly support quantitative and qualitative claims about interference in a given domain.

\section{Formal definitions} \label{sec:definitions}

For each concept, we differentiate between one specific instance of the concept (denoted by the first letter in lowercase, 
analogous to the convention for a ``object'' in object oriented programming) and the space/set of possible instances of the concept (denoted by the first letter in uppercase, similar to a ``class'' in object-oriented programming). \textit{Italic} and camelCase are used when referring to concepts or their values, unless the word is used in its common meaning. If more than one function argument possesses the same type, they are subscripted by an integer number.

{\bf Entity $e$}. A distinct semantic concept that may be subject to unlearning and may cause or receive interference. \underline{Formalization}: a member of a finite set $E$. \underline{Example of entity}: \textit{GeorgeBush}.

{\bf Attribute $a$}. Annotated property of an \textit{entity}. One \textit{entity} possesses several \textit{attributes}, and an \textit{attribute} $a$ can be scalar or categorical. \underline{Formalization}: function $a: E \to V$, for an \textit{attribute} $a$ with value space $V$; a scalar \textit{attribute} $a$ can assume any real number as a value; a categorical \textit{attribute} $a$ can assume values from a finite set $a_{1}, a_{2}, a_{3}, \dots, a_{|a|}$, where $|a|$ denotes the number of possible values for $a$. \underline{Example}: \textit{profession} (a hypothetical categorical \textit{attribute} assuming the possible values \textit{politician}, \textit{artist}, or \textit{athlete}).

{\bf Task $t$}. Set of related \textit{entities}, all of which are annotated with the same \textit{attributes}. 
\underline{Formalization}: $t \subseteq E$, annotating the \textit{attributes} $\{a_1, a_2, \dots , a_{|a|}\}$. 
\underline{Example of task}: \textit{forgettingPoliticians}.

{\bf Model $m$}. Function that generates an image given an \textit{entity}. Generative parameters (such as the number of denoising steps) are considered part of the \textit{model}. 
\underline{Arguments}: \textit{entity}.
\underline{Example}: \textit{stableDiffusion} \cite{rombach2021stable_diffusion}.

{\bf UnlearningMethod $u$}. Creates a new \textit{model} that no longer contains knowledge about the requested \textit{entity}. Hyperparameters and other forms of configuration are considered part of the \textit{unlearningMethod}. 
\underline{Arguments}: \textit{entity}, \textit{model}. 
\underline{Formalization}: $u : E \times M \to M$. 
\underline{Example}: \textit{\ac{uce}} \cite{Gandikota2023UCE}.

{\bf LatentEmbedding $l$}. Vector-valued function that computes an $n$-dimensional representation of images generated by a \textit{model} for an \textit{entity}. Potentially internally uses an auxiliary foundation model, unrelated to the generative \textit{model}.
\underline{Arguments}: \textit{entity}, \textit{model}. 
\underline{Formalization}: $l : E \times M \to \mathbb{R}^n$. 
\underline{Example}: computing the 768-dimensional vector using the end-of-text output token of a CLIP model (henceforth, \textit{clipEmbedding} \cite{Radford2021Clip}).

{\bf SimilarityBetweenEntities $s$}. Scalar-valued function that measures proximity between two \textit{entities}, following some human-intuitive notion of ``proximity''. 
\underline{Arguments}: \textit{entity\textsubscript{1}}, \textit{entity\textsubscript{2}}. 
\underline{Formalization}: $s : E \times E \to \mathbb{R}$. 
\underline{Example}: \textit{clipEmbedding} $(entity_1)^T$ \textit{clipEmbedding} $(entity_2)$ (also referred to as \textit{clipCosineSimilarity} \cite{Radford2021Clip}).

{\bf MetricQuality $q$}. Scalar-valued function that measures how well a \textit{model} generates an image for a given \textit{entity}.
\underline{Arguments}: \textit{entity}, \textit{model}. 
\underline{Formalization}: $q : E \times M \to \mathbb{R}$. 
\underline{Examples}: probability of the generated images being associated with the desired entity by a pretrained classifier \cite{fan2023salun}, a reference-free image complexity measure \cite{mittal2012brisque}, or the CLIP similarity score \cite{Radford2021Clip} between prompt and image (from now on simply referred to as $Clip$).

{\bf MetricInterferencePerEntityPair $m_p$}. Scalar-valued function that measures, when forgetting \textit{entity\textsubscript{1}}, how much worse the generation quality of \textit{entity\textsubscript{2}} becomes. 
Potentially internally averages across several seeds and assumes a certain \textit{metricQuality}, such as in $m_p = \frac{1}{|seeds|} \sum_{seeds} \big(q(e_2, m) - q(e_2, u(e_1, m))\big)$. 
\underline{Arguments}: \textit{entity\textsubscript{1}}, \textit{entity\textsubscript{2}}, \textit{model}, \textit{unlearningMethod}.
\underline{Formalization}: $m_p : E \times E \times M \times U \to \mathbb{R}$. 
\underline{Example}: difference between $Clip$ of \textit{entity\textsubscript{2}} before and after unlearning \textit{entity\textsubscript{1}} (henceforth, $\Delta Clip$).

{\bf MetricInterferencePerEntity $m_e$}. Scalar-valued function that measures an inter\-fe\-re\-nce-related aspect of a single \textit{entity}, potentially by aggregating \textit{metricInterferencePerEntityPair} across all \textit{entities} of a \textit{task}. 
\underline{Arguments}: \textit{entity}, \textit{model}, \textit{unlearningMethod}.
\underline{Formalization}: $m_e : E \times M \times U \to \mathbb{R}$.
\underline{Example}: average, across all \textit{entity\textsubscript{i}}, of the difference between $Clip$ of \textit{entity\textsubscript{i}} before and after unlearning \textit{entity} (or simply \textit{avgClipEmitter}); or average, across all \textit{entity\textsubscript{j}}, of the difference between $Clip$ of \textit{entity} before and after unlearning \textit{entity\textsubscript{j}} (or simply \textit{avgClipReceiver}).

Other concepts could be formalized but are not strictly necessary for a basic notation. Similarly, types other than numerical and categorical could be defined, as well as nullable \textit{attributes}. 
Furthermore, no internal performance details are discussed unless clarifying the concept. Naturally, in the software implementation, all images are generated ahead of time, as well as all unlearning processes, and we just ``keep track'' of which result refers to which \textit{model} and \textit{unlearningMethod}. 
We expect that the meaning of function arguments can be deduced from their context. 
For example, for a function that receives a \textit{model}, \textit{unlearningMethod}, and \textit{entity}, it is reasonable to interpret that its internal implementation involved the stateless operation $u(e_2, m)(e_1)$ (or, equivalently, fetching from a data store the images previously generated using the previously unlearned \textit{model}). 

\section{Task choice}

While any subset of possible entities could constitute a \textit{task}, researchers must carefully select a diverse and representative set of \textit{entities} in order to draw meaningful conclusions, as well as make pragmatic design choices that ease the practical observation of interference.
With this in mind, we impose the following constraints in the choice of tasks:

\begin{enumerate}
    \item \textit{Entities} should be fine-grained (that is, they represent similar concepts, so as to maximize the observation of interference \cite{Thakral2025fade}).
    \item \textit{Entities} should be nameable in natural language, and most off-the-shelf foundation \textit{models} should recognize the \textit{entity} given its name/short string, as opposed to anonymous instances.
    \item For each \textit{entity}, there should be a set of at least 10 images. We highly recommend using real images, since generated images vary in quality between \textit{entities} and generative models have biases of their own. All \textit{entities} should contain the same number of images.
    \item All \textit{entities} should be generated equally well by the chosen \textit{model}.
    \item All \textit{entities} should be at the same ``level of abstraction''; there should not be a case in which one entity ``contains'' another, so in a perfect unlearning session one entity does not affect any other. For example, if \textit{``George W. Bush''} and \textit{``the USA president during 9/11''} were two \textit{entities} of a \textit{task}, then this criterion would be violated.
    \item An initial pool of \textit{entities} (that is, before the \textit{data balancing} procedure described in \cref{sec:balancing}) should ideally contain at least 200 \textit{entities}.
\end{enumerate}

\section{Attribute choice}

The annotated \textit{attributes} play a crucial role when analyzing broader societal impacts of interference-caused problems. As such, the choice of \textit{attributes} to be annotated, as well as the procedure used to annotate the \textit{attributes} of each \textit{entity}, must conform to the following guidelines:

\begin{enumerate}
    \item Researchers should be able to annotate \textit{attributes} at the class level; if not possible, annotations at the image level (such as those often present in image classification datasets) may be aggregated per \textit{entity}. Aggregating annotations at the instance level (for object detection tasks, with multiple objects in the scene) is strongly discouraged.
    \item \textit{Attribute} values should be unambiguous and their definitions undisputed.
    \item \textit{Attributes} should be immutable per \textit{entity} (that is, one \textit{entity} should have a single value for each \textit{attribute}, approximately stable over time). This eliminates picture-specific \textit{attributes} such as pose, mood, etc.
    \item A clear methodological procedure to annotate \textit{attributes} should be defined by the researchers, including the disclosure of possible biases introduced by the annotation itself. Well-established authoritative datasets are preferred over custom inference rules.
\end{enumerate}

\section{Data preparation} \label{sec:balancing}

Since the training images directly impact the \textit{unlearning} process, researchers should take special care to ensure that every retain set contains all other \textit{entities}. Moreover, the distribution of images in both forget and retain sets should be similar across \textit{entities}, decreasing the possibility of new external biases being introduced.

\subsection{Data balancing}
If conclusions are to be drawn based on specific \textit{attributes} (such as socially relevant dimensions), the dataset should be balanced across those \textit{attributes} (approximately the same number of \textit{entities} per \textit{attribute} combination). We refer to this process as \textit{data balancing}. 
Furthermore, it is of paramount importance to document preprocessing steps performed to make this process feasible; for instance, it may be necessary to remove low-frequency categories, bin numerical \textit{attributes}, and/or perform other preprocessing operations on the \textit{attributes} to ease \textit{data balancing}.

\section{Interference measurement}

I-CARE quantifies interference by unlearning $entity_i$ and evaluating its impact on all $entity_j$ of the same \textit{task}, using one or more metrics $m_p(e_i, e_j, m, u)$ as well as one or more metrics $m_e(e_i, m, u)$, for every $entity_i$ in the \textit{task}. If interference is high, the generative quality of the unlearned \textit{model} will differ from the original (better or worse). If the goal of the benchmark is to compare \textit{models}, we recommend choosing at least 3 different base \textit{models}. Similarly, 3 or more \textit{unlearningMethods} if the aim is to compare \textit{methods}. In any case, in the absence of a theory that predicts which pairs of entities will show interference, this all-vs-all strategy is the safest strategy for analyzing interference dynamics.

\subsection{Equalization}
Since the goal of I-CARE is to compare how much interference is caused, rather than the ability to forget \textit{per se}, we recommend configuring \textit{unlearningMethods} such that they achieve comparable levels of unlearning quality. That is, controlling for unlearning quality in order to estimate the direct effect of \textit{unlearningMethod} choice on observed interference, excluding the mediation effect induced by unlearning quality. 
We refer to this unlearning adjustment procedure as \textit{equalization}. In contrast, a non-equalized setup would require explicitly accounting for unlearning quality during the analysis, but would enable joint benchmarking of methods with respect to both their unlearning effectiveness and their interference dynamics.

\section{Result Templates}

Each \ac{RT} provides a tool to answer potentially interesting questions about a specific dataset. 
That is, they declare using the I-CARE notation how data will be aggregated and plotted, thus easing comparisons between studies and increasing the overall reproducibility of I-CARE-compatible works.
We strongly recommend researchers to report their results exclusively in terms of RTs or simple aggregations of RT results, instead of ad hoc or one-time analyses.

Each \ac{RT} is uniquely identified by a name; it receives parameters (whose types are exclusively concepts defined in Section~\ref{sec:definitions}) and returns standardized results (that can be saved in a json file as specified in \cref{ap:metadata}). We provide the interpretation of each \ac{RT} so their internal workings become irrelevant and the user can treat these definitions as pure notation. When an \ac{RT} involves a statistical significance test, one parametric test and one non-parametric test are computed, and the user is free to choose which one to analyze depending on the broader context.

{\bf MetricMetricAlignment}. Measures how strongly two \textit{MetricInterferencePerEntity} metrics are correlated. 
\underline{Arguments}: $m$, $t$, $u$, $m_{e1}$, $m_{e2}$. 
\underline{Result}: Pearson p-value, Spearman p-value, Pearson correlation, scatter plot. 
\underline{Interpretation}: quantitative; the higher the correlation, the lower the need to calculate both metrics for this specific choice of $m$, $t$, and $u$.

{\bf MetricSimilarityAlignment}. Measures to what degree similar \textit{entities} interfere more with each other.
Formalized in \cref{ap:prediction}, which also proposes its natural expansion to a multivariable and non-linear predictive regression. 
\underline{Arguments}: $m$, $t$, $u$, $m_p$, $s$, optional $coloringAttribute$. 
\underline{Result}: Pearson p-value, Spearman p-value, Pearson correlation, scatter plot. 
\underline{Interpretation}: quantitative; if this value is high, interference between two \textit{entities} can be approximated by \textit{similarity} (which is cheaper to compute for any new \textit{entity}); equivalently, the amount of ``transmission wires'' can be summarized by this single \textit{similarity} function; the optional parameter $coloringAttribute$ specifies one \textit{attribute} that, for visualization purposes, colors the dots of the scatter plot based on the \textit{attribute value} of the emitter \textit{entity}, allowing the researcher to qualitatively identify intra-task patterns.

{\bf InterferenceMatrix}. \textit{MetricInterferencePerEntityPair} between each possible combination of two \textit{entities} within a \textit{task}.
\underline{Arguments}: $m$, $t$, $u$, $m_p$. 
\underline{Result}: $|t|\times|t|$ real-valued tensor. 
\underline{Interpretation}: qualitative; visual patterns may be spotted, especially when rearranging indices in a meaningful manner (for example, grouping professions together). Further quantitative values may be derived, such as the average value or the ratio between the diagonal-average value and the non-diagonal-average value.

{\bf SimilarityMatrix}. \textit{Similarities} between each possible combination of two \textit{entities} within a \textit{task}.
\underline{Arguments}: $m$, $t$, $s$. 
\underline{Result}: $|t|\times|t|$ real-valued tensor. 
\underline{Interpretation}: qualitative; visual patterns may be spotted, similarly to \textit{InterferenceMatrix}.

{\bf SignificantRelationshipNumerical}. Measures whether a numerical \textit{attribute} is significantly correlated with a \textit{MetricInterferencePerEntity}. 
Formalized in \cref{ap:rt_relationship}. 
\underline{Arguments}: $m$, $t$, $u$, $m_e$, $a$. 
\underline{Result}: Pearson p-value, Spearman p-value, Pearson correlation, scatter plot. 
\underline{Interpretation}: qualitative; the researcher should decide if it is ethical or desirable that this \textit{attribute} propagates interference.

{\bf SignificantRelationshipCategorical}. Statistical significance of the average \textit{Me\-tric\-Inter\-ference\-PerEntity} across all \textit{entities}, when grouped by each of its values. Formalized in \cref{ap:rt_relationship}, which also describes a simple extension that allows the analysis of interference flow distribution (such as, for instance, whether \textit{politicians} cause more interference to other \textit{politicians} than \textit{artists} cause to other \textit{artists}).
\underline{Arguments}: $m$, $t$, $u$, $m_e$, $a$. 
\underline{Result}: ANOVA p-value, Kruskal-Wallis p-value, 
average value of $m_e$ grouped by each value of $a$, grouped box plot. 
\underline{Interpretation}: qualitative; similar to \textit{SignificantRelationshipNumerical}.

{\bf CountSignificantRelationship}. Number of significant relationships across all combinations of \textit{attributes} and \textit{MetricInterferencePerEntity}.
\underline{Arguments}: $m$, $t$, $u$, list of $m_e$, list of $a$.
\underline{Result}: integer, list of significances.
\underline{Interpretation}: quantitative; the lower the better. Since the \textit{attributes} for which it is ethical to propagate interference are constant across all \textit{models} and \textit{methods}, a higher value directly implies a higher number of ethical violations, that is, a larger number of ``transmission wires'' in a given \textit{task} effectively used by this \textit{method} and \textit{model}.

{\bf MethodComparisonByMetricEntity}. Compares the distribution of one \textit{MetricInterferencePerEntity} across multiple \textit{unlearningMethods}.
\underline{Arguments}: $m$, $t$, $m_e$, list of $u$.
\underline{Result}: per-method mean, median, standard deviation, $n$, list of values; box plot.
\underline{Interpretation}: quantitative; if lower values are preferred for this $m_e$, then \textit{methods} with lower distributions can be considered to perform better.

{\bf ImplicitAssociationTest}. Measures how the strength of automatic associations $B$ \cite{Sirotkin2022,Steed2021} between two pairs of \textit{entities} changes after unlearning ($\Delta B$, as formalized in \cref{ap:rt_iat}). 
\underline{Arguments}: $m$, $t$, $u$, $a_1$, $a_2$, $l$. 
\underline{Result}: $|a|\times|a|$ real-valued tensor $\Delta B$. 
\underline{Interpretation}: qualitative; a human should decide whether it is ethical or desirable for the unlearning process to cause this change in implicit association between the chosen \textit{attributes}.

{\bf MinimumCutInterference}. Interprets a \textit{task} as a directed weighted graph and computes the minimum cut separating two \textit{entities}, as formalized in \cref{ap:graph}. As a consequence of the max-flow min-cut theorem, it directly follows that the minimum cut is the smallest influence whose removal eliminates every directed influence path from $e_1$ to $e_2$ (see formal proof in \cref{ap:flow_isolation}). Based on this, we conjecture that if we need to unlearn $e_1$ while minimizing harm to $e_2$, then the ideal intervention in the unlearning process is to increase the preservation of the emitter-side nodes. More intuitively, we can think of this intervention as ``blocking the interference path,'' as performed in electrical circuits to protect sensitive components (such as ground partitioning, shielding traces, among others \cite{Zhong2025emi}). 
\underline{Arguments}: $m$, $t$, $u$, $e_1$, $e_2$, $m_p$. 
\underline{Result}: list of \textit{entities} (corresponding to the emitter-side nodes). 
\underline{Interpretation}: qualitative; small set of nodes through which most of the interference from $e_1$ propagates to $e_2$.

\clearpage
\chapter{Feasibility demonstration}

Building upon the I-CARE methodology, we develop a feasibility demonstration to empirically study interference effects in generative unlearning.
The experimental setup involves one base \textit{model} (\textit{Stable Diffusion~1.4}), three state-of-the-art \textit{unlearningMethods}, four random seeds for image generation, four \textit{SimilarityBetweenEntities} functions, five \textit{InterferencePerEntityPair} metrics, and thirty-nine \textit{MetricInterferencePerEntity} metrics. 
Three \textit{tasks} were selected, each composed of 100 \textit{entities}.
As such, the exhaustive execution of all \acp{RT} over all valid combinations of inputs requires unlearning 900 \textit{models} and generating 360000 images, alongside computation of all metrics and \textit{similarity} functions. \cref{ap:materials} describes the hardware and software utilized.

As argued by \cite[pg. 760]{donoho2017}, ``an article about computational science in a scientific publication is not
the scholarship itself, it is merely advertising of the scholarship. The actual scholarship is the complete software development environment and the complete set of instructions which generated the figures''. As such, only a few informative results are discussed in \cref{sec:results}, with a larger set of results provided in \cref{ap:results}, and the complete set of precomputed results available in a HuggingFace repository
\footnote{
https://huggingface.co/datasets/LeonardoBenitez/VisionUnlearningEvaluationTestbeds
}
, following a standardized naming convention and indexed via metadata files (described in \cref{ap:metadata}), and can be explored through the Forgety UI (\cref{sec:forgety} describes its architecture and functionalities). 

Its software implementation closely follows the I-CARE naming and function signatures, and extensive documentation of the concept instantiations (for example, the implemented metric function for $\Delta Clip$) is provided.
The entire demonstration is implemented as fully free and open-source software, integrated into the 
Vision-Unlearning library
\footnote{
https://pypi.org/project/vision-unlearning/
}. The implementation is modular by design, allowing straightforward reuse and extension in future works that adhere to the I-CARE methodology.

\section{Design choices}
We define three \textit{tasks} leveraging existing image datasets, chosen to satisfy the methodological constraints while minimizing the implementation effort: unlearning one person from the dataset Labeled Faces in the Wild (LFW) \cite{dataset_lfw}, unlearning a dog breed from the dataset AtharvaTaras Dog Breeds Dataset \cite{dataset_taras_dog_breeds}, and unlearning a scene from the SUN Attributes dataset \cite{dataset_sun_attributes}. From now on, these \textit{tasks} will be referred to as \textit{breeds}, \textit{scenes}, and \textit{people}. Justification for the better suitability of these tasks compared to others is discussed in \cref{ap:tasks}. 

Each \textit{entity} is annotated with two \textit{attributes} that are used both for \textit{data balancing} and for the majority of the reported downstream analyses. These \textit{attributes} are selected to be meaningful for social or semantic analysis, approximately independent from each other, and feasible to balance across \textit{entities}. We refer to these as \textit{attributesOfInterest}. Any additional \textit{attributes} annotated are referred to simply as \textit{attributes}.
While SUN \cite{dataset_sun_attributes} already had annotated \textit{attributes}, the additional attribute datasets \cite{dataset_akc,dataset_pawsomeauthority} and \cite{dataset_pantheon} were used for \textit{breeds} and \textit{people}, respectively.
The precise preprocessing and complete final set of \textit{attributes} are described in \cref{ap:attributes}.

\begin{figure}[tb]
  \centering
  \begin{subfigure}{\linewidth}
    \centering
    \includegraphics[width=0.8\linewidth]{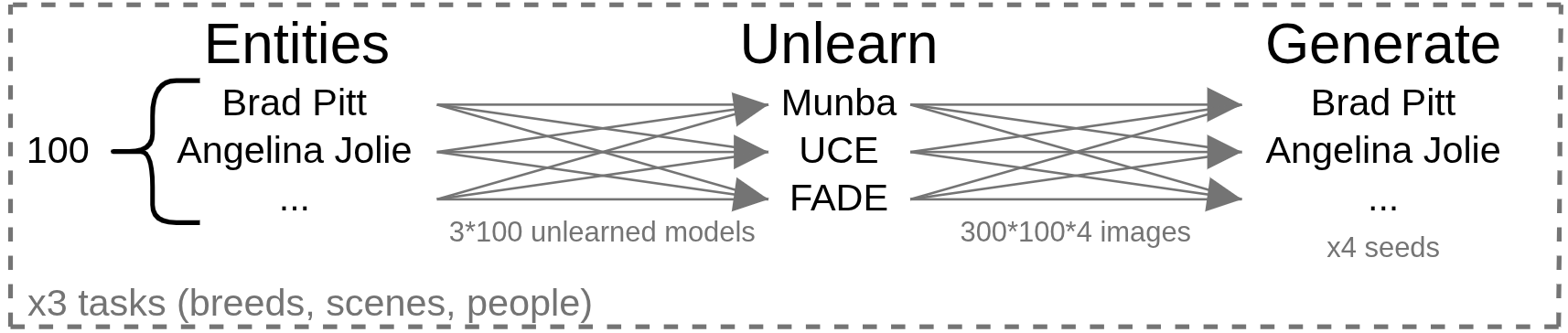}
    \caption{Testbed setup pipeline, including data preparation, unlearning, and image generation.}
    \label{fig:pipeline_testbed}
  \end{subfigure}

  \vspace{0.5em}

  \begin{subfigure}{\linewidth}
    \centering
    \includegraphics[width=0.8\linewidth]{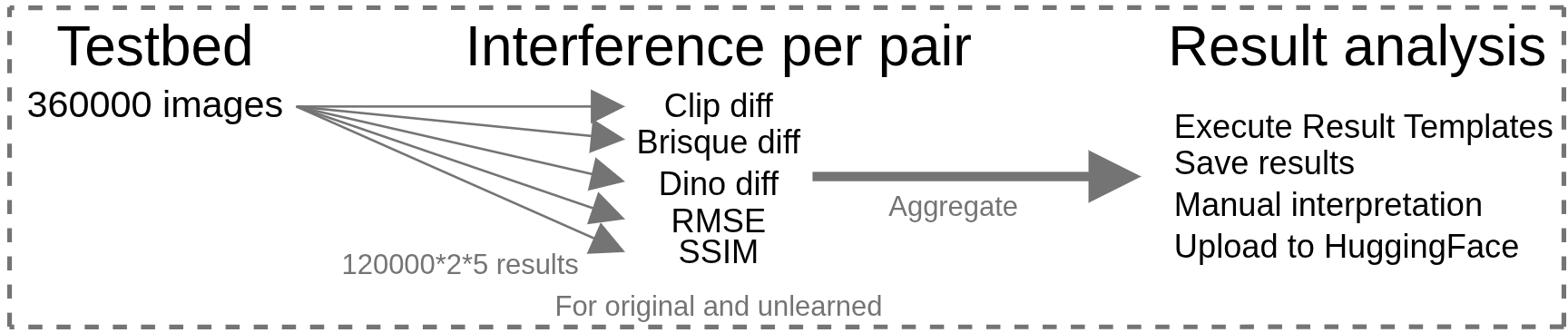}
    \caption{Interference measurement pipeline, covering metric computation, RT execution, and analysis.}
    \label{fig:pipeline_interference}
  \end{subfigure}

  \caption{Overview of the experimental workflow. This design enables full reproducibility of the interference analysis without rerunning the computationally expensive unlearning and image generation stages, as well as allowing partial reuse of the testbed in future studies.}
  \label{fig:pipeline}
\end{figure}

The \textit{similarity} between \textit{entities} was measured by four independent techniques, each capturing different aspects that may potentially be relevant for the chosen \textit{tasks}: one text-only method (\textit{Clip}), one image-only method (\textit{Dino}), one mechanistic activation-based method (\textit{Act}), and a simple intersection-based metric computed over the sets of annotated \textit{attributes} associated with each \textit{entity} (\textit{Jacc}).
The chosen \textit{InterferencePerEntityPair} metrics
reflect different perspectives on image quality: changes in semantic alignment with respect to the textual prompt (measured by $m_p$ \textit{$\Delta$Clip} \cite{Radford2021Clip}), reference-less structural image quality (\textit{$\Delta$Brisque} \cite{mittal2012brisque}), changes in the DINOv2 representation (\textit{$\Delta$Dino} \cite{oquab2023dinov2}), pixel-wise changes (\textit{RMSE}), and structural changes in image content (\textit{SSIM} \cite{wang2004ssim}).
From these, we derive 49 \textit{InterferencePerEntity} metrics using a simple combinatorial aggregation procedure. 
The complete description of the chosen metrics is provided in \cref{ap:metrics}.

We evaluated three \textit{unlearningMethods}, selected to represent the main algorithmic families in the unlearning literature: one data-free method (\textit{\ac{uce}} \cite{Gandikota2023UCE}), one distillation-based method (\textit{\ac{spare}} \cite{kelsch2026fade}, also referred to in the code simply as \textit{distil}), and one game-theoretic method (\textit{\ac{munba}} \cite{Wu2024_munba}). 
We performed \textit{equalization} in the \textit{scenes} and \textit{people} tasks, while \textit{breeds} was not equalized (hyperparameters were optimal for each \textit{method}), thus allowing to analyze the effects of the \textit{equalization} procedure. Equalization was performed by manually tuning hyperparameters such that \textit{$\Delta$Clip} in forget and retain sets is as similar as possible across the three \textit{methods}. The main hyperparameters of each \textit{method} are listed in \cref{ap:unlearning_method}.

After selecting the 100 \textit{entities} for each \textit{task}, intersectionally \textit{balanced} across the \textit{attributesOfInterest}, all unlearning sessions were executed and their resulting \textit{models} were used to generate the necessary images. All \acp{RT} were executed over all valid combinations of inputs. As an illustrative example, the \ac{RT} \textit{MetricSimilarityAlignment} (with parameters: $m$, $t$, $u$, $m_p$, $s$) is executed 180 times: for each of the three \textit{tasks}, accounting for the combination of three \textit{unlearningMethods}, five \textit{InterferencePerEntityPair} metrics, and four \textit{similarity} functions. 
To abstract the complexity of the software implementation, the workflow is explicitly separated into two sequential ``processing pipelines'' (\cref{fig:pipeline}).


\section{Seed implementation mistake} \label{sec:seed}

An important implementation mistake was only noticed after the image generation for \textit{scenes} and \textit{people} was completed: images that should have been generated with the same seed were actually generated with different seeds. Due to the heavy computational requirements of generating the images, the mistake was only corrected for the image generations of the \textit{breeds} task. The practical implication is that the interference metrics became noisier, especially for \textit{RMSE} and \textit{SSIM} (which directly compare images before and after unlearning).

\section{Analysis of specific results} \label{sec:results}

\FloatBarrier
\subsection{Relationships between attributes and interference.} 
A complementary question is whether interference is organized by the annotated \textit{attributes} of the \textit{entities}, rather than by pairwise \textit{similarity}. This lens is particularly relevant for analyzing broader impacts: an \textit{attribute} along which interference propagates systematically acts as a ``transmission wire'', and whether such propagation is acceptable is an ethical rather than a technical judgment.

\begin{figure}[tb]
  \centering
  \includegraphics[width=0.7\linewidth]{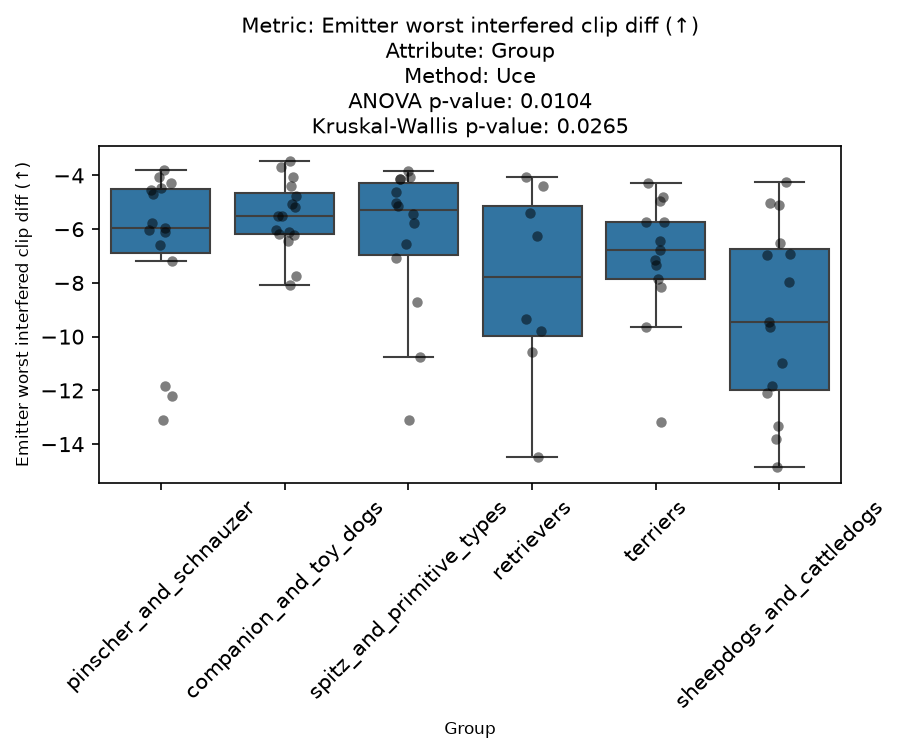}
  \caption{Result obtained from computing the RT \textit{SignificantRelationshipCategorical} for the parameters $m=$\textit{Stable Diffusion v1.4}, $t=$\textit{breeds}, $u=$\textit{UCE}, $m_e=$\textit{emitterWorstInterferedClipDiff}, and $a=$\textit{group}. The statistical tests indicate that not all groups have the same mean (tested by ANOVA) nor the same distribution (Kruskal-Wallis), without indicating which pair drives the difference. Each points represents one \textit{entity} (thus, there are 100 points in total).}
  \label{fig:sig_breeds_group}
\end{figure}

\begin{figure}[htb]
  \centering
  \includegraphics[width=0.7\linewidth]{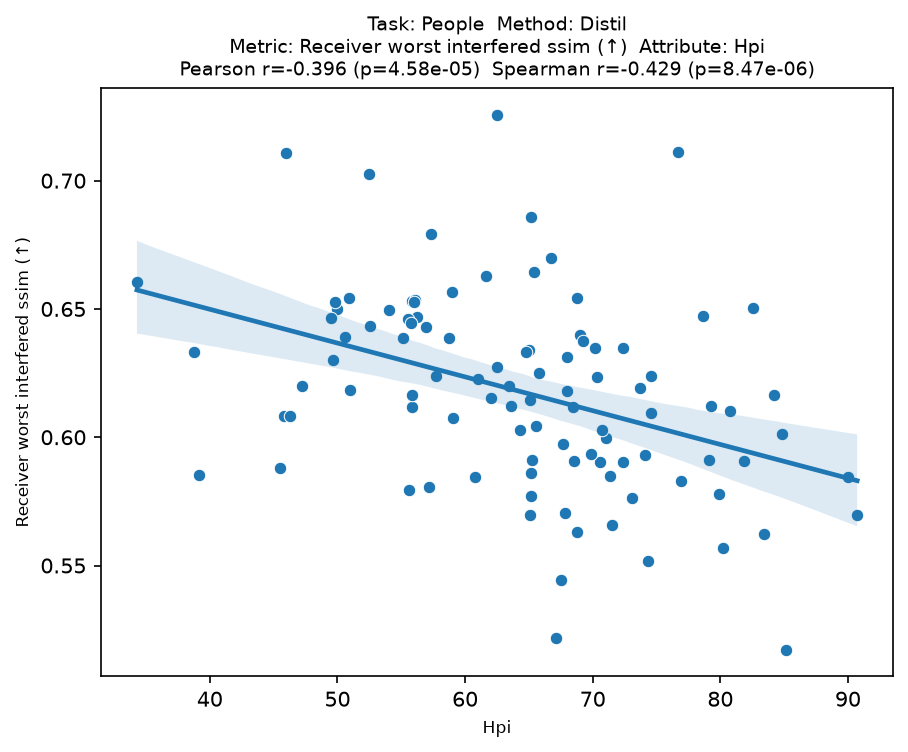}
\caption{\textit{SignificantRelationshipNumerical} for \textit{Stable Diffusion v1.4}, \textit{people}, \textit{SPARE}, \textit{receiverWorstInterferedSSIM}, and \textit{HPI}. The statistical tests demonstrate that \textit{HPI} is correlated (Pearson test) and has a monotonic relationship (Spearman test) with \textit{receiverWorstInterferedSSIM}.
}
\label{fig:hpi_clip}
\end{figure}

Using \textit{SignificantRelationshipCategorical}, we observe that 
\textit{\ac{uce}} causes significantly different amounts of \textit{EmitterWorstInterferedClipDiff} $m_e$ interference (a \textit{MetricInterferencePerEntity}, as defined in \cref{ap:metrics}) across \textit{group} when unlearning \textit{breeds}. More specifically, \textit{Sheepdogs and cattledogs} cause the most interference, while \textit{Spitz and primitive types} cause the least, as shown in \cref{fig:sig_breeds_group}. 
Similarly, by invoking \textit{SignificantRelationshipNumerical} with parameters $m=SD$, $t=people$, $u=SPARE$, $m_e=$\textit{Receiver\-Worst\-Interfered\-SSIM}, and $a=HPI$ (Historical Popularity Index), we observe that more famous \textit{entities} tend to receive significantly more interference (\cref{fig:hpi_clip}).
Naturally, such differences could be related to aspects not measured by the I-CARE methodology, such as the pose variety of different groups or biases in how CLIP \cite{Radford2021Clip} measures semantic alignment.

\begin{figure}[htb]
  \centering
  \includegraphics[width=1\linewidth]{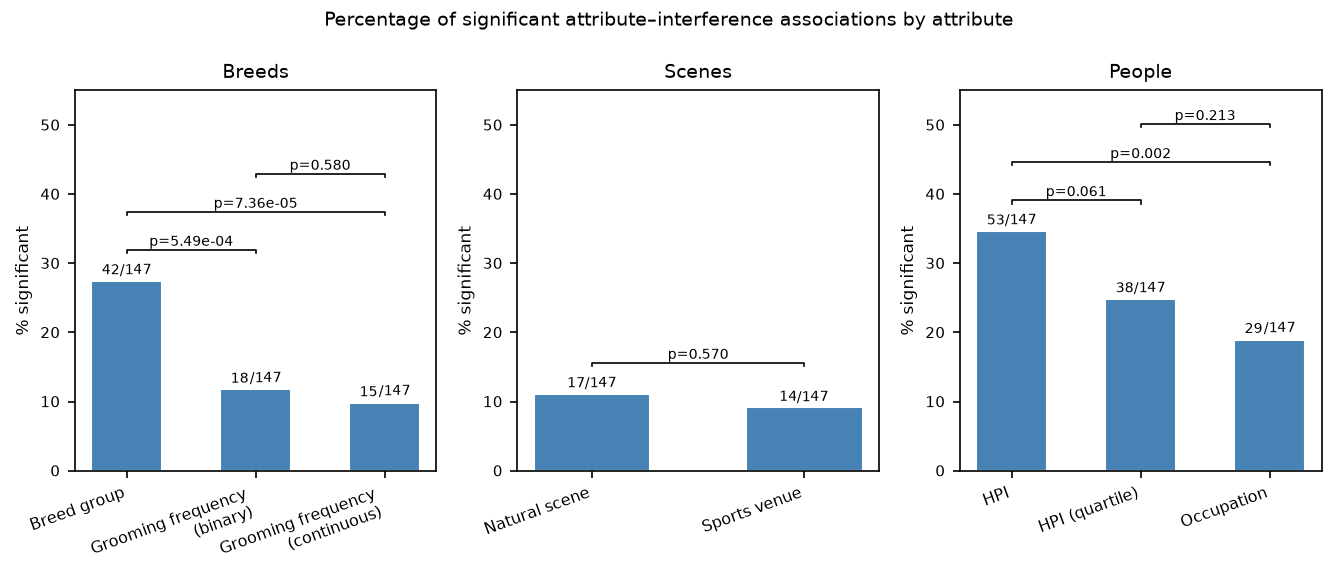}
    \caption{\textit{AttributesOfInterest} ranked by the percentage of significant relationships they show across all ($m_e$ $\times$ \textit{unlearningMethod}) combinations, aggregated per \textit{task}.}
    \label{fig:sig_by_attribute}
\end{figure}

\begin{figure}[htb]
  \centering
  \includegraphics[width=0.6\linewidth]{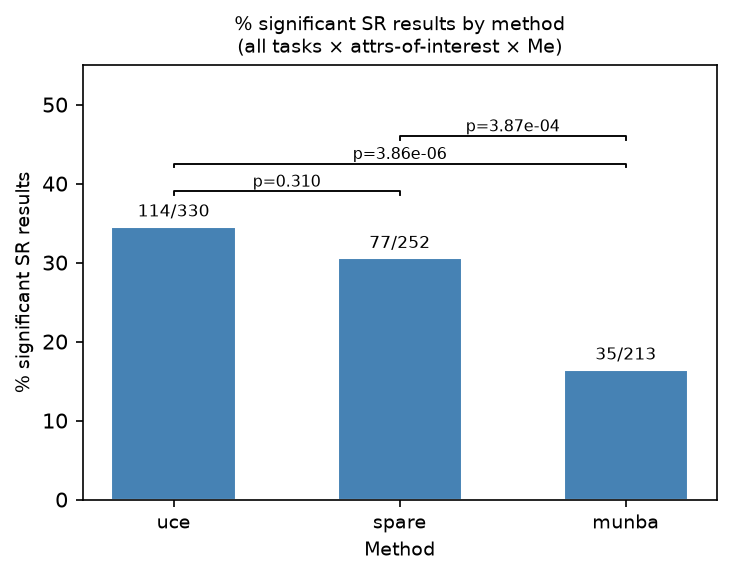}
    \caption{\textit{Methods} ranked by \textit{CountSignificantRelationship}, that is, the number of significant ($m_e$ $\times$ \textit{attribute}) relationships each \textit{method} shows across its sessions.}
    \label{fig:sig_by_method}
\end{figure}

Beyond those specific combinations of parameters, \cref{fig:sig_by_attribute} 
ranks the \textit{attributesOfInterest} by the percentage of significant relationships they exhibit across the 147 possible combinations of 49 \textit{metricInterferencePerEntities} and 3 \textit{unlearningMethods}, aggregated per \textit{task}.
If the goal is to compare \textit{methods} rather than \textit{attributes}, the RT \textit{CountSignificantRelationship} counts, per \textit{method}, how many (\textit{attribute} $\times$ $m_e$) relationships reach significance across its unlearning sessions; \cref{fig:sig_by_method} reports this ranking. \textit{\ac{uce}} and \textit{\ac{spare}} exhibit more such interference channels than \textit{\ac{munba}}.

\begin{figure}[tb]
  \centering
  \includegraphics[width=1\linewidth]{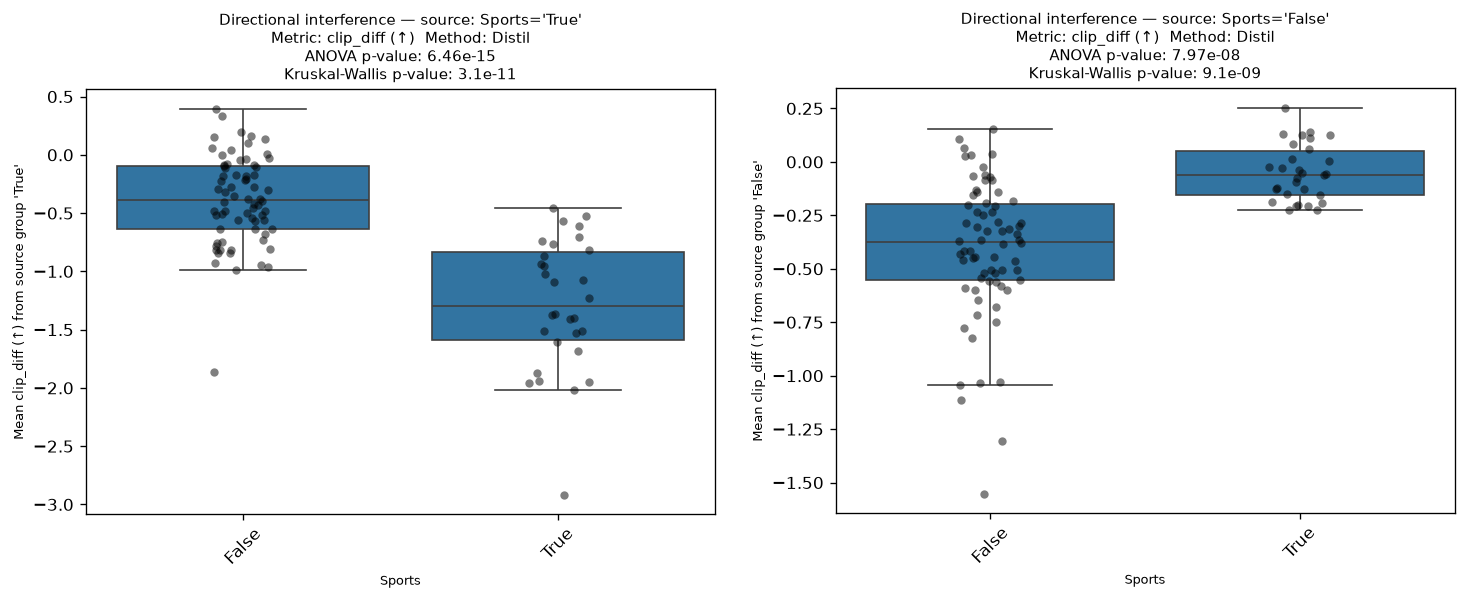}
    \caption{Directional extension of \textit{SignificantRelationshipCategorical} for the binary \textit{sport attribute} in \textit{scenes} (\textit{SPARE}, \textit{$\Delta$Clip}). Left: emitters restricted to \textit{sport scenes}; right: emitters restricted to \textit{non-sport scenes}. In both, interference concentrates within the emitter's own group.}
    \label{fig:sig_dir_sports}
\end{figure}

A complementary analysis uses the directional (flow) extension of \textit{SignificantRelationshipCategorical} (see \cref{ap:rt_relationship}), which conditions on a fixed emitter group and tests how the interference it emits is distributed across receiver groups.
\cref{fig:sig_dir_sports} shows the two invocations over the binary \textit{attribute} \textit{sport} of the \textit{scenes} \textit{task} (under \textit{\ac{spare}} and measured by \textit{$\Delta$Clip}): emitters restricted to \textit{sport scenes} (left) and to \textit{non-sport scenes} (right). In both, interference concentrates significantly within the emitter's own group, so \textit{sport scenes} interfere most with other \textit{sport scenes}, and likewise for \textit{non-sport}.

\FloatBarrier
\subsection{Relationships between similarities and interference} 
\begin{figure}[tb]
  \centering
  \includegraphics[width=0.9\linewidth]{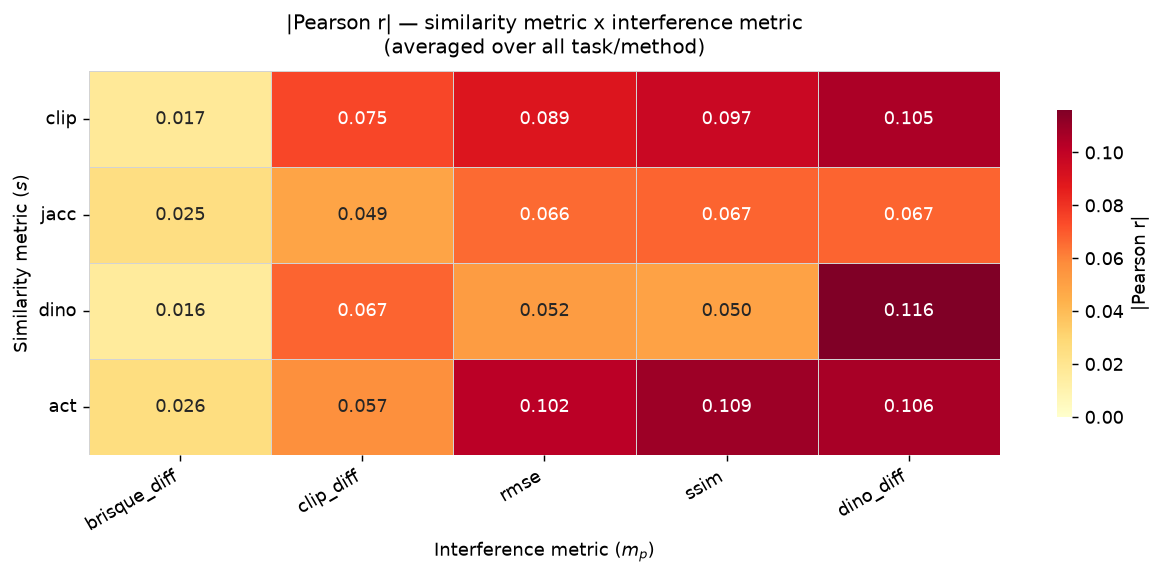}
    \caption{Mean Pearson correlations obtained with RT \textit{MetricSimilarityAlignment} across all combinations of \textit{MetricInterferencePerEntityPair} and \textit{SimilarityBetweenEntities}.}
    \label{fig:msa_full_heatmap_sim_mp_abs}
\end{figure}

\begin{figure}[tb]
  \centering
  \includegraphics[width=0.8\linewidth]{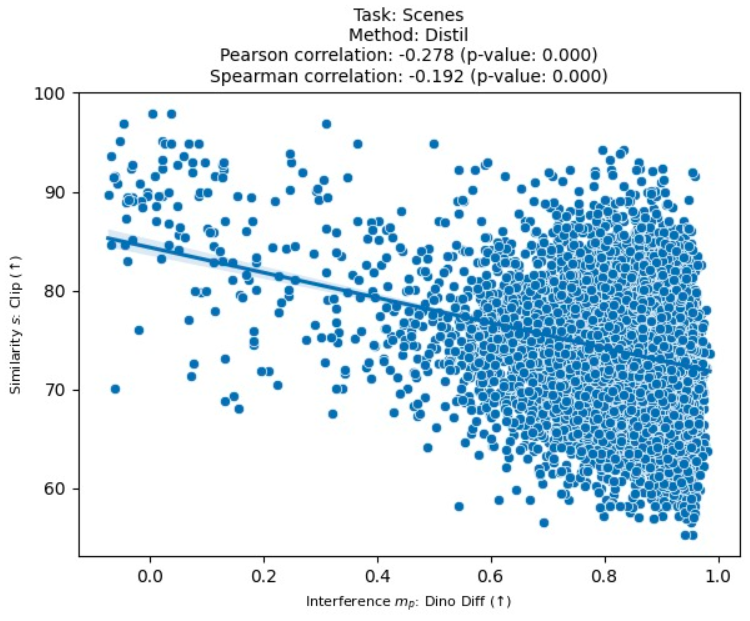}
    \caption{\textit{MetricSimilarityAlignment} for $m_p = \Delta$\textit{Dino} and $s = $\textit{Clip} under \textit{SPARE} in \textit{scenes}. Each point correspond to one pair of an unlearned \textit{entity} and an interfered \textit{entity}, giving $100\times 99=9900$ points.}
    \label{fig:results_temp_1}
\end{figure}

The RT \textit{MetricSimilarityAlignment} can be used to analyze whether the \textit{MetricInterferencePerEntityPair} $m_p$ can be estimated by a \textit{SimilarityBetweenEntities} $s$. No single $m_p$ was well predicted by any $s$ across all \textit{tasks} and \textit{unlearningMethods} (\cref{fig:msa_full_heatmap_sim_mp_abs}).
A reliable predictor would have been desirable, since it would let interference be estimated cheaply for new \textit{entities}; its absence indicates that more advanced methods are needed to predict the occurrence of interferences in a computationally efficient manner.
The multivariable and non-linear predictive regression proposed in \cref{ap:prediction} was implemented, but its results were no better than the best performing \textit{similarity}, thus the results will not be reported.
Restricting the analysis to specific \textit{tasks} and \textit{unlearningMethods}, interferences are slightly more predictable, such as the relationship between $m_p=\Delta Dino$ and $s=Clip$ for the \textit{\ac{spare}} method in \textit{scenes} (\cref{fig:results_temp_1}), which has Pearson correlation of $-0.278$.

\begin{figure}[tb]
  \centering
  \includegraphics[width=1\linewidth]{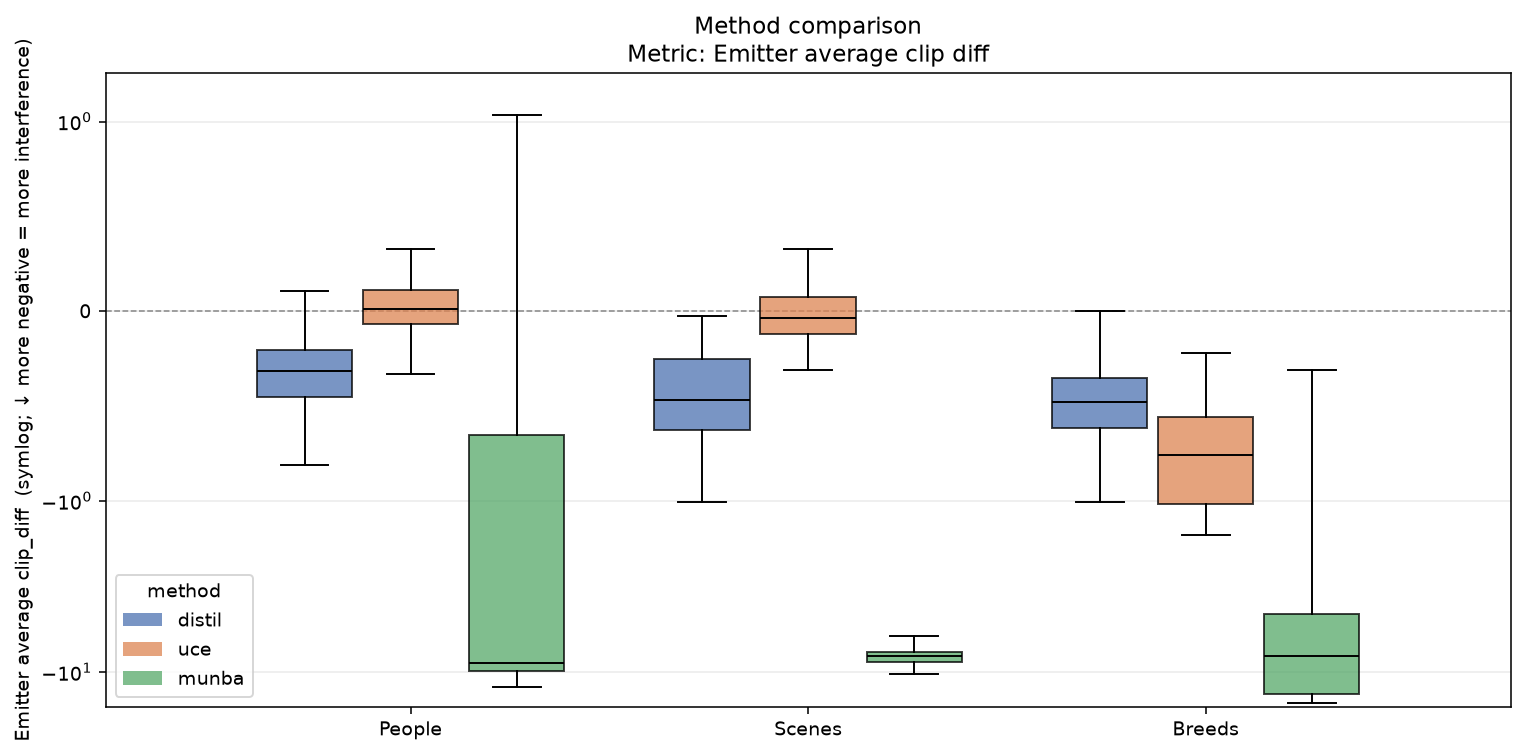}
    \caption{Result from \textit{MethodComparisonByMetricEntity} across all \textit{methods} and all \textit{tasks} of the feasibility demonstration, for the $m_e$ \textit{emitterAverageClipDiff} (one of the most meaningful and easy to interpret interference metrics, but the same analysis can be done for each one of the 49 available $m_e$). Logarithmic scale is used in the vertical axis due to the large difference between \textit{\ac{munba}} and the other two \textit{methods}.}
    \label{fig:MethodComparisonByMetricEntity}
\end{figure}

\begin{figure}[tb]
  \centering
  \includegraphics[width=0.8\linewidth]{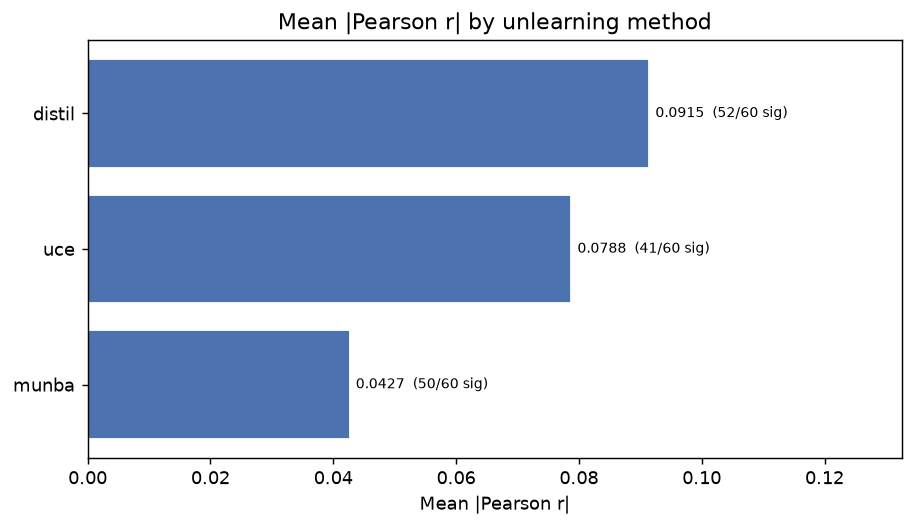}
    \caption{Aggregation by \textit{method} of the results obtained from \textit{MetricSimilarityAlignment}. Each bar aggregates results from 5 \textit{InterferencePerEntityPair} metrics × 4 \textit{similarity} metrics × 3 \textit{tasks}, totaling 60 executions of the RT.}
    \label{fig:msa_full_groupby_method}
\end{figure}

Intermediate levels of aggregation are naturally possible, such as by \textit{task} or by \textit{MetricInterferencePerEntityPair}, depending on what aspect of the unlearning process the researcher is interested in analyzing. In \cref{fig:msa_full_groupby_method} we analyze which \textit{unlearningMethods} have more easily predictable interferences, in which \textit{SPARE} shows a small advantage.
Beyond the predictability of its interferences, when comparing \textit{unlearningMethods} it is important to analyze the total amount of caused interference, which can be done with the RT \textit{MethodComparisonByMetricEntity}. 
In \cref{fig:MethodComparisonByMetricEntity}, measured by the $m_e$ \textit{emitterAverageClipDiff}, we observe that \textit{\ac{uce}} causes less interference than \textit{SPARE}, and \textit{\ac{munba}} causes substantially more than both.

\FloatBarrier

\begin{figure}[tb]
  \centering
  \includegraphics[width=0.7\linewidth]{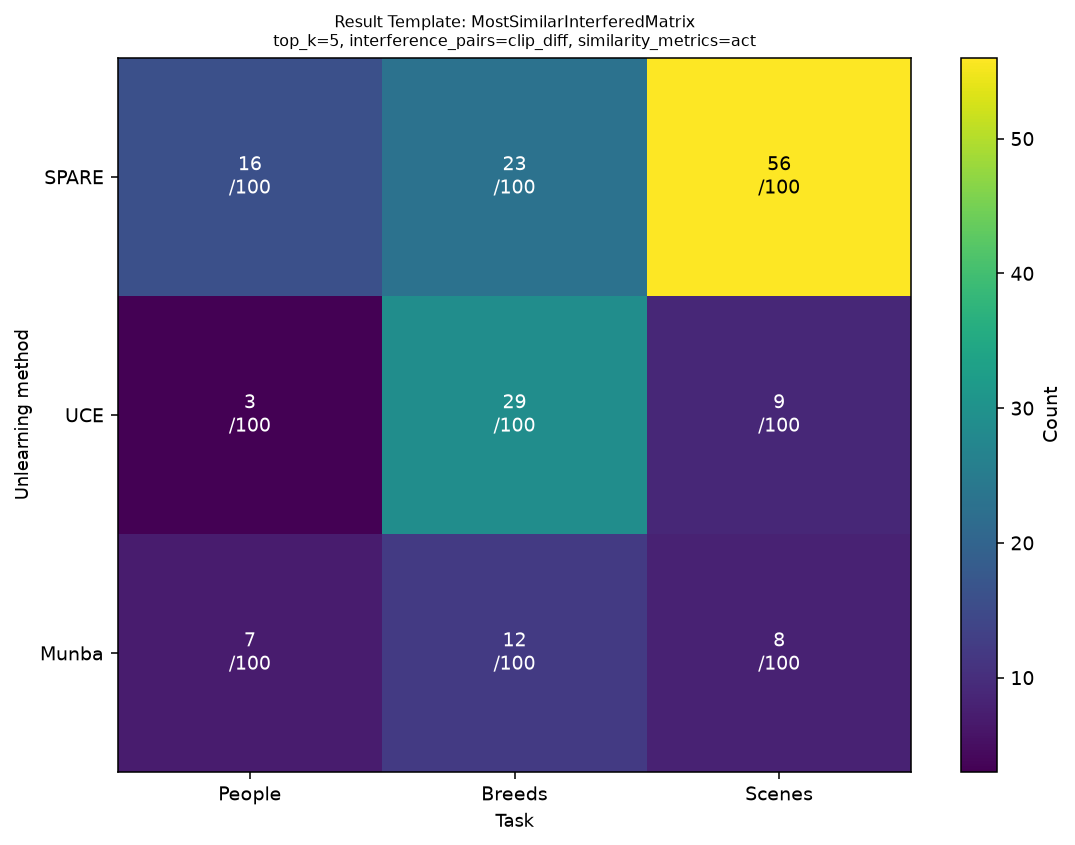}
    \caption{RT \textit{MostSimilarInterferedMatrix} counting, for every combination of $t$ and $u$, how many of the 100 $e_1$ have the property that the $e_2$ most similar to $e_1$ (as measured by $s=act$, the \textit{similarity} function with strongest predictive effect) is among the 5 $e_i$ most interfered by the unlearning of $e_1$ (as measured by $m_p=\Delta$\textit{Clip}).}
    \label{fig:msa_grid_clipdiff_act_top5}
\end{figure}

Instead of looking at overall correlations, another possibility of analysis is looking only at the most similar pairs. That is, when unlearning one \textit{entity}, what happens to the other \textit{entity} that is the most similar to it. Such questions can be answered by the RT \textit{MostSimilarInterferedMatrix}, an aggregation of the same \textit{MetricSimilarityAlignment} and \textit{InterferencePerEntityPair} used by \textit{MetricSimilarityAlignment}, described in \cref{ap:prediction_most_similar}, that focuses only on the single most-similar pair.
In \cref{fig:msa_grid_clipdiff_act_top5} we observe an execution of that RT, in which \textit{\ac{spare}} clearly has more predictable interferences. For \textit{scenes}, the most similar \textit{entity} was among the 5 most interfered in 56 out of the 100 \textit{entities}, more than ten times higher than random chance, which would be about 5 occurrences. This predictability is lower for the other tasks, and overall just above chance for unlearning sessions performed with \textit{\ac{munba}}.

\FloatBarrier
\subsection{Interference and similarity matrices}
\begin{figure}[tb]
\centering
\includegraphics[width=0.8\linewidth]{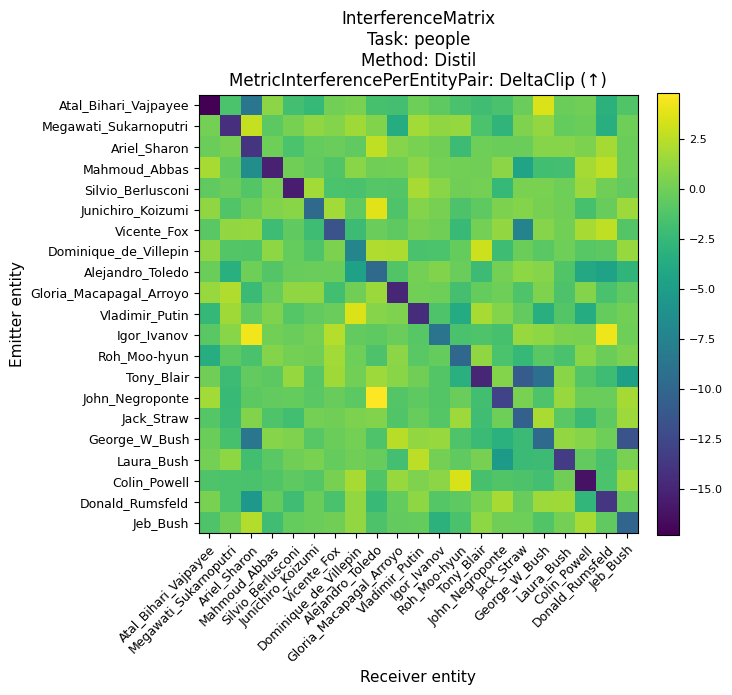}
\caption{\textit{InterferenceMatrix}, filtered, sorted, and zoomed to an area of interest.}
\label{fig:matrix}
\end{figure}

\cref{fig:matrix} shows a subset of the \textit{InterferenceMatrix} for the task \textit{people}, computed using the \textit{MetricInterferencePerEntityPair} $\Delta Clip$ for the method \textit{\ac{spare}}; the indices were sorted by nationality (alphabetically) and then by popularity (decreasing), and a sequential subset of \textit{entities} was manually selected. As expected, an \textit{entity} is usually the most affected by its own unlearning (hence the dark diagonal). However, no large-scale organization is distinguished, and subtle interferences (such as between \textit{George W. Bush} and \textit{Tony Blair}, respectively former President of the United States and Prime Minister of the United Kingdom during a largely overlapping period of time) are not captured by the sorting attributes.

\begin{figure}[tb]
\centering
\includegraphics[width=0.8\linewidth]{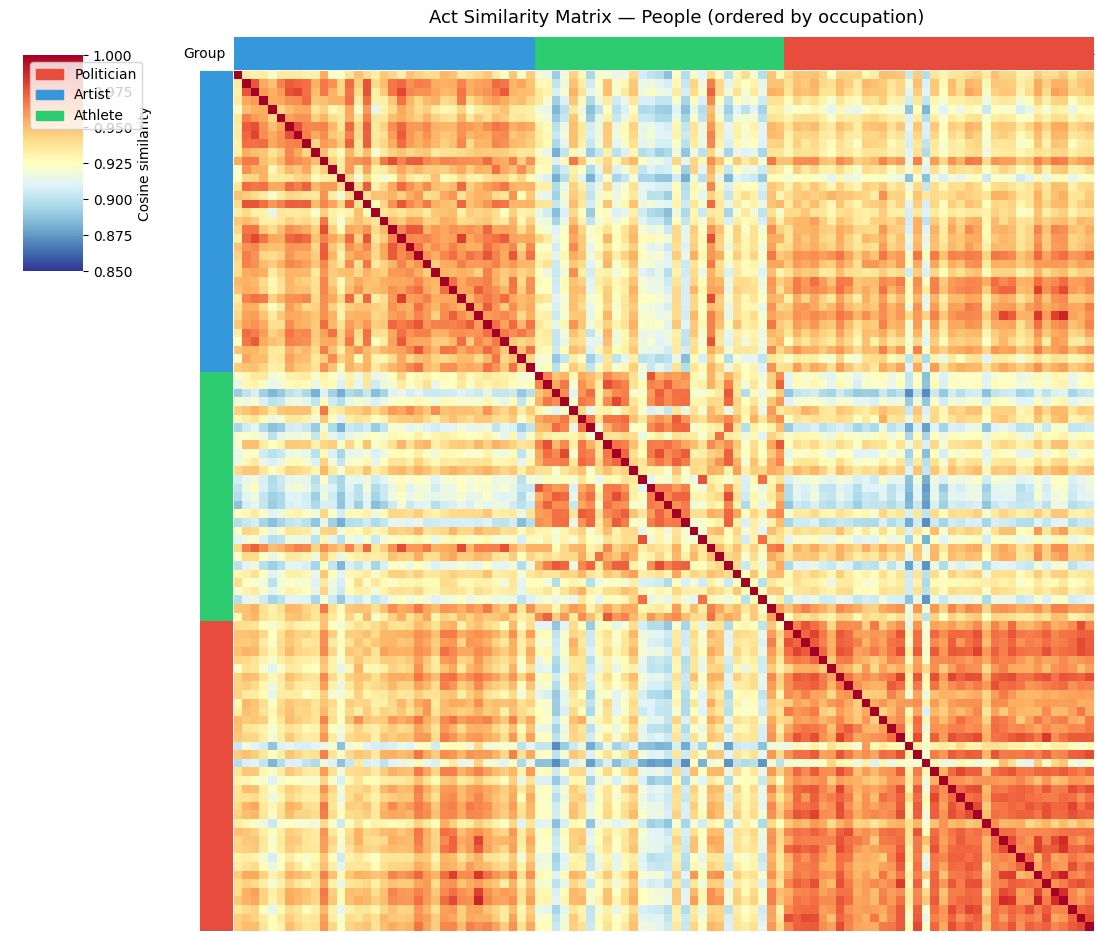}
\caption{RT \textit{SimilarityMatrix} for the \textit{task people}, with the rows sorted by the \textit{attribute occupation} (which can assume the values \textit{Politician}, \textit{Artist}, or \textit{Athlete}).}
\label{fig:sim_matrix_act}
\end{figure}

\cref{fig:sim_matrix_act} shows the \textit{SimilarityMatrix} for the same
\textit{task}, computed with the \textit{similarity} \textit{act}, this time
with the \textit{entities} sorted by \textit{occupation} (\textit{Artist}, \textit{Athlete},
and \textit{Politician}). Unlike the \textit{InterferenceMatrix} of
\cref{fig:matrix}, here the sorting \textit{attribute} organizes the matrix at a
large scale: a block structure is visible along the diagonal, with \textit{entities} of the same
\textit{occupation} tending to be more similar to one another than to \textit{entities} of other
occupations. The effect is uneven across groups: the \textit{Politician} block is the
most cohesive (a dense high-similarity cluster in the lower-right corner), the
\textit{Artist} block is moderately cohesive, and the \textit{Athlete} block is the
least similar, both internally and toward the other groups, appearing as comparatively
cooler bands. A small number of \textit{entities} are dissimilar to almost all others, visible as
isolated light-colored rows and columns. We emphasize that this organization concerns the
similarity space alone, and we do not claim it translates into the interference space.

\FloatBarrier
\subsection{ImplicitAssociationTest} 
\begin{figure}[tb]
  \centering
  \includegraphics[width=1\linewidth]{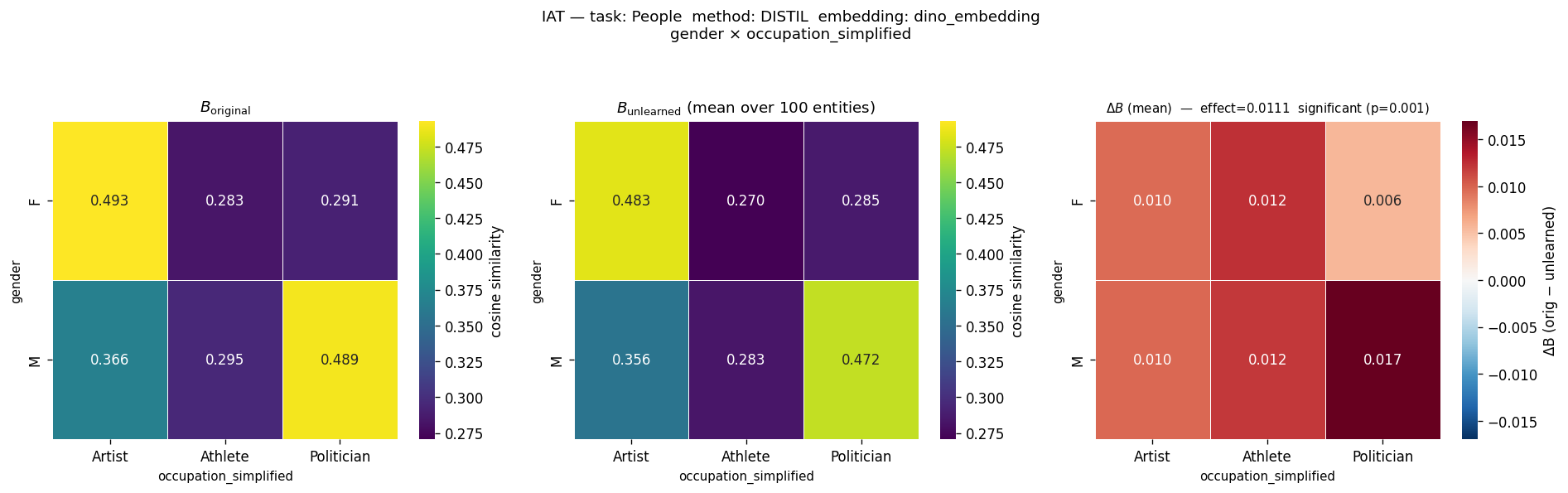}
    \caption{RT \textit{ImplicitAssociationTest} between \textit{gender} and \textit{occupation} for the \textit{people} task with \textit{\ac{spare}} (using \textit{dino} embeddings): $B$ before and after unlearning, and the shift $\Delta B = B_{\text{original}}-B_{\text{unlearned}}$ (positive = weakened association).}
    
  \label{fig:iat_gender}
\end{figure}

\begin{figure}[tb]
  \centering
  \includegraphics[width=1\linewidth]{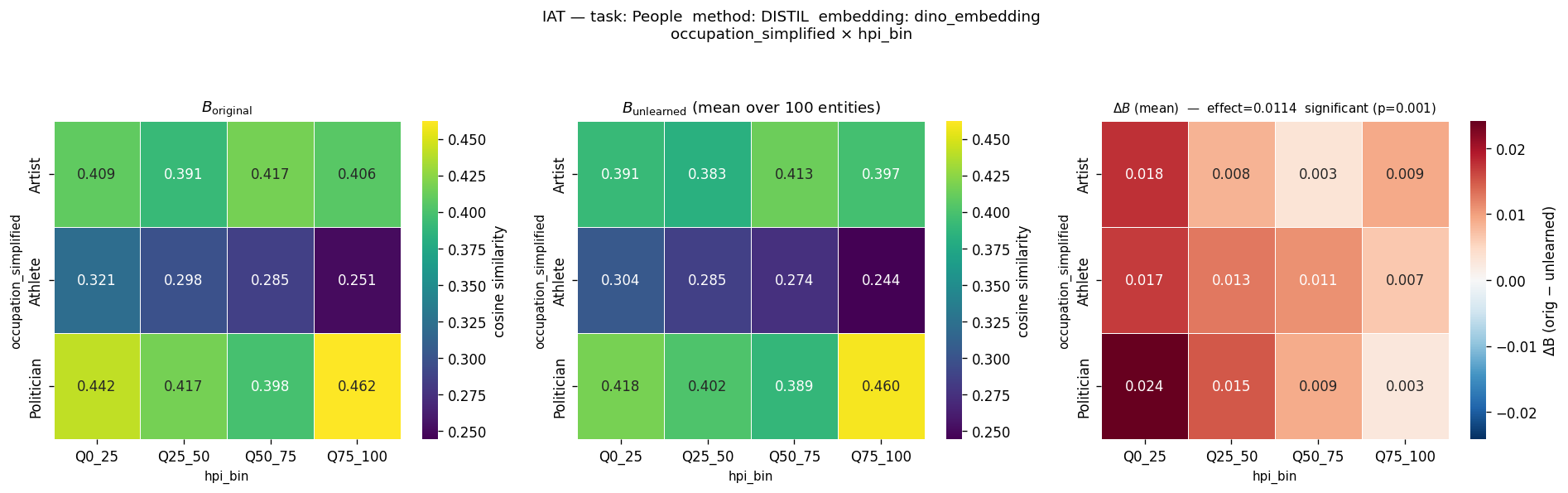}
    \caption{RT \textit{ImplicitAssociationTest} between \textit{occupation} and \textit{HPI bin} for the \textit{people} task with \textit{\ac{spare}}, using \textit{Dino embeddings}. The shift $\Delta B$ shows that associations involving low-\textit{HPI} (low popularity) \textit{entities} weakened more than those involving high-\textit{HPI} \textit{entities}.}
  \label{fig:iat_hpi}
\end{figure}

By using the RT \textit{ImplicitAssociationTest}, we observe that the original \textit{Stable Diffusion model} contains several biased associations among attributes, whose discussion is beyond the scope of this work. For our purposes, the relevant analysis is how those associations change during unlearning. Across all \textit{tasks} and all \textit{unlearning methods}, associations tended to weaken, consistent with the interpretation that unlearning harms the overall knowledge of the \textit{model} (including knowledge about biases).
Moreover, the decline in association was more intense in stereotypical combinations (such as \textit{male} and \textit{politician}, as shown in \cref{fig:iat_gender} for the \textit{\ac{spare}} method, in which the effect was especially pronounced), and low popularity (measured by \textit{HPI bin}) \textit{entities} were more changed than high popularity ones (\cref{fig:iat_hpi}).

\FloatBarrier
\subsection{Graph interpretation and minimum cut} 
A motivation for the \ac{RT} \textit{MinimumCutInterference} is that interference between two \textit{entities} may not stem from their direct \textit{similarity}, but from an indirect ``propagation path'' through intermediate \textit{entities}. 
For example, if two \textit{entities} $e_1$ and $e_3$ are both strongly tied to a common \textit{entity} $e_2$ (e.g., two people married to the same person), unlearning $e_1$ may end up harming $e_3$ even when $e_1$ and $e_3$ are not themselves similar. 

\begin{figure}[tb]
  \centering
  \includegraphics[width=1\linewidth]{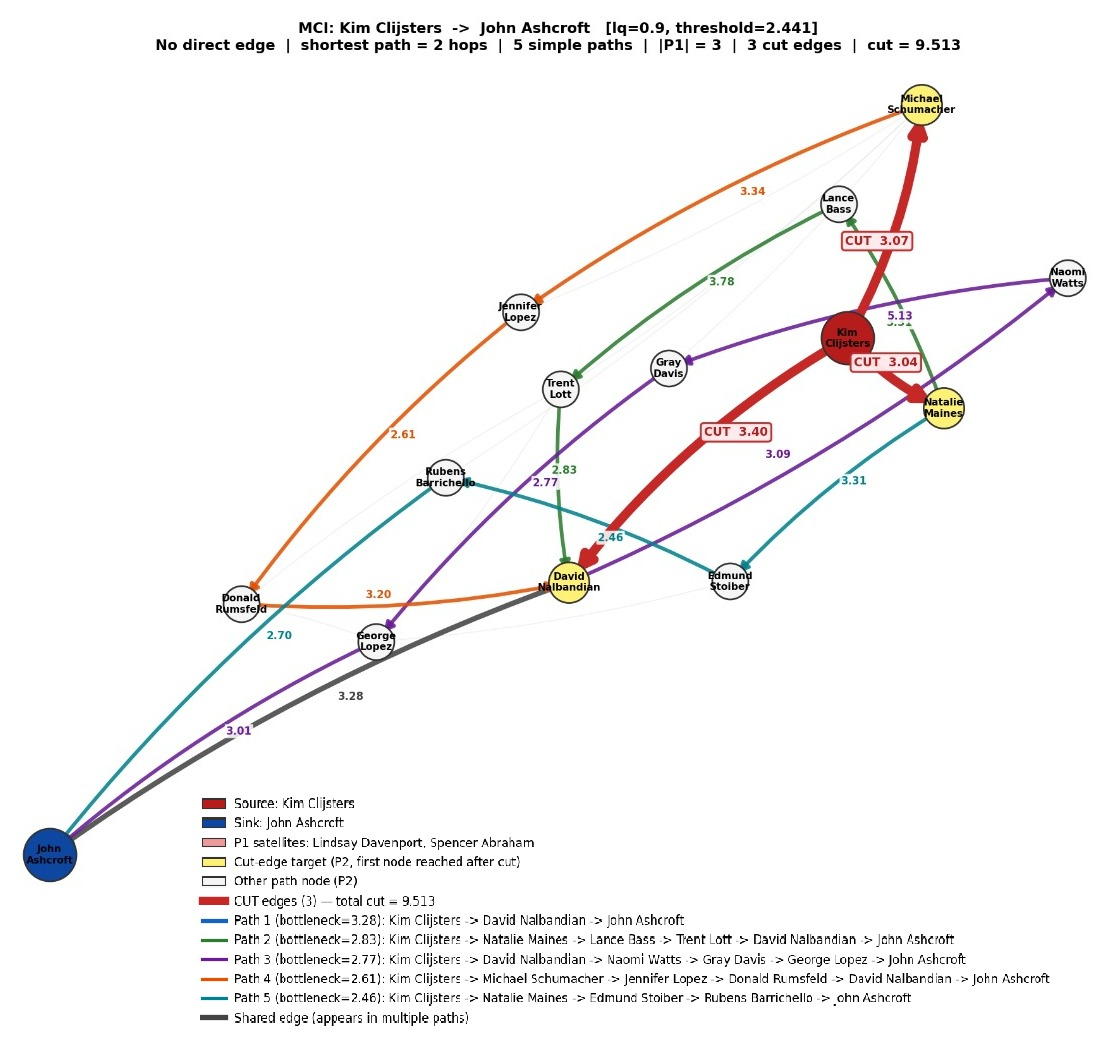}
  \caption{RT \textit{MinimumCutInterference} for source \textit{Kim Clijsters} and sink \textit{John Ashcroft} (top-10\% strongest edges retained): with no direct edge between them, the minimum cut is the three edges from the source-side partition to \textit{David Nalbandian}, \textit{Natalie Maines}, and \textit{Michael Schumacher}.}

  \label{fig:graph}
\end{figure}

Since such relationships are hard to express through \textit{attributes}, we computed several interference graphs with the \ac{RT} \textit{MinimumCutInterference}, varying the interference threshold, the \textit{metricInterferencePerEntityPair} used to measure interference, and the \textit{unlearningMethod}. 
After mining the resulting ``2-hop interference pairs'' (pairs of \textit{entities} connected only through a single intermediate \textit{entity}) and manually searching for meaningful combinations, we were not able to find clear interpretable ``stories'' in the data.
One example of the commonly found patterns is shown in \cref{fig:graph}, for investigating why unlearning \textit{Kim Clijsters} (a female Belgian tennis player, born in 1983) has any effect at all on \textit{John Ashcroft} (American politician, born in 1942).
Pruning the graph retaining only the strongest 10\% of edges, there is no direct connection between the two entities (so their interference is not strong, albeit not zero). The minimum cut contains 3 edges going from \textit{Clijsters} (that is, had we eliminated the interference in those 3 receiver entities, there would be no connection to \textit{Ashcroft}).

Under closer inspection, and with a considerable dose of subjective interpretation, it seems that all 3 intermediate \textit{entities} share properties with both \textit{Clijsters} and \textit{Ashcroft}. 
\textit{Nalbandian} is a male Argentinian tennis player, born in 1982 and at his professional peak at approximately the same years as \textit{Clijsters}; even though he lived his entire life in Argentina, he spent time in the U.S. for tournaments and training during the time \textit{Ashcroft} was Attorney General (2001–2005); furthermore, Nalbandian is an Armenian surname, and there are several American public figures with that surname, such as John Baylor Nalbandian (judge, not in our dataset).
\textit{Schumacher}, a male German F1 driver, was born in 1969, thus an intermediate age between \textit{Clijsters} and \textit{Ashcroft}. 
\textit{Maines}, a female American singer born in 1974, toured extensively in Europe; her most famous concert was, arguably, at Shepherd's Bush Empire in London, in 2003, the year in which 
\textit{Clijsters} was World Number 1 for the first time in her career. Beyond sharing the same nationality, \textit{Ashcroft} was an amateur singer and part of the band The Singing Senators, and a strong defender of the security and military U.S. activities that were strongly condemned by \textit{Maines}.
We do not claim that those relationships cause unlearning \textit{Clijsters} to harm \textit{Ashcroft}, nor that intervening in the unlearning process so as to increase the preservation of \textit{Nalbandian}, \textit{Schumacher}, and \textit{Maines} would eliminate the interference in \textit{Ashcroft}, but we do suggest those are research lines worth pursuing.

\FloatBarrier
\section{Effects of the equalization}
\begin{figure}[tb]
  \centering
  \includegraphics[width=1\linewidth]{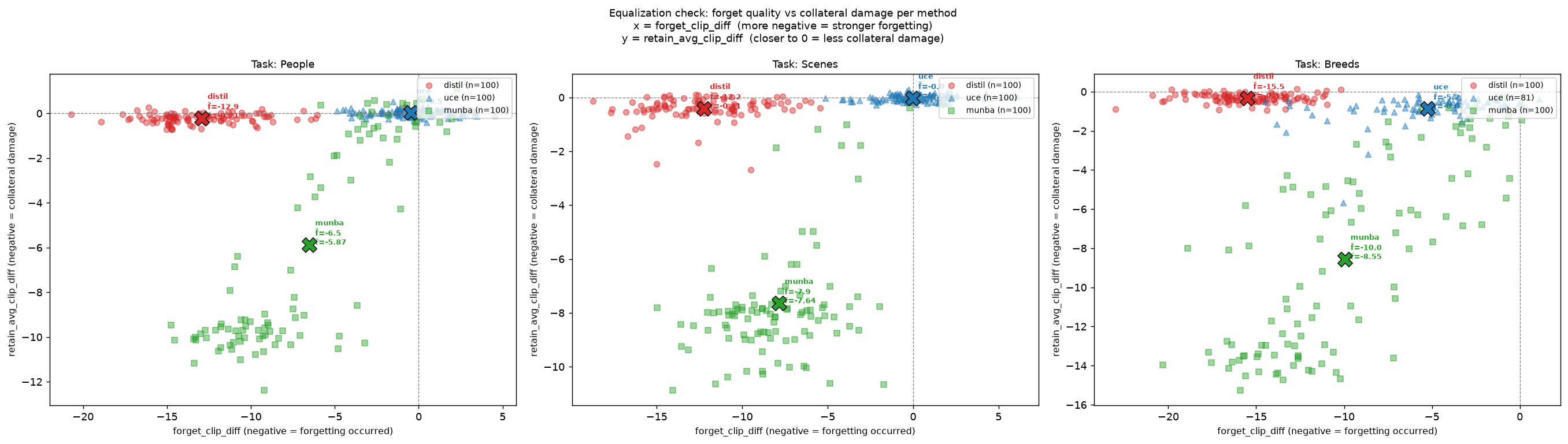}
  \caption{Comparison of forget quality (horizontal axis, more negative = stronger forgetting) against collateral damage (vertical axis, more negative = more interference) per \textit{method} and \textit{task}.}
  \label{fig:equalization}
\end{figure}

In \cref{fig:equalization} (obtained from a minor modification of the RT \textit{MetricMetricAlignment} to visualize all \textit{methods} together), each dot represents one unlearning session. The horizontal axis is the forget quality of the forgotten \textit{entity} (its own \textit{$\Delta$Clip}, how well the concept was erased) and the vertical axis is the interference on the 99 retain \textit{entities} (their aggregate \textit{$\Delta$Clip}, the aggregated collateral damage of forgetting).
While for \textit{people} and \textit{scenes} we attempted to equalize the hyperparameters, we observe that the results still greatly vary between the methods, albeit less than in the \textit{breeds task} (in which the hyperparameters were independently configured, without attempting to reach similar levels of forgetting). 
Across all \textit{tasks}, \textit{\ac{munba}} is the clear outlier: its sessions sit far below zero on the retain axis, forgetting at the cost of large collateral damage, whereas \textit{\ac{uce}} and \textit{\ac{spare}} remain close to zero. The per-\textit{method} means (marked with $\times$) stay dispersed along both axes, confirming that \textit{equalization} did not fully bring the three methods to a common operating point.

\begin{figure}[tb]
  \centering
  \includegraphics[width=1\linewidth]{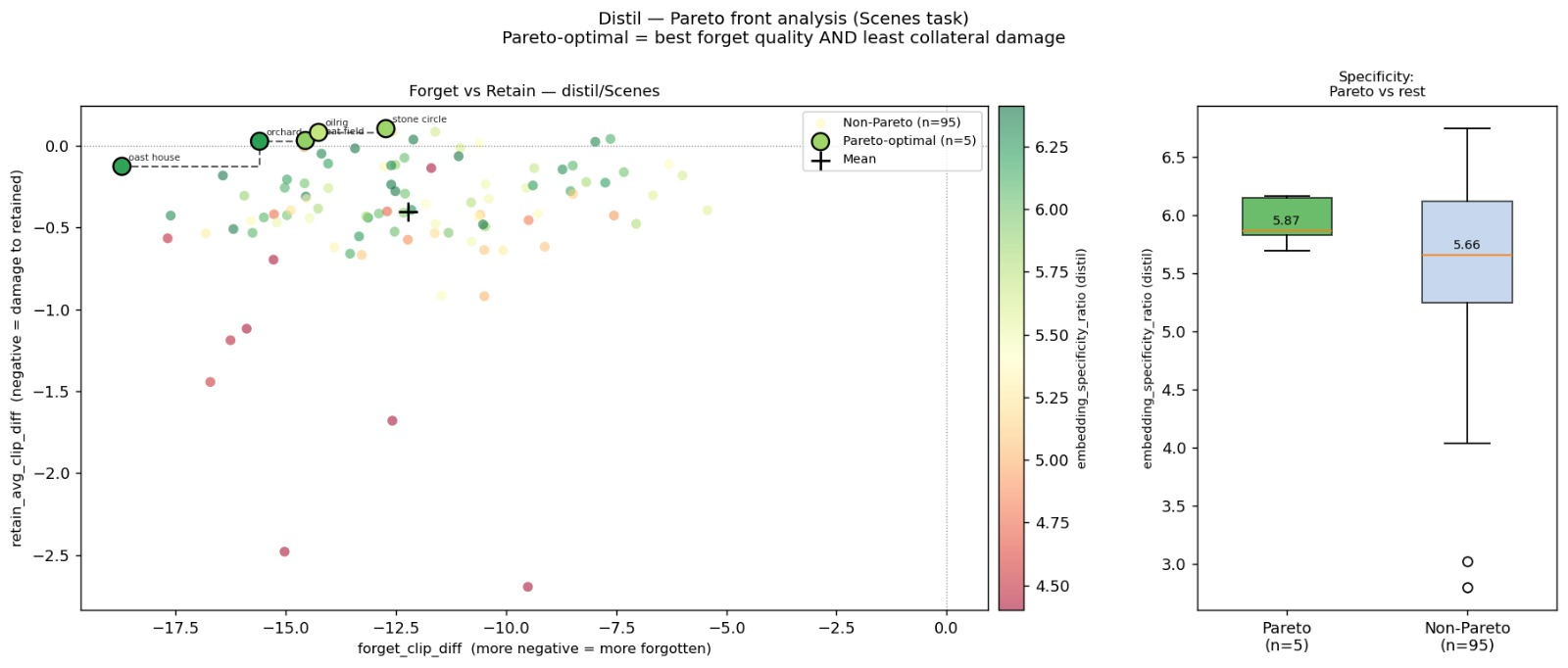}
  \caption{Forget-vs-retain trade-off for \textit{\ac{spare}} on \textit{scenes}, coloured by \textit{embeddingSpecificityRatio}, highlighting the few optimal sessions.}
  \label{fig:paretto}
\end{figure}

Zooming into a single \textit{method} and \textit{task} (\textit{\ac{spare}} on \textit{scenes}, \cref{fig:paretto}), the per-\textit{entity} sessions form a diffuse cloud rather than a sharp frontier, with only a handful on the Pareto front that jointly achieve strong forgetting and little interference. The same trade-off is visible through the fully unsupervised $m_e$ \textit{embeddingSpecificityRatio}: Pareto-optimal sessions tend toward higher values (median $5.87$ versus $5.66$ for the rest), though the distributions overlap substantially, so the metric is at best a weak indicator.

\FloatBarrier
\section{Effects of retain data} \label{sec:retain}
To assess the validity of the \textit{balancing} procedure, we performed a small-scale experiment in which one single \textit{entity} (\textit{football field}, from \textit{scenes}, chosen due to its high interference values) was unlearned with one \textit{method} (\textit{\ac{spare}}, chosen due to its clear interference patterns) using two different retain datasets: one containing only sport-related \textit{entities}, and another containing only \textit{entities} unrelated to sports. Interference was measured with \textit{$\Delta$Clip} in a reduced set of 10 \textit{entities}, chosen to include sport-related \textit{entities} and non-sport-related \textit{entities} representative of the variety found in the rest of the \textit{task} (\textit{basketball court indoor, basketball court outdoor, football field, tennis court indoor, tennis court outdoor, rainforest, ocean, abbey, oast house, pharmacy}).
The results, reported in detail in \cref{tab:results_retain}, clearly highlight the importance of the retain set for preserving \textit{entities} against interference: sport \textit{entities} are highly damaged by the unlearning of \textit{football field} when they are not present in the retain set, while just lightly harmed when they are present.
\begin{table}[tb]
\centering
\begin{tabular}{llrrr}
\hline
\textbf{Entity} & \textbf{Group} & \textbf{Standard} & \textbf{Sport} & \textbf{No sport} \\
\hline
Basketball Court Indoor  & sport     & -12.4099 & -0.0811  & -15.5222 \\
Basketball Court Outdoor & sport     & -12.9907 & 0.1392   & -15.8938 \\
\textbf{Football Field}           & \textbf{sport}     & \textbf{-9.9067}  & \textbf{-15.2208} & \textbf{-14.6320} \\
Tennis Court Indoor      & sport     & -6.3022  & -1.5764  & -15.5752 \\
Tennis Court Outdoor     & sport     & -7.1779  & -0.3616  & -14.6376 \\
Rainforest               & non-sport & -0.7492  & -0.9754  & -0.1574 \\
Ocean                    & non-sport & -0.2297  & -0.4408  & -0.2516 \\
Abbey                    & non-sport & 0.0680   & -1.2310  & -0.6414 \\
Oast House               & non-sport & -1.3410  & -1.3572  & -0.8500 \\
Pharmacy                 & non-sport & 3.2518   & -1.4439  & 0.2770 \\
\hline
\end{tabular}
\caption{$\Delta$Clip on 10 \textit{entities} after unlearning \textit{football field} (in bold) with \textit{\ac{spare}}, under three retain-set compositions: Standard (the default retain set used for comparison), Sport (only sport-related \textit{entities}), and No sport (only \textit{entities} unrelated to sports); lower (more negative) values indicate stronger interference.}
\label{tab:results_retain}
\end{table}

\FloatBarrier
\section{Effects of evaluation prompt}
\begin{figure}[tb]
  \centering
  \includegraphics[width=1\linewidth]{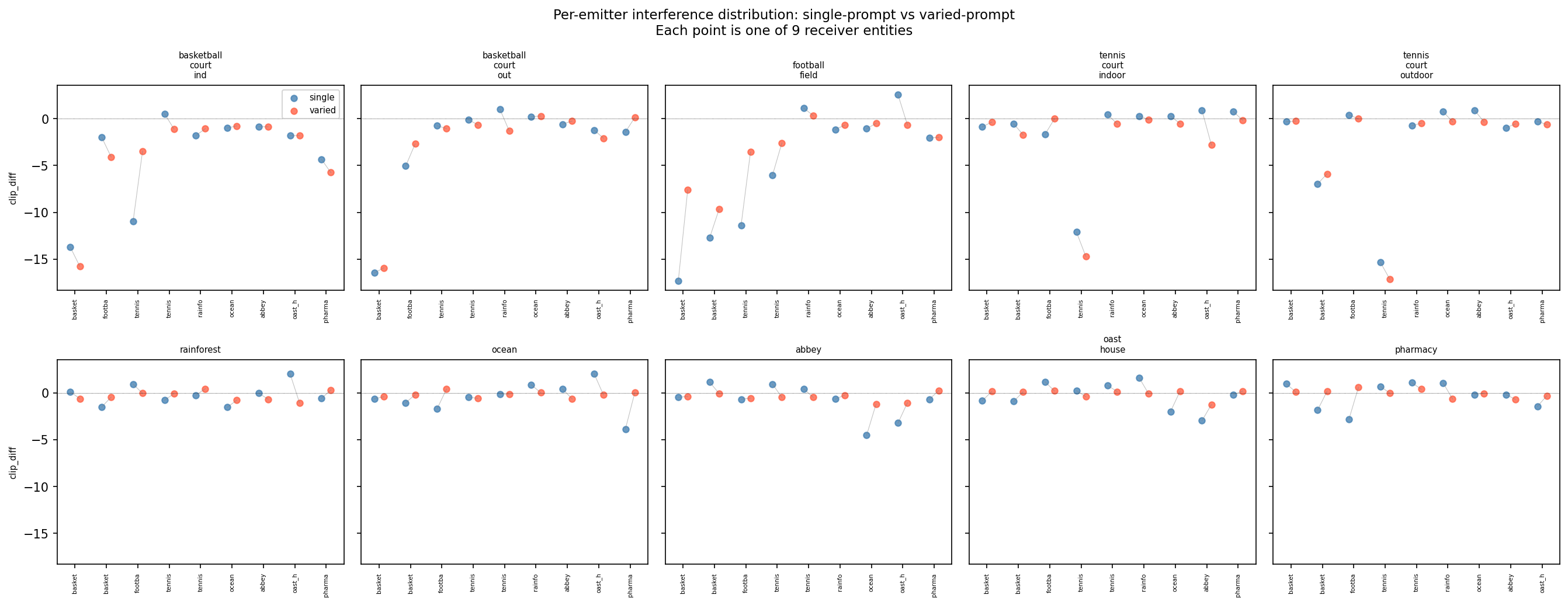}
  \caption{Per-emitter \textit{$\Delta$Clip} on the nine receiver \textit{entities} for each of the ten emitters, comparing single-prompt (blue) and varied-prompt (red) evaluation; the two largely overlap, and no systematic pattern is observed.}
  \label{fig:varied_all}
\end{figure}

\begin{figure}[tb]
  \centering
  \includegraphics[width=1\linewidth]{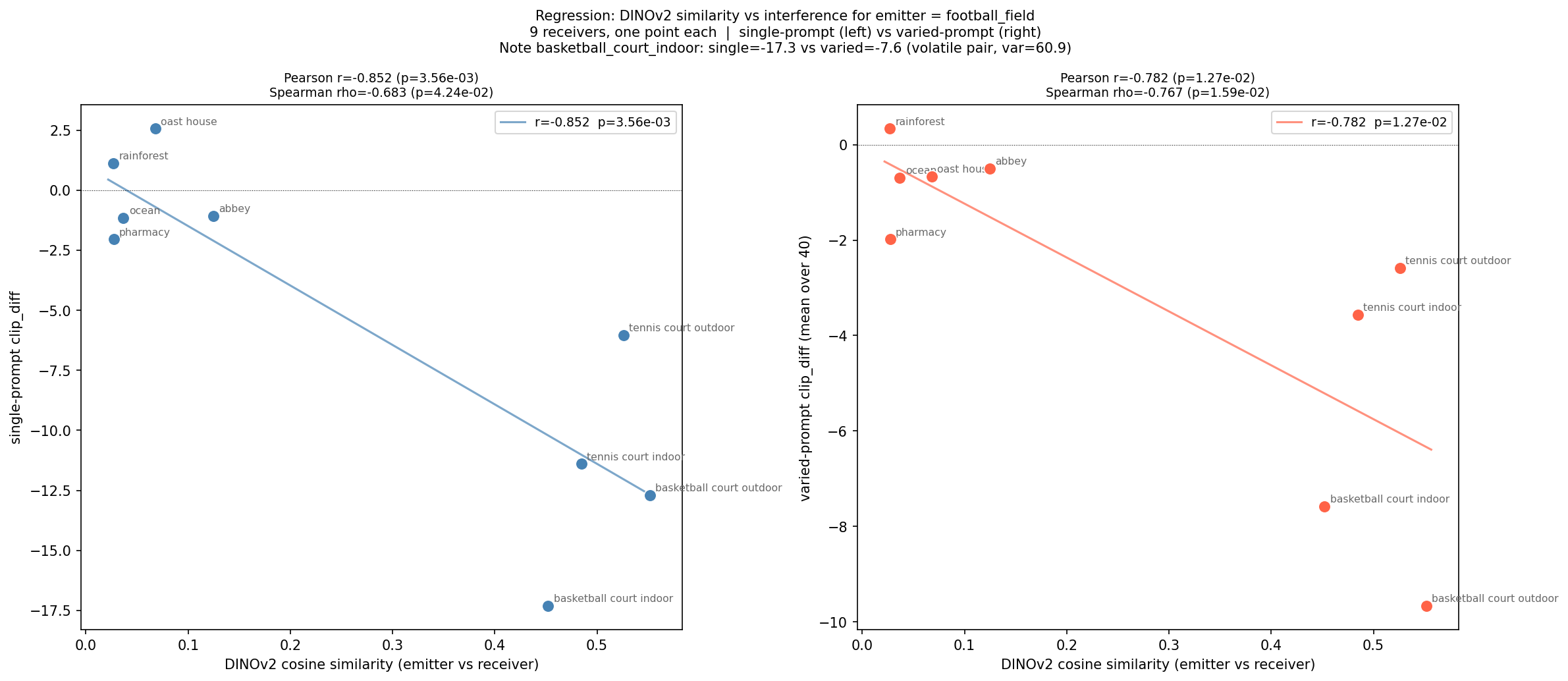}
  \caption{\textit{DINOv2} \textit{similarity} versus \textit{$\Delta$Clip} for the emitter \textit{football field}, under single-prompt (left) and varied-prompt (right) evaluation; the negative relationship persists in both, but the correlations are approximately the same (no statistical test comparing the two variations was performed due to the small sample size).}
  \label{fig:varied_one}
\end{figure}

\begin{figure}[tb]
  \centering
  \includegraphics[width=1\linewidth]{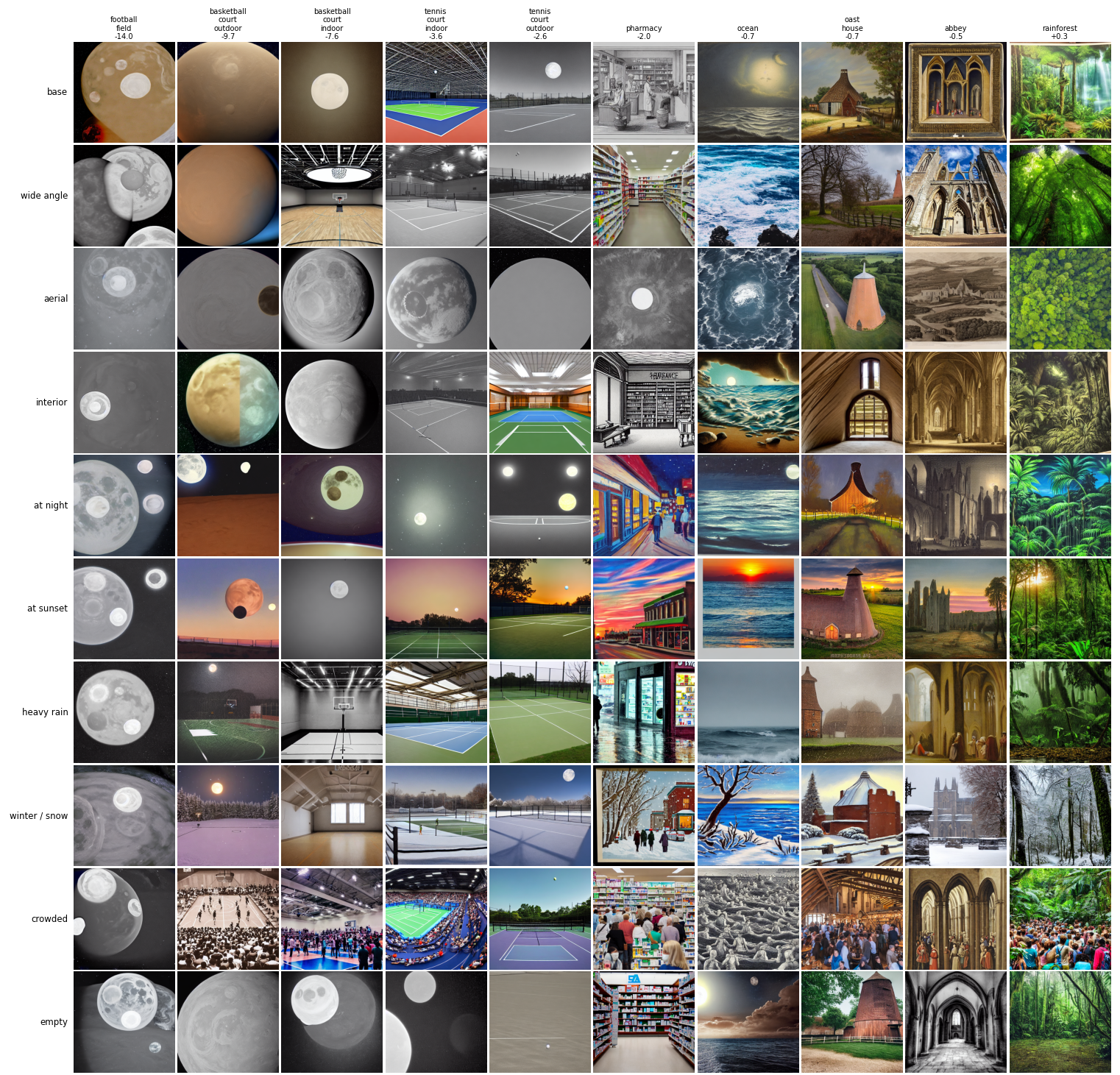}
  \caption{Unlearning session replacing \textit{football field} by \textit{the moon}. Columns ordered by \textit{$\Delta$Clip} interference, with the most interfered \textit{entities} displayed to the left.}
  \label{fig:varied_images}
\end{figure}

In the I-CARE methodology, the unlearned \textit{models} are assessed by generating images with different seeds (four seeds, in this feasibility demonstration) but always with the same prompt. In order to investigate the consequences of that design, we perform a small-scale experiment using instead 10 different prompts: ``an image of \{entity\}'', ``a wide angle view of \{entity\}'', ``an aerial view of \{entity\}'', ``a detailed interior view of \{entity\}'', ``\{entity\} at night'', ``\{entity\} at sunset'', ``\{entity\} during heavy rain'', ``\{entity\} in winter with snow'', ``a crowded \{entity\} full of people'', and ``an empty \{entity\} with no people''.
The same 10-\textit{entity} setup of \cref{sec:retain} is used, separately unlearning each one of the 10 \textit{entities} and evaluating all 10 \textit{entities} with 10 prompts and 4 seeds. The results indicate little difference from the single-prompt setup, as shown in \cref{fig:varied_all}. A closer inspection of the interferences caused by unlearning \textit{football field} illustrates that the interference pattern remains nearly unchanged when evaluating with more prompts (\cref{fig:varied_one}), thus supporting the I-CARE design of single-prompt evaluation.
However, varied phrasings also reveal that some prompts can qualitatively ``avoid'' interference even for heavily affected \textit{entities}: in \cref{fig:varied_images}, where \textit{football field} is overwritten by \textit{the moon}, the basketball courts show the moon contamination under most prompts, yet the ``crowded'' phrasing instead yields unaffected crowded sports venues, potentially because a moon cannot be crowded. We report this as a single illustrative case rather than a quantified effect.

\FloatBarrier
\section{Constructive interference} \label{sec:results_constructive}
\begin{figure}[tb]
  \centering
  \includegraphics[width=1\linewidth]{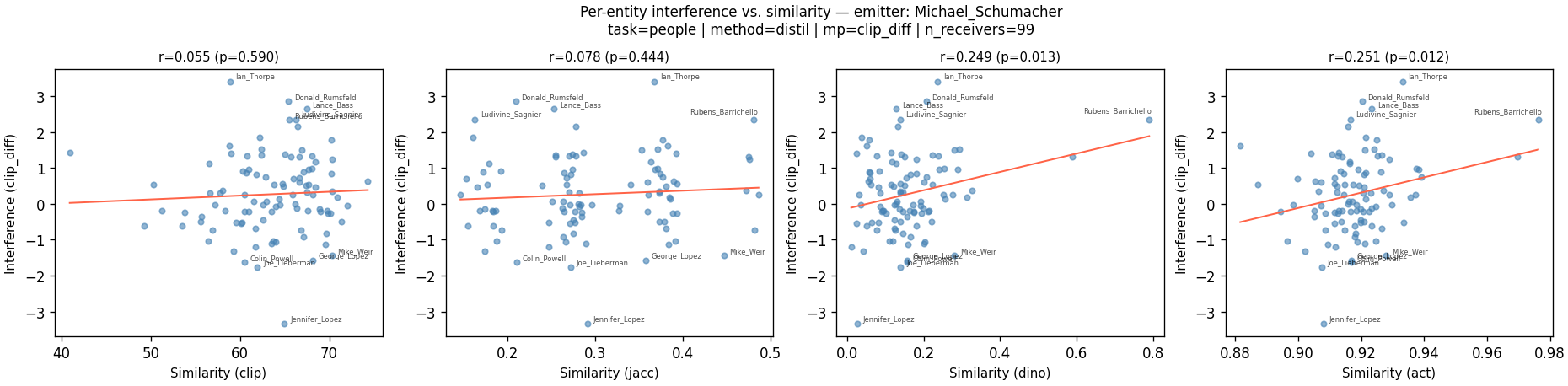}
  \caption{Regression between \textit{$\Delta$Clip} and each of the \textit{similarity} metrics for the unlearning of \textit{Michael Schumacher}.}
  \label{fig:improvements}
\end{figure}

Throughout this work, it is assumed that interference is a destructive phenomenon, and that if a retain \textit{entity} changes during unlearning then this change is for the worse. However, 
among all the unlearning sessions we observed that 2.7\% of the \textit{entities} show a statistically significant positive correlation between \textit{$\Delta$Clip} (the most meaningful metric when it comes to semantic quality) and at least one \textit{similarity} metric. That is, \textit{entities} similar to the one being forgotten end up improving their generative quality, albeit slightly.
The strongest effect was observed while unlearning \textit{Michael Schumacher} (a German F1 driver, \cref{fig:improvements}), in which the 5 most improved \textit{entities} are respectively: \textit{Ian Thorpe} (Australian swimmer, famous at about the same period of time as \textit{Michael Schumacher}), \textit{Donald Rumsfeld} (American politician), \textit{Lance Bass} (American singer), \textit{Ludivine Sagnier} (French actress), and \textit{Rubens Barrichello} (Brazilian F1 driver). 
While it is possible to dismiss those situations as statistical anomalies, it is relevant to highlight that I-CARE would be applicable to a study focused on identifying and predicting the occurrence of such ``constructive interference''.

\section{Analysis of selected sessions}
In order to provide the reader with an intuition about the interference phenomenon, we look closer at two unlearning sessions: the \textit{scene} ``\textit{ice skating rink indoor}'' unlearned with \textit{\ac{spare}}, and the \textit{breed} ``\textit{giant schnauzer dog}'' unlearned with \textit{\ac{uce}}.
In both cases, \textit{$\Delta$Clip} is used as $m_p$ (the most semantic of the five measured \textit{InterferencePerEntityPair}), together with the strongest correlated among the four measured \textit{SimilarityBetweenEntities}.

\begin{figure}[tb]
  \centering
  \includegraphics[width=0.85\linewidth]{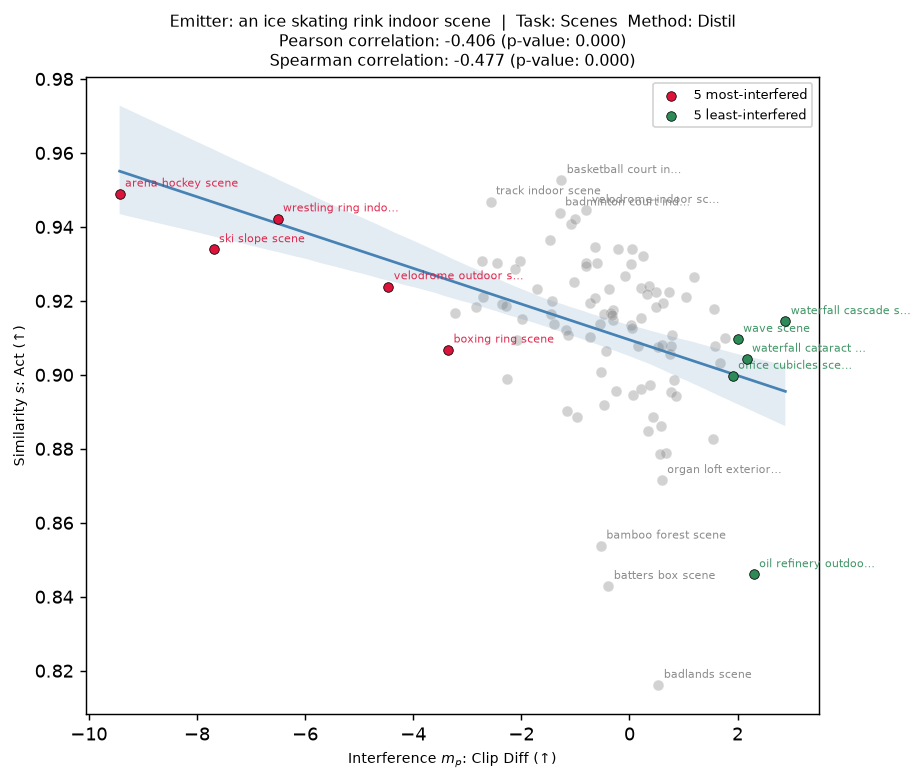}
  \caption{Variation of \textit{MetricSimilarityAlignment} in which only one emitter \textit{entity} is displayed. Five most and least interfered \textit{entities} are highlighted, as well as the five most and least similar.}
  \label{fig:grid_clipdiff_act_top5}
\end{figure}

\begin{figure}[tb]
  \centering
  \includegraphics[width=0.85\linewidth]{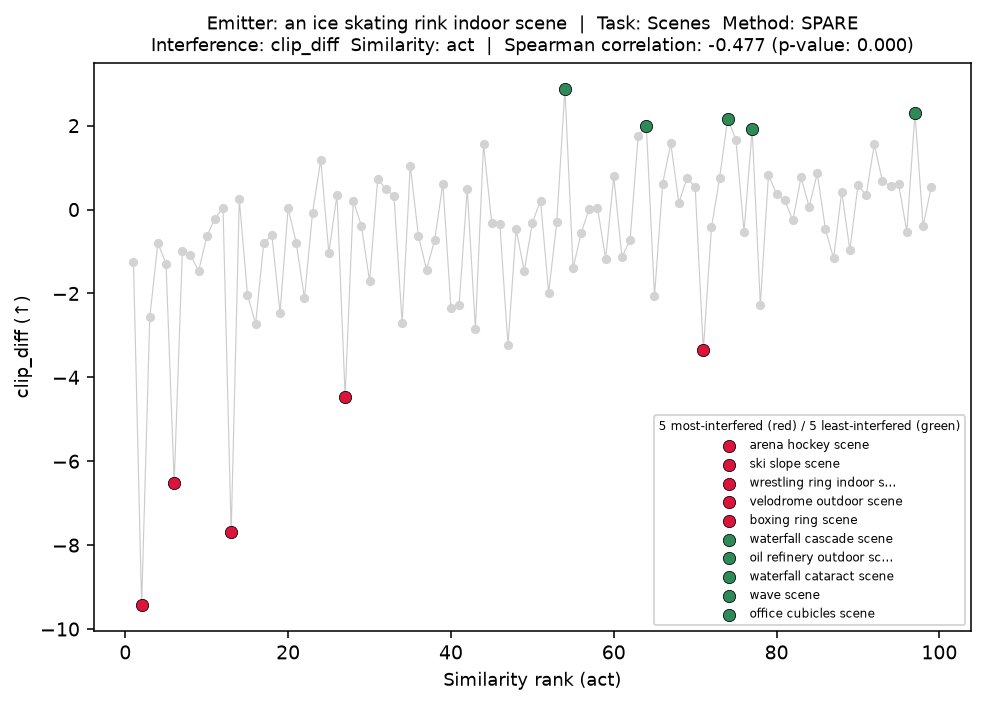}
  \caption{Fundamentally the same information as \cref{fig:grid_clipdiff_act_top5}, but with \textit{similarity} values displayed only as a ranking for clarity.}
  \label{fig:msaone_rank_ice_skating_distil_clip_diff_act}
\end{figure}

\begin{figure}[tb]
  \centering
  \includegraphics[width=1\linewidth]{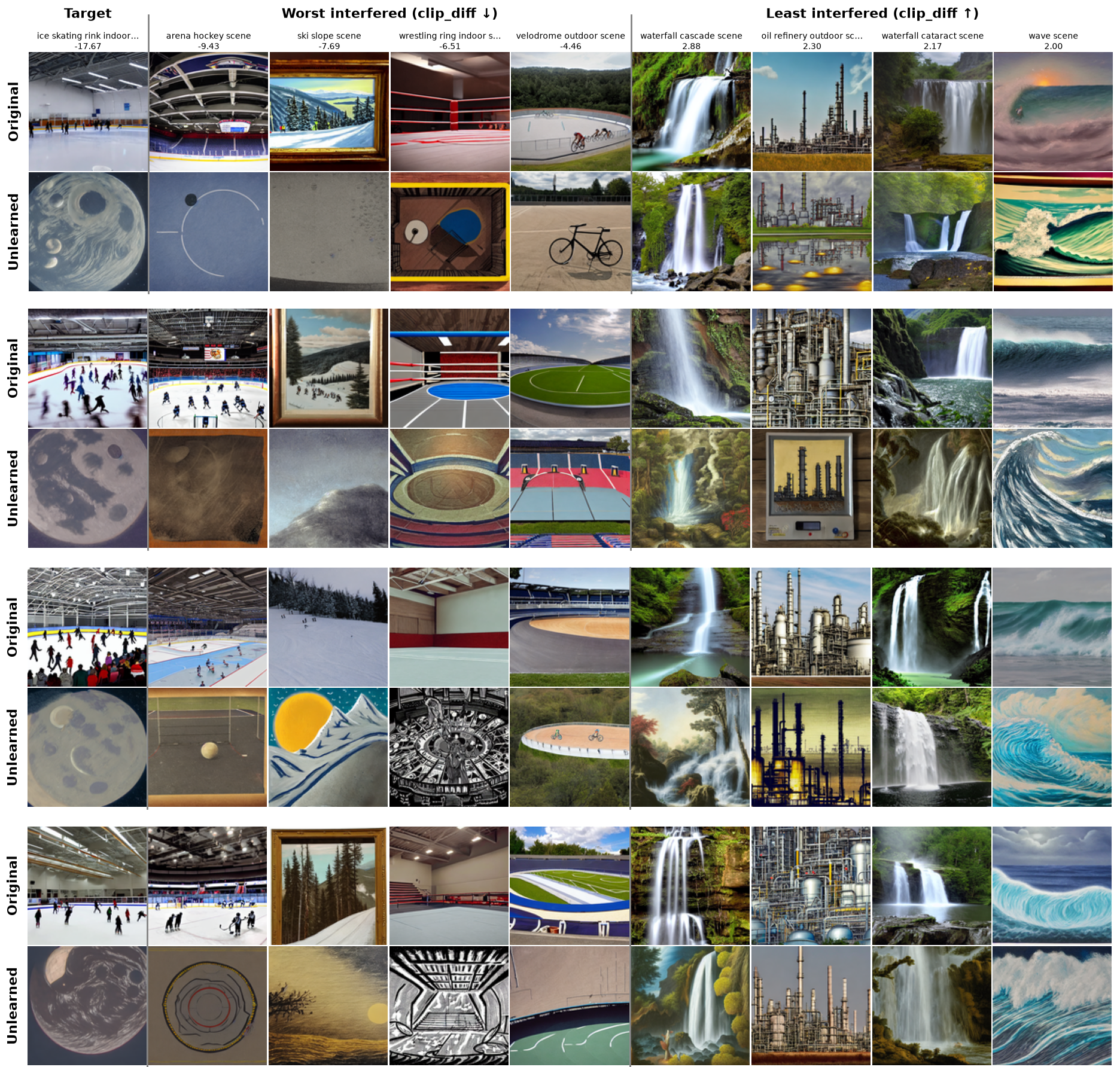}
  \caption{Generated images for the original \textit{model} and for the unlearned \textit{model}, for the \textit{entity} being unlearned as well as for the 4 most and least interfered \textit{entities}, for each of the 4 used seeds. Due to the problem described in \cref{sec:seed}, original and unlearned \textit{models} used different seeds.}
  \label{fig:visual_summary_skating_distil}
\end{figure}

When unlearning ``\textit{ice skating rink indoor}'', replacing it by the concept ``the moon'' the five most interfered \textit{entities} are ``\textit{arena hockey}'', ``\textit{ski slope}'', ``\textit{wrestling ring indoor}'', ``\textit{velodrome outdoor}'', and ``\textit{boxing ring}'', as shown in \cref{fig:grid_clipdiff_act_top5} and \cref{fig:msaone_rank_ice_skating_distil_clip_diff_act}.
While ``\textit{ice skating rink indoor}'' is clearly displayed as a moon in the unlearned \textit{model} (\cref{fig:visual_summary_skating_distil}), the interfered \textit{entities} show varied degrees of degradation, ranging from lower image quality to becoming moon-like.

The least interfered \textit{entities} show a small improvement in their \textit{Clip} \textit{quality}.
While ``\textit{waterfall cascade}'' and ``\textit{oil refinery outdoor}'' are more similar to the target \textit{entity} than ``\textit{boxing ring}'', the latter suffered more interference. 
We refrain from attempting a explanation for this situation, since we are analyzing a single unlearning session from a single \textit{method}.

\begin{figure}[tb]
  \centering
  \includegraphics[width=0.85\linewidth]{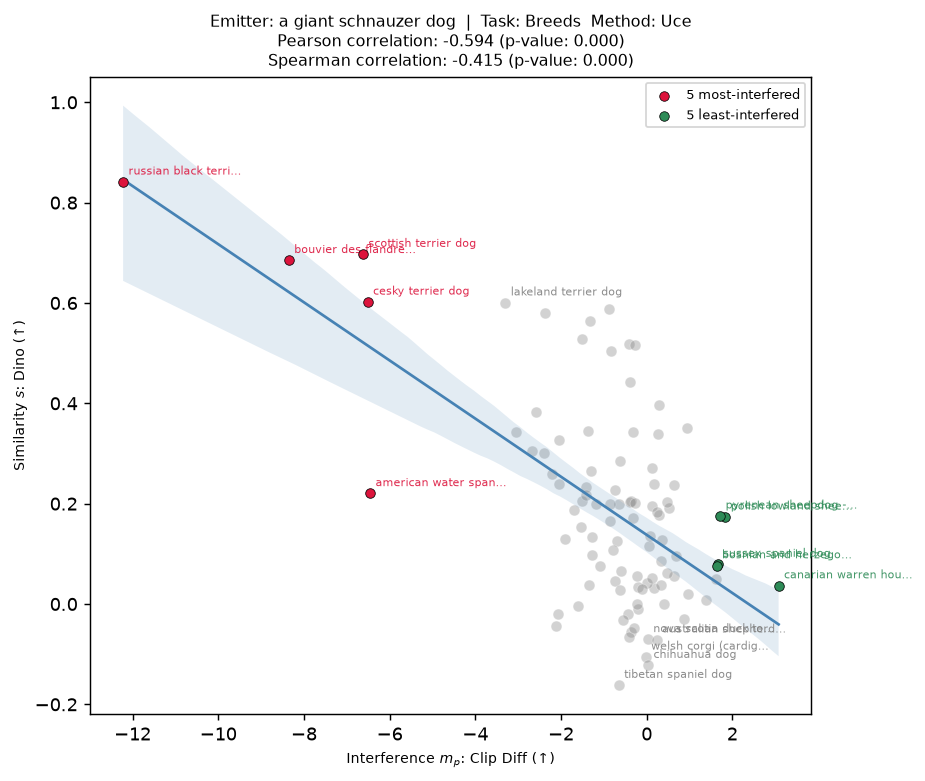}
  \caption{Variation of \textit{MetricSimilarityAlignment} in which only one emitter \textit{entity} is displayed.}
  \label{fig:msaone_giant_schnauzer_clip_diff_dino}
\end{figure}

\begin{figure}[tb]
  \centering
  \includegraphics[width=0.85\linewidth]{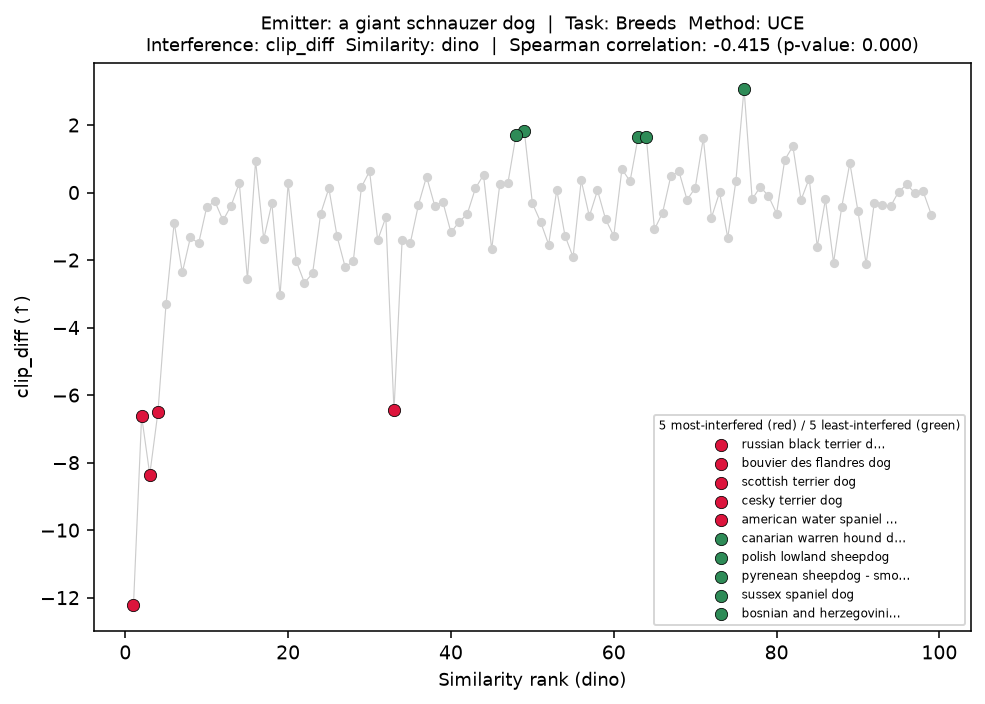}
  \caption{\textit{Similarity} values displayed only as a ranking for clarity.}
  \label{fig:msaone_rank_giant_schnauzer_uce_clip_diff_dino}
\end{figure}

\begin{figure}[tb]
  \centering
  \includegraphics[width=1\linewidth]{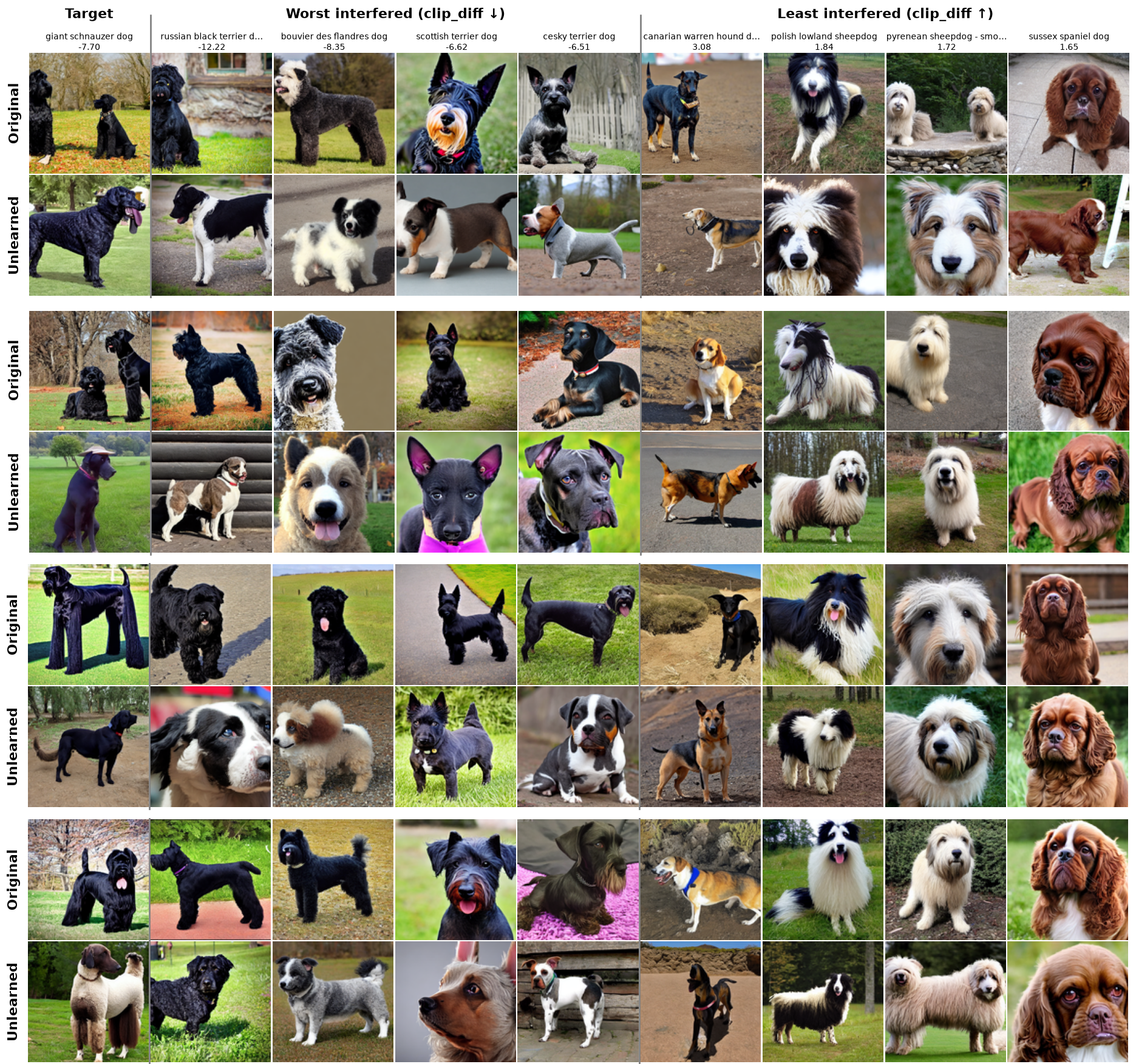}
  \caption{Generated images for the original \textit{model} and for the unlearned \textit{model}. As described in \cref{sec:seed}, for \textit{breeds} the seed problem was corrected, and original and unlearned \textit{models} used the same seeds.}
  \label{fig:visual_summary_schnauzer_uce}
\end{figure}

When unlearning ``\textit{giant schnauzer dog}'' with \textit{\ac{uce}} (that does not perform concept replacement), the four most harmed \textit{entities} correspond exactly to the four most similar \textit{entities} (\cref{fig:msaone_giant_schnauzer_clip_diff_dino} and \cref{fig:msaone_rank_giant_schnauzer_uce_clip_diff_dino}, measured by \textit{Dino}). We observe that the main features distinguishing that \textit{breed} (such as long beard on the muzzle and prominent eyebrows, as shown in \cref{fig:visual_summary_schnauzer_uce}) are gone, even though a large dog is still generated. The two most interfered \textit{entities}, 
``\textit{russian black terrier}'' and ``\textit{bouvier des flandres}'', are also erased, even to a higher $\Delta$\textit{Clip} than the target. The least interfered \textit{entities} correspond mainly to smaller \textit{breeds} or with a fluffier coat.

\clearpage
\chapter{Forgety architecture and features} \label{sec:forgety}
\begin{figure}[tb]
  \centering
  \includegraphics[width=0.8\linewidth]{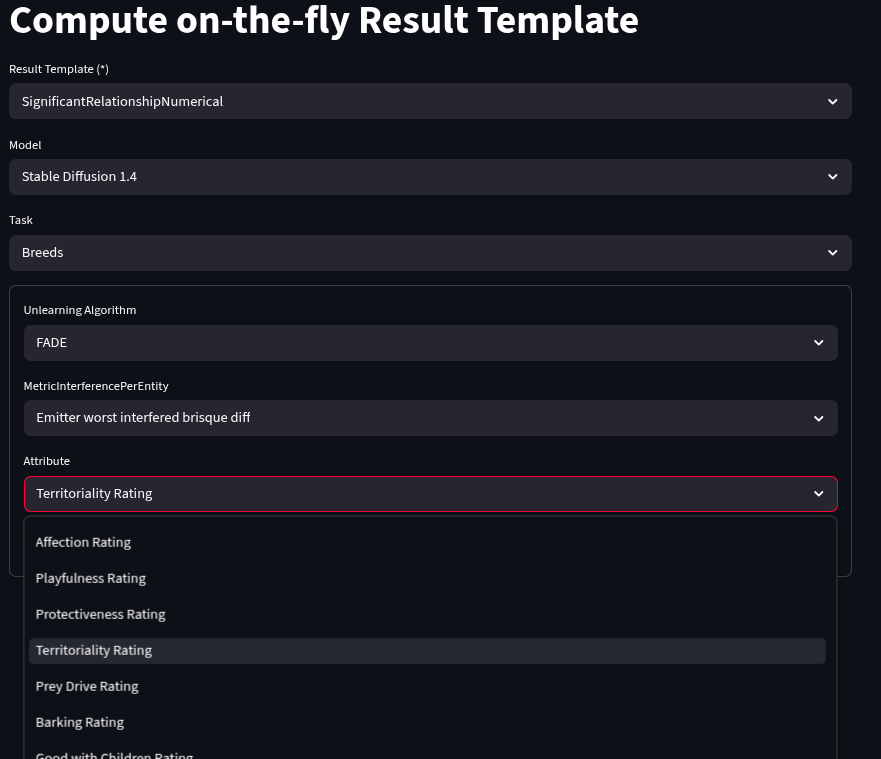}
  \caption{Screen capture for requesting the computation of a \textit{SignificantRelationshipNumerical} \ac{RT}.}
  \label{fig:screen_rt_1}
\end{figure}

\begin{figure}[tb]
  \centering
  \includegraphics[width=0.8\linewidth]{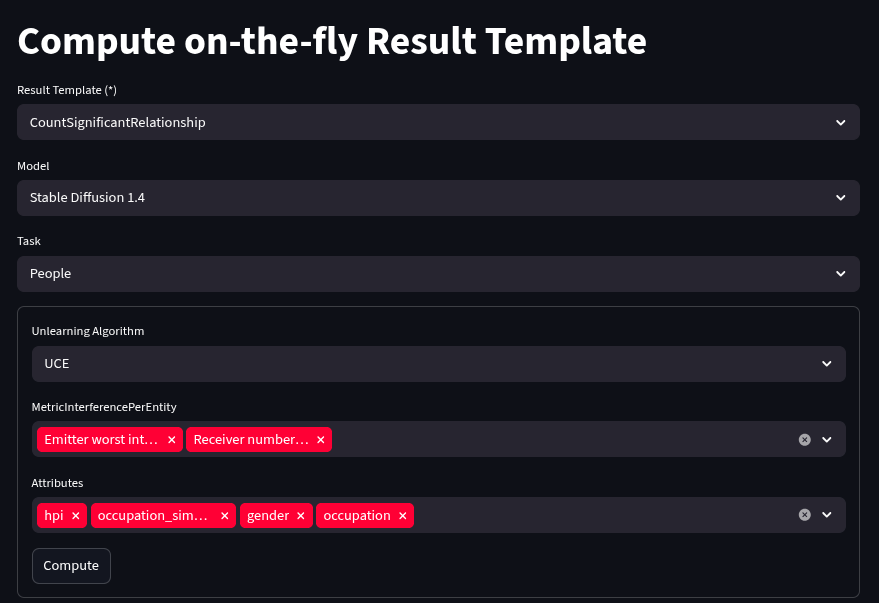}
  \caption{Screen capture for requesting the computation of a \textit{CountSignificantRelationship} \ac{RT}.}
  \label{fig:screen_rt_2}
\end{figure}

\begin{figure}[tb]
  \centering
  \includegraphics[width=0.8\linewidth]{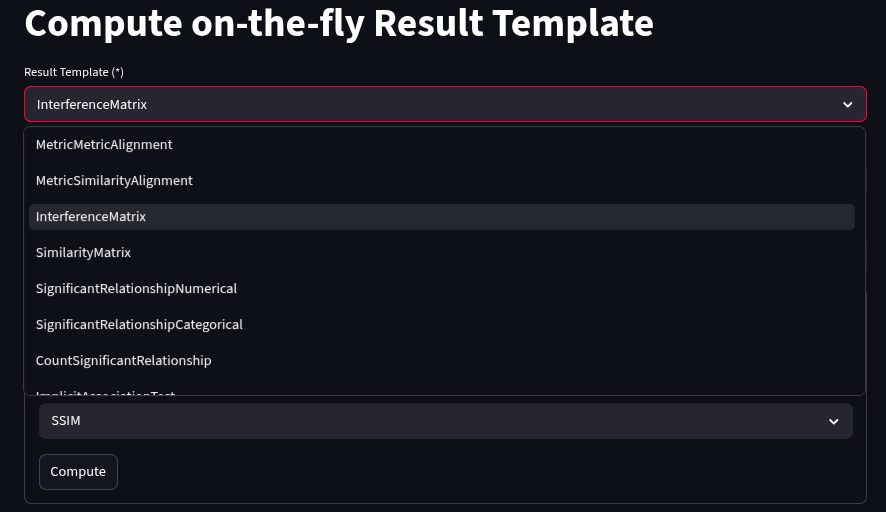}
  \caption{Screen capture for requesting the computation of an \textit{InterferenceMatrix} \ac{RT}.}
  \label{fig:screen_rt_3}
\end{figure}

\begin{figure}[tb]
  \centering
  \includegraphics[width=1\linewidth]{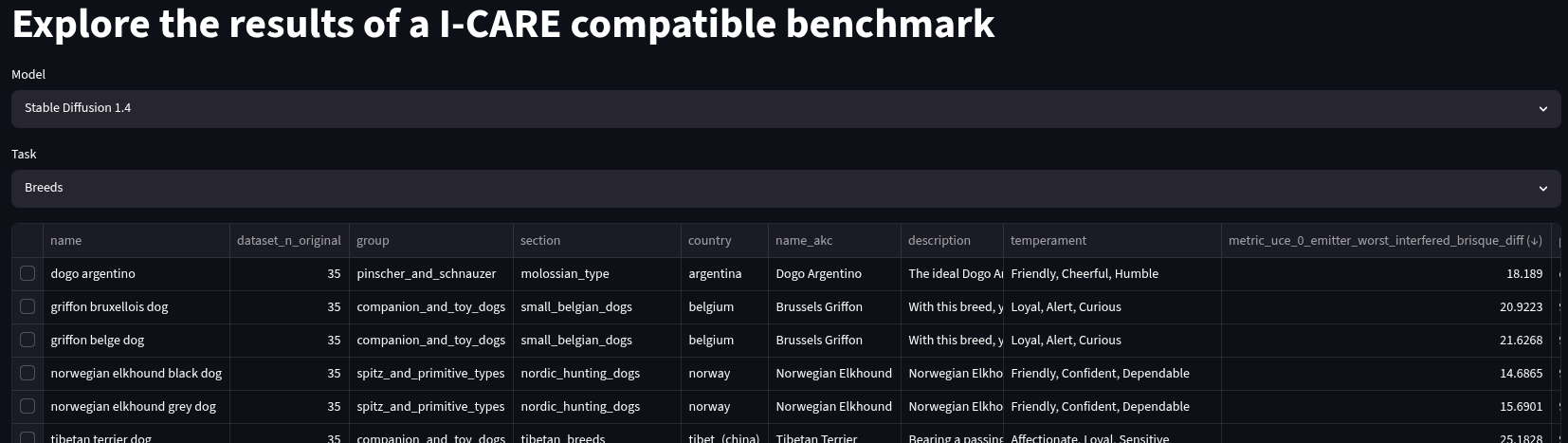}
  \caption{Screen capture for listing \textit{entities}.}
  \label{fig:screenshot_list}
\end{figure}

\begin{figure}[tb]
  \centering
  \includegraphics[width=0.8\linewidth]{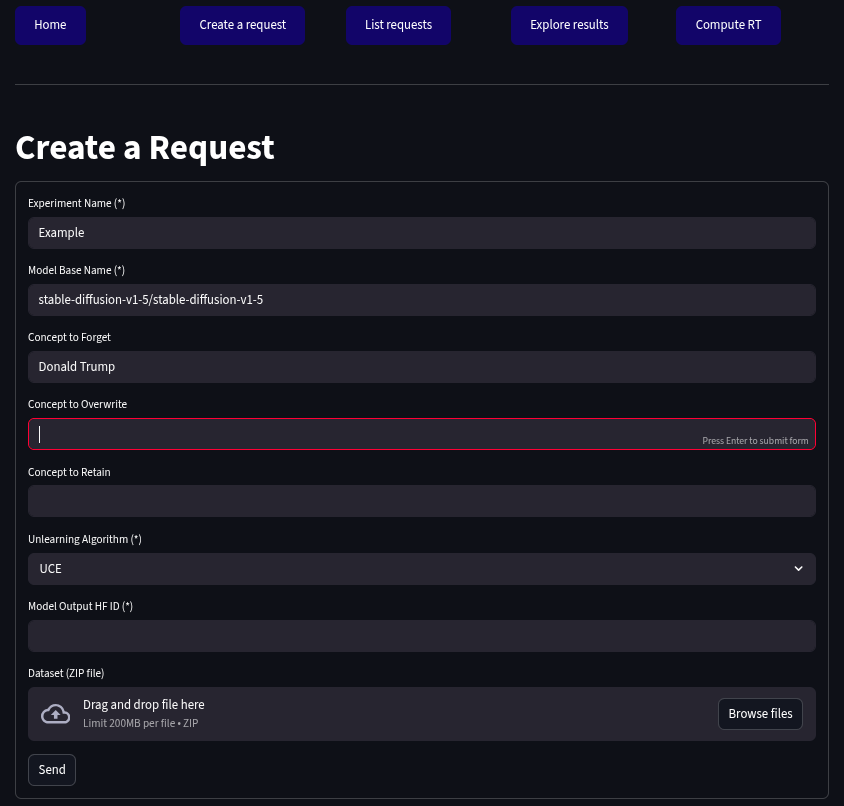}
  \caption{Screen capture for performing a new unlearning session.}
  \label{fig:screen_form}
\end{figure}

\begin{figure}[tb]
  \centering
  \includegraphics[width=1\linewidth]{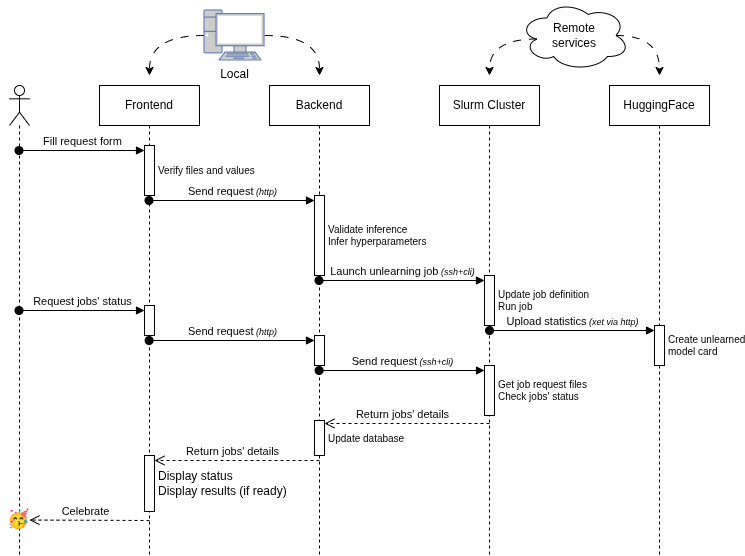}
  \caption{Sequence diagram for performing an unlearning session.}
  \label{fig:uml_sequence}
\end{figure}

Forgety is a web-based Graphical User Interface that enables experimenting with unlearning in a no-code manner, empowering a broader public (such as policy-makers, philosophers, among others) to explore the benefits and shortcomings of \ac{MU}.
Its main features are: 
\begin{itemize}
    \item Compute \acp{RT} on-the-fly, where the user specifies the parameters and receives the computed result (\cref{fig:screen_rt_1}, \cref{fig:screen_rt_2}, and \cref{fig:screen_rt_3} display a few screen captures of this process).
    \item List \textit{entities}, their \textit{attributes}, and their \textit{InterferencePerEntity} values (\cref{fig:screenshot_list}).
    \item Perform a new unlearning session. This feature requires configuring a Slurm cluster to execute the unlearning process. A screen capture is shown in \cref{fig:screen_form}, and its sequence diagram is shown in \cref{fig:uml_sequence}.
    \item List previous unlearning sessions.
\end{itemize}

\begin{figure}[tb]
  \centering
  \includegraphics[width=1\linewidth]{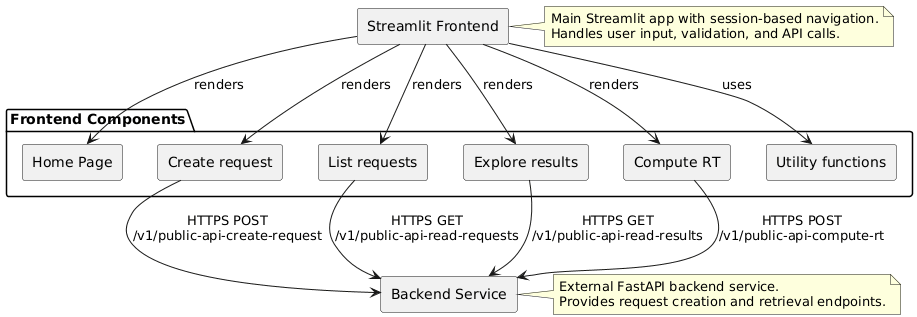}
  \caption{Component diagram for the frontend service.}
  \label{fig:uml_components}
\end{figure}

These features are orchestrated by the frontend services, as shown in the component diagram of \cref{fig:uml_components}. Although initially developed as a proof of concept for I-CARE, there are several possible improvements and expansions, and we envision that the Forgety project could ultimately grow into a platform for exploring results of I-CARE-compatible experiments.

\clearpage
\chapter{Future works}

Several directions would strengthen both the I-CARE methodology and its
empirical grounding. A first line concerns the prediction of interference.
Our current definition of \textit{SimilarityBetweenEntities} characterize each pair of
\textit{entities} in isolation, whereas interference may be better anticipated
by inspecting the unlearning process itself, for example through the gradient
conflicts that arise between the forget and retain objectives. Incorporating
such a metric would, however, break the current abstraction: \textit{SimilarityBetweenEntities}
would no longer receive only \textit{entity}$_1$ and \textit{entity}$_2$, but
also the \textit{model} and the \textit{unlearningMethod}. The formalism would
therefore have to be adapted before a process-aware \textit{similarity} could be
integrated.

A complementary line concerns the sensitivity of the interference metrics
themselves, which presently appear noisy and capture only the most severe
effects. Domain-specific metrics could expose subtler interferences: for the
\textit{people} \textit{task}, a classifier predicting the profession of the generated
person could be compared against the ground-truth \textit{attribute} to reveal shifts
that the generic metrics miss; for the \textit{breeds} \textit{task}, when an
\textit{unlearningMethod} overwrites the \textit{entity} with ``a cat'', a
cat-classifier could detect small changes that made the retained images more
``cat-like''. Because these \textit{tasks} are hard and their interference patterns are
complex, it would also be valuable to test the methodology in simpler,
controlled domains (such as forgetting geometric shapes), where the core
mechanisms of interference could be isolated before returning to naturalistic
\textit{tasks}.

An important generalization of the I-CARE methodology is to account for
concept revival during sequential unlearning \cite{George2025}. Apart from the
high computational cost of combining unlearning sessions, there is no
conceptual obstacle: \textit{unlearningMethod} can be elegantly modified to accept
several \textit{entities} as arguments, and similarly every \ac{RT} that
receives $u$ as an argument can be generalized. Furthermore, the current
methodology can be generalized by allowing \textit{unlearningMethod} to represent,
instead of only a \ac{MU} method, any finetuning process that incurs the risk
of catastrophic forgetting or similar issues, such as model adaptation,
transfer learning, knowledge editing, and continuous learning.

Beyond the provided feasibility demonstration, it is important to benchmark
more \textit{models} and \textit{unlearningMethods}. 
Also, the seed inconsistency described in
section~\ref{sec:seed} should also be addressed by rerunning the dataset generation
and the subsequent metric computation under matched seeds. We speculate that the
\textit{methods} causing the least interference will share common characteristics; for
example, it could be that data-free \textit{methods} tend to cause less interference
than distillation-based \textit{methods}, as tentatively suggested by the
\textit{CountSignificantRelationship} between \textit{\ac{uce}} and
\textit{\ac{spare}}, even though the relationship does not hold for
\textit{\ac{munba}}. Similarly, the hyperparameter choices for a given
\textit{method} may play a crucial role in balancing unlearning quality
(forget and retain metrics) against caused interference. The obtained results
may in turn be used to design unlearning \textit{methods} that reduce interference, for
example by finetuning the entities identified by
\textit{minimumCutInterference} so as to increase their retention. They may
also inspire and validate explanation methods that shed light on the causes of
interference, such as underlying entanglements, and on how those causes affect
the unlearning process itself.

Finally, the tooling could be made considerably easier to adopt. A fully
declarative interface, in which the user specifies the \textit{entities} and
the metrics of interest and the framework prepares everything needed to compute
the results, would lower the barrier for new \textit{tasks}. For Forgety specifically,
adding tools for interactive analysis of the results would further empower
non-technical professionals to draw meaningful conclusions. Moreover, while
Forgety can orchestrate the execution of an unlearning session, it does not yet
orchestrate the image generation and metric computation steps of the pipeline;
implementing this feature would greatly decrease the knowledge required to
reproduce the results of this work and to extend the analysis to other tasks.

\section{Mapping existing benchmarks into I-CARE} \label{sec:unlearncanvas}

A promising direction is to express existing unlearning benchmarks in the I-CARE notation. Most benchmarks define their concepts, metrics, and reporting formats in an ad hoc manner, which makes their results hard to compare and their tooling hard to reuse. Re-describing a benchmark in terms of the I-CARE ontology yields three benefits: heterogeneous benchmarks become mutually comparable under a shared grammar; the reusable software components and \acp{RT} can be applied without reimplementation; and any benchmark that emits results in the standardized format of \cref{ap:metadata} becomes immediately consumable by Forgety and other downstream applications. Many of these benchmarks also evaluate aspects beyond interference, such as forget quality, retainability, efficiency, and robustness, so mapping them extends the I-CARE formalization and software to the broader unlearning field. Any I-CARE-compliant implementation would be complementary to the original work, which retains its own design and findings.

UnlearnCanvas~\cite{Zhang2024UnlearnCanvas} is a well-established benchmark, which makes it a good first candidate for such a mapping. Its targets are 60 artistic styles and 20 objects, which can be modeled as a single \textit{task} of 80 \textit{entities}, each annotated with a categorical \textit{attribute} $domain$ valued \textit{style} or \textit{object}. Since every image contains both a style and an object, the \textit{model} generates a style \textit{entity} by averaging over many objects (and an object \textit{entity} over many styles), while the accompanying ViT classifier reads only the relevant axis and induces a natural \textit{metricQuality}.
Three of UnlearnCanvas' core metrics follow from this. Unlearning Accuracy (the proportion of target-prompt images no longer recognized as the target) can be formalized as a \textit{InterferencePerEntity}, and similarly In-domain and Cross-domain Retain Accuracy by additionally filtering the receiver entities according to whether their $domain$ matches that of the unlearned \textit{entity}. Because the testbed makes recognition nearly perfect before unlearning, these post-unlearning accuracies conceptually coincide with the change-based interference quantities used elsewhere in I-CARE, and the same filter applied at the pairwise level could reproduce UnlearnCanvas' accuracy heatmap as an \textit{InterferenceMatrix}. The remaining metrics are not about interference but could reuse the same machinery: distribution-level quality (FID) could be expressed as a task-level \ac{RT}, and the efficiency metrics (run-time, memory, storage) are accommodated by the existing \textit{MetricInterferencePerEntity} signature without modification.

The Holistic Unlearning Benchmark (HUB)~\cite{Moon2024HUB} evaluates 33 concepts across six perspectives: faithfulness, alignment, pinpoint-ness, multilingual robustness, attack robustness, and efficiency. Of these, only pinpoint-ness concerns interference, and even it selects the receiver concepts dynamically per target rather than over a fixed shared set, which conflicts with I-CARE's all-versus-all design; recovering that structure would require running the remaining sessions so that every concept also acts as an emitter, at substantial computational cost. The other perspectives could still be borrowed into the framework, for instance by treating each language or adversarial prompt generator as a distinct \textit{model}.

\section{Generalization beyond unlearning} \label{sec:care}
Drawing inspiration from the term ``Specification-Driven Development'' \cite{ostroff2004}
we propose the term ``Specification-Driven Science'' as a future extension of the guiding philosophy of I-CARE.

In this approach, the main contribution of a novel scientific work is to propose a conceptual scheme (formal definitions, restrictions on the possible choices, and RTs) for solving a certain type of problem, rather than the solution for one particular problem.
The specific results reported can be thought of as mere automated tests: pairs of input-output that serve for checking the correctness of software implementations of the proposed methodology.
On a broader framing, the role of academia becomes then to provide the framework, and the role of industry to apply the framework to economically relevant problems.

Exploring the philosophical foundations and methodological implications of such future development lies beyond the scope of this work, and we do not claim I-CARE provides even a example of how such generalization could be. 
We also do not claim complete novelty in this proposal, and several prior works \cite{wilson2014, sandve2013, chen2025aria, hellum2025} have indicated similar directions.


\clearpage
\chapter{Conclusions}

I-CARE provides a starting point for the design and implementation of novel studies and benchmarks, as well as inspiration for new methodologies that aim to improve the standardization and reusability in the \ac{MU} field. To the best of our knowledge, no prior work has provided a similar methodology for analyzing interference or related phenomena. 
The software implementation is designed to flexibly handle new \textit{tasks}, new \textit{similarity} metrics, new \textit{metricInterferencePerEntityPairs}, as well as every other explicitly defined concept within I-CARE. Similarly, Forgety is designed to work out-of-the-box with any HuggingFace repository whose files are structured as specified in the software implementation. Both are provided as free and open-source software, and welcome contributions from the community.

The presented feasibility demonstration validates the main aspects of the proposed methodology, but drawing general conclusions about the interference phenomenon was shown to be very difficult. 
One reasons is that the all-vs-all strategy of I-CARE uncovered that often similar \textit{entities} are not the ones that interfere the most, and even statistically significant improvement are in some cases observed (to which we coined the novel term ``constructive interference'').
Another indication of the underlying difficulty is that, in most RTs, the Pearson correlation was higher than the Spearman correlation, indicating that a few influential samples are responsible for the observed correlations.
Even though several techniques for measuring similarities and for quantifying interferences were tested, none demonstrated predictive power that is likely to be useful for \ac{MU} practitioners in a completely new \textit{task}.
Given these challenges, we can not assert the broader validity and generalization to other datasets of the specific results obtained. However, all computed results were made available, and we invite social scientists to explore the thousands of computed \ac{RT} outputs and discuss the broader consequences of interference, such as algorithmic biases and loss of variety in generative models. 

While only three \ac{MU} \textit{methods} were tested, the three analyzed \textit{tasks} are the first benchmarks ever developed for analyzing interference. 
We do not claim to have analyzed all dimensions in which those \textit{methods} differ, but our results uncover initial differences between \textit{\ac{spare}}, \textit{\ac{munba}}, and \textit{\ac{uce}} in the extent and predictability of their caused interferences. 
Most importantly, \textit{\ac{spare}} caused more $\Delta$\textit{Clip} than \textit{\ac{uce}}, but its interference was more easily predicted by \textit{similarity} metrics and equally aligned to the annotated \textit{attributes}; By all analyzed criteria, \textit{\ac{munba}} performed worse than the other two \textit{methods}.


\bibliographystyle{ieeetr}
\addcontentsline{toc}{chapter}{\bibname}
\bibliography{references/bib_ai,references/bib_ai_unlearning,references/bib_IT,references/bib_electronics,references/bib_others}

\begin{appendices}

\chapter{Hardware and software utilized}
\label{ap:materials}

The experiments and development described in this work were executed on an NVIDIA Tesla V100-PCIE-16GB GPU and an NVIDIA Tensor Core A100-PCIE-40GB made available through the Pázmány Péter Catholic University High-Performance Computing cluster (Esztergom).
Cluster access took place between January and June via the institutional Slurm scheduler (version 23.11.4) running on the operating system CentOS 8.10 ``Green Obsidian''. 
We provisioned approximately 1200 GPU hours for the reported experiments; using the Hungarian grid carbon intensity, this corresponds to an estimated footprint of 180.69
 kgCO\textsubscript{2}e.
The overall compute cost is substantial: each unlearning session takes approximately one hour, image generation for a single unlearned \textit{entity} takes about 1.5 hours, metric computation takes roughly 30 minutes, and the final analysis stage can take up to one hour (although it is executed only once).

The software implementation was written in Python 3.10 and relies on PyTorch 2.8, Diffusers 0.30, Transformers 4.55, Peft 0.17, Vision-Unlearning 0.1.6, among other minor dependencies. Model artifacts and checkpoints were managed using the Hugging Face Hub.  
Development and experiment orchestration used Poetry for dependency management, Apptainer for containerized execution, and a Makefile to automate common workflows.

Stable Diffusion v1.4 was utilized as the base generative model. 
The designed unlearning tasks and their associated image datasets are: unlearning one person from the dataset Labeled Faces in the Wild (LFW) \cite{dataset_lfw}, which is intended for studying unconstrained face recognition problems (in which there is little control over parameters such as position, pose, lighting, background, camera quality); 
unlearning a dog breed from the AtharvaTaras Dog Breeds Dataset \cite{dataset_taras_dog_breeds}, with 356 breeds recognized by the FCI (Fédération Cynologique Internationale) and containing 35 images for each breed; 
unlearning a scene from the SUN Attributes \cite{dataset_sun_attributes} dataset, intended for fine-grained scene categorization and containing more than 14,000 images from 717 classes.


\clearpage
\chapter{Standardized data and metadata files}
\label{ap:metadata}

Across all tasks, a standardized set of files/media/metadata/content is provided:

{\bf MetadataFiltered}. List of dictionaries where each entry refers to one \textit{entity}. One file for the entire \textit{task}.  
\underline{Save path}: \path{metadata_{task}_2_enriched_filtered.json}.  
\underline{Fields}: \path{name}, as labeled in the main image dataset (also referred to as ``non-preprocessed''); \path{index}, an integer identifier of the \textit{entity}; \path{is_unlearned}, indicating whether the \textit{entity} belongs to the unlearning target set (if true, then of course \path{dataset_n}>0; unused for now, since all chosen \textit{entities} are unlearned and analysed); \path{dataset_n_original}, the number of samples in the original dataset (0 if the \textit{entity} does not appear even in the retain set; unused for now, since all \textit{entities} have data); All the \textit{attributes} specific to the \textit{task} under consideration.

{\bf Similarity}. $100\times100$ matrix with pairwise \textit{similarities} of one \textit{SimilarityBetweenEntities}. Indexed by \path{name} field of \textit{metadataFiltered}. One file per combination of \textit{task} and \textit{SimilarityBetweenEntities}.
\underline{Save path}: \path{similarity_{similarity}_{task}.json}.

{\bf TrainingDataSplits}. Data splits ready for training, with separate folders for forget and retain images.  
\underline{Save path}: \path{{dataset_base_path}/{target}/train_forget/}.  
\underline{Save path}: \path{{dataset_base_path}/{target}/train_retain/}.

{\bf UnlearnedModels}. Finetuned stable-diffusion-v1-4 \textit{models} that forgot the target \textit{entity}. Each \textit{model} is stored in a separate folder.  
\underline{Save path}: \path{models/{task}_{target}_{method}_{num_train_epochs:03d}/}.

{\bf GeneratedData}. Data generated by unlearned \textit{models}.  
\underline{Prompt}: ``An image of \{name\}''.  
\underline{Save path}: \path{datasets/generated_{task}_{target}_{method}_{num_train_epochs:03d}/}.

{\bf InterferencePerPair}. Dictionary storing computed \textit{MetricInterferencePerEntityPair}. Each key refers to one \textit{entity} (including the target), as it appears in the name fields of \textit{metadataFiltered} (non-preprocessed). The entire dictionary refers to interferences caused by one emitter to each receiver (the keys of the dictionary). Values are the metrics computed, averaged across all seeds. The dictionary is always complete (all values computed). One file per unlearning session.  
\underline{Computed by}: \path{3_compute_caused_interferences.py} (benefits from GPU but can run on CPU).  
\underline{Save path}: \path{datasets/interferences_caused_by_{task}_{index}_{method}_{num_train_epochs}.json}.

{\bf InterferencePerPairInverse}. Dictionary representing interferences received by one \textit{entity} from each emitter (the keys of \textit{InterferencePerPair}). It may be incomplete if not all unlearning sessions were performed. One file per \textit{entity} and per unlearning session, but using information aggregated from all unlearning sessions.  
\underline{Save path}: none, calculated on-the-fly.

{\bf InterferencePerEntity}. List of dictionaries storing computed \textit{MetricInterferencePerEntity}, together with all information from \textit{metadataFiltered}. Each item corresponds to one \textit{entity}. One file for the entire \textit{task}, for all \textit{methods}. Columns include the fields from \textit{metadataFiltered}, followed by each \textit{MetricInterferencePerEntity} prefixed by \path{metric_{method}_{num_train_epochs}_} and suffixed by $(\uparrow)$ or $(\downarrow)$.  
\underline{Computed by}: \path{4. Compute interference per entity.ipynb} (does not benefit from GPU).  
\underline{Save path}: \path{interference_per_entity_{task}.json}.

{\bf ResultTemplatesOutputs}. Outputs of the result templates. All parameters that uniquely identify the performed analysis are included, as well as all data required to plot the results (thus no GPU nor any form of heavy computation is needed).  
\underline{Save path}: \path{results/{result_template_name}/{named_arguments_serialized}.json}.

\clearpage
\chapter{RT SignificantRelationship}
\label{ap:rt_relationship}

Considering first the categorical formulation, let $t \subseteq E$ denote the set of \textit{entities} in the \textit{task}, and let
$a : E \rightarrow V$ be a categorical \textit{attribute} whose possible
values are $a_1, a_2, \dots, a_{|a|}$.
For each \textit{attribute} value $v \in V$ we define the subset
\begin{equation}
t_v = \{\, e \in t \mid a(e) = v \,\}.
\end{equation}

Let $m_e : E \rightarrow \mathbb{R}$ denote the chosen
\textit{MetricInterferencePerEntity}.
The statistical test evaluates whether the distribution of $m_e$
differs significantly across the groups induced by the \textit{attribute} $a$.
Formally, we compute

\begin{equation}
\operatorname{StatisticalTest}\Big(
\big\{
\big( m_e(e) \big)_{e \in t_v}
\;\big|\;
v \in \{\, a(e) \mid e \in t \,\}
\big\}
\Big).
\end{equation}

Intuitively, \textit{entities} are partitioned according to the values of the
categorical \textit{attribute}, and a statistical test (e.g., ANOVA or
Kruskal--Wallis) is applied to the resulting groups of metric values, determining whether the mean interference differs
significantly across \textit{attribute} values.
The same procedure can be expressed in abstract pseudo-code:

\begin{verbatim}
INPUT:
  task t
  categorical attribute a
  entity metric m_e

groups ← empty collection

for each value v in domain(a):
    group ← []
    for each entity e in t:
        if a(e) = v:
            group.append( m_e(e) )
    groups.add(group)

return StatisticalTest(groups)
\end{verbatim}

\section{Considerations about numerical attributes}
The same conceptual framework extends naturally to numerical
\textit{attributes}. Instead of partitioning \textit{entities} into discrete groups, the
analysis evaluates the statistical relationship between the numerical
\textit{attribute} values and the \textit{entity}-level interference metric $m_e$.
Concretely, the procedure computes correlation statistics (e.g.,
Pearson and Spearman coefficients) between the \textit{attribute} values
$a(e)$ and the metric values $m_e(e)$ across all \textit{entities} in the \textit{task}.
The result quantifies whether higher or lower values of the \textit{attribute}
are associated with systematically stronger or weaker interference
effects.

\section{Directional analysis}

The analysis described above evaluates whether $m_e$ differs across groups of \textit{entities}.
Since $m_e$ is a per-\textit{entity} aggregate that marginalizes over all counterparties, it does not preserve information about the source of interference; it is therefore unable to express directional claims of the form ``\textit{entities} of
group $v_1$ cause more interference to \textit{entities} of $v_2$ than of $v_3$.''

To support such claims, the directional extension of this RT replaces $m_e$ with the pairwise \textit{MetricInterferencePerEntityPair} $m_p$, which retains the identity of both the emitting and the receiving \textit{entity}, thus allowing one to configure the subset of emitter \textit{entities} considered in the analysis. It can be formalized as: for a fixed source value $v \in V$, let $t_v$ denote the corresponding source group; for each receiver $r \in t$, the conditioned interference received from the source group is
\begin{equation}
\phi_v(r) \;=\; \frac{1}{|t_v \setminus \{r\}|} \sum_{\substack{e \,\in\, t_v \\ e \,\neq\, r}} m_p(e, r).
\end{equation}
The same statistical procedure as in the non-directional case is then applied to the values $\{\phi_v(r)\}_{r \in t}$, partitioned by $a(r)$. If a statistical difference in the distribution of the values in each partition is observed, then we can interpret that emitter \textit{entities} of type $v$ interfere differently with receiver \textit{entities} depending on their $a(r)$. In practice, the results of these repeated evaluations can be visualized
as directional or flow-based diagrams (e.g., Sankey representations) that highlight how interference propagates between subsets of \textit{entities}.

\clearpage
\chapter{RT MetricSimilarityAlignment}
\label{ap:prediction}

We consider a fixed \textit{model} $m$, \textit{task} $t$, and \textit{unlearningMethod} $u$, which will be omitted for brevity.
The objective is to quantify whether interference between \textit{entities} is aligned with their \textit{similarity}, i.e., to which degree similar \textit{entities} interfere more with each other.

For every ordered pair of distinct \textit{entities} $e_i, e_j \in t$ with $i \neq j$, we observe several \textit{SimilarityBetweenEntities} and several \textit{MetricInterferencePerEntityPair}, which will be respectively differentiated by a superscript $l=1, 2, ..., |S|$ and $k=1, 2, ..., |M_p|$. 
Each ordered pair $(e_i,e_j)$ is therefore treated as one data point with feature vector
\begin{equation}
\mathbf{X}_{ij} = \big(     s^{(1)}(e_i, e_j)  , \dots, s^{(|S|)}(e_i, e_j)  \big)
\end{equation}
and target
\begin{equation}
\mathbf{Y}_{ij} = \big(    m_p^{(1)}(e_i, e_j), \dots, m_p^{(|M_p|)}(e_i, e_j)       \big).
\end{equation}

The resulting dataset is
\begin{equation}
D = \{(\mathbf{X}_{ij}, \mathbf{Y}_{ij}) \mid e_i, e_j \in t,\, i \neq j\}.
\end{equation}

from which a regression model can be estimated using standard regression procedures with appropriate validation.
In the linear case,
\begin{equation}
Y^{(k)}_{ij} =
\beta^{(k)}_0 +
\sum_{\ell=1}^{|S|}
\beta^{(k)}_{\ell} X^{(\ell)}_{ij}
+
\varepsilon^{(k)}_{ij}.
\end{equation}

Given a specific \textit{entity} $e_i$ whose removal is considered, similarities 
$X^{(\ell)}_{ij} = s^{(\ell)}(e_i,e_j)$
can be computed for all remaining \textit{entities} $e_j \in t$. The fitted model then yields predictions
$\hat{Y}^{(k)}_{ij} = f^{(k)}(\mathbf{X}_{ij})$,
which approximate the expected interference on each receiver \textit{entity}.

This formulation already encompasses both the simple case of measuring alignment between a single $m_p$ and a single $s$, and the more general case where multiple metrics and similarities are jointly considered.
Furthermore, the concept of \textit{similarity} can also represent several forms of practical data engineering; for example, one may define a \textit{similarity} function for each \textit{attribute}, or a \textit{similarity} function that is based only on the \textit{attributes} of the emitter \textit{entity}.

\clearpage
\chapter{RT MostSimilarInterferedMatrix} \label{ap:prediction_most_similar}
Analyses whether the single most-similar receiver is among the most-interfered receivers, and then counting how often this holds across many configurations.
\textit{MostSimilarInterferedMatrix} operates on the data used by \textit{MetricSimilarityAlignment} (\cref{ap:prediction}), but summarizes it differently. Instead of fitting a regression between \textit{similarity} and \textit{interference per-pair}, it analyzes whether the single most-similar receiver is among the most-interfered receivers, and counts how often this holds across many configurations. It receives as parameter a single \textit{model}, a list of \textit{tasks}, list of \textit{unlearning methods}, list of \textit{metricInterferencePerEntityPairs}, list of \textit{similaritiesBetweenEntities}, and a integer \textit{topK} specifying how many of the most-interfered receivers are to be considered. 

The output is a $method \times task$ grid of counts, sweeping across every combination of emitter, one or more $m_p$, one or more $s$.
For example, for 4 \textit{unlearningMethods} and 3 \textit{tasks}, the output is a $4\times3$ matrix; for 100 \textit{entities} per \textit{task}, 5 \textit{InterferencePerEntityPair} metrics, and 4 \textit{SimilarityBetweenEntities} metrics, the nominal maximum per cell is $100 \times 5 \times 4 = 2000$. For each of the 2000 combinations, it is counted as 1 if the single most-similar receiver is among the emitter's \textit{topK} most-interfered receivers, and 0 otherwise.

Under the null hypothesis that similarity is unrelated to interference, the
most-similar receiver falls in the top-$k$ of the $|t|-1$ receivers with
probability $k/(|t|-1)$, so the expected cell count is
$|M_p|\cdot|S|\cdot|t|\cdot k/(|t|-1)$. A cell well above this
baseline indicates that, for that task and method, the most-similar receiver is
preferentially among the most-interfered.

\clearpage
\chapter{RT ImplicitAssociationTest}
\label{ap:rt_iat}

The Implicit Association Test (IAT) measures the differential associations between \textit{target attributes} $a^{(t)}$ and \textit{protected attributes} $a^{(p)}$ \cite{Sirotkin2022}, while the Image Embedding Association Test (iEAT) \cite{Steed2021} extends this methodology to the visual domain, quantifying the differential associations of different values of \textit{target attributes} $a^{(t)}_1$ and $a^{(t)}_2$ with values of \textit{protected attributes} $a^{(p)}_1$ and $a^{(p)}_2$ (assuming, for simplicity, two categorical \textit{attributes} with two possible values each). The null hypothesis for the iEAT posits that \textit{latent embeddings} of images generated for \textit{entities} with \textit{target attributes} $a^{(t)}_1$ and $a^{(t)}_2$ are equally similar to those with \textit{protected attributes} $a^{(p)}_1$ and $a^{(p)}_2$. Rejection of the null hypothesis indicates that one \textit{target attribute} exhibits a stronger association with one \textit{protected attribute} than the other, thereby detecting an association bias \cite{Sirotkin2022}.

Following the computation of image embeddings per \textit{entity} (averaged across seeds), we construct a matrix $B \in \mathbb{R}^{|a^{(t)}| \times |a^{(p)}|}$ representing the average embedding alignment for each pair of \textit{attribute value} combinations, facilitating the comparison of within-group versus cross-group distances. 

This matrix is independently computed for the original and for the unlearned \textit{models}, respectively denoted as $B_{\text{original}}$ and $B_{\text{unlearned}}$, and the shift in associations during the unlearning process is captured by:
\begin{equation}
\Delta B = B_{\text{original}} - B_{\text{unlearned}}
\end{equation}
from which researchers can evaluate the ethical implications of the changes in implicit associations between the selected \textit{attributes} induced by the unlearning process.

\clearpage\chapter{RT MinimumCutInterference}
\label{ap:graph}

Given a fixed \textit{model} $m$, \textit{task} $t$, \textit{unlearningMethod} $u$, and a chosen \textit{MetricInterferencePerEntityPair} $m_p$, we interpret the \textit{task} as a directed weighted graph.

Let $G = (V, A, w)$ denote a directed weighted graph, where $V = t$ is the set of vertices, $A \subseteq V \times V$ is the set of directed edges, and

\begin{equation}
w : A \rightarrow \mathbb{R}_{\ge 0}
\end{equation}

is the edge weight function.
Edges are defined using the \textit{MetricInterferencePerEntityPair} and a predefined positive threshold $\lambda$.  
For every ordered pair of distinct \textit{entities} $e_i,e_j \in t$, an edge $(e_i,e_j)$ is included in $A$ if

\begin{equation}
m_p(e_i,e_j,m,u) > \lambda,
\end{equation}

and its weight is

\begin{equation}
w(e_i,e_j) = m_p(e_i,e_j,m,u).
\end{equation}

In principle the resulting graph could be fully connected. In practice, the analysis is restricted to the strongest interference relations by keeping only edges within a local neighbourhood. This reduces computational cost and mitigates noise in the interference measurement, such as variability due to poorly performing unlearning runs or a small number of sampling seeds.

Based on this graph representation, the \ac{RT} computes the minimum $e_1$--$e_2$ cut, a partition of the vertex set $V$ into two disjoint subsets $(P_1,P_2)$ such that

\begin{equation}
e_1 \in P_1, \qquad e_2 \in P_2, \qquad P_1 \cup P_2 = V, \qquad P_1 \cap P_2 = \varnothing.
\end{equation}

The capacity of a cut is defined as the total weight of all directed edges that cross from $P_1$ to $P_2$:

\begin{equation}
\mathrm{cap}(P_1,P_2)
=
\sum_{\substack{(e_i,e_j)\in A \\ e_i\in P_1,\; e_j\in P_2}}
w(e_i,e_j).
\end{equation}

The minimum $e_1$--$e_2$ cut is the partition $(P_1,P_2)$ that minimizes $\mathrm{cap}(P_1,P_2)$ among all valid cuts.
Without loss of generality, the cut can be represented either by the set of crossing edges (all edges from $P_1$ to $P_2$), or by the emitter-side node set $P_1$, or by the receiver-side node set $P_2$. The \ac{RT} returns the set of \textit{entities} corresponding to this emitter-side partition.

\clearpage
\chapter{Max-Flow / Min-Cut and Isolation Duality}
\label{ap:flow_isolation}

Let \(G=(V,A,w)\) be a finite directed graph with nonnegative edge capacities \(w:A\to\mathbb{R}_{\ge 0}\).
Fix distinct vertices \(s,t\in V\). (In our application \(V=t\) is the \textit{task} entity set and \(w(u,v)=m_p(u,v)\) is the chosen \textit{MetricInterferencePerEntityPair}). A flow is an assignment \(\{f_{uv}\}_{(u,v)\in A}\) satisfying
\begin{equation}
0 \le f_{uv} \le w_{uv}\qquad\forall (u,v)\in A,
\end{equation}
and, for every \(v\in V\setminus\{s,t\}\),
\begin{equation}
\sum_{w:(v,w)\in A} f_{vw} \;-\; \sum_{u:(u,v)\in A} f_{uv} \;=\; 0.
\end{equation}
The value of a flow \(f\) is
\begin{equation}
|f| \;=\; \sum_{v:(s,v)\in A} f_{sv} - \sum_{u:(u,s)\in A} f_{us}.
\end{equation}

An ordered partition \((S,T)\) of \(V\) with \(s\in S,\; t\in T\) is called an \(s\!-\!t\) cut. Its capacity is
\begin{equation}
\mathrm{cap}(S,T) \;=\; \sum_{\substack{(u,v)\in A\\ u\in S,\ v\in T}} w_{uv}.
\end{equation}

\textbf{Weak upper bound.}
For any feasible flow \(f\) and any cut \((S,T)\) we have
\begin{equation}
|f| \le \mathrm{cap}(S,T).
\end{equation}
Proof: sum the flow conservation equations over all \(v\in S\setminus\{s\}\) and rearrange; only edges from \(S\) to \(T\) contribute net outflow, and each such edge contributes at most its capacity \cite{ahuja1993network}.

\textbf{Residual graph and augmenting paths.}
Given a feasible flow \(f\), define the residual capacity
\begin{equation}
c_f(u,v) :=
\begin{cases}
w_{uv}-f_{uv}, & (u,v)\in A,\\[4pt]
f_{vu}, & (v,u)\in A \text{ (reverse residual)},\\[4pt]
0, & \text{otherwise.}
\end{cases}
\end{equation}
The residual graph \(G_f\) has vertex set \(V\) and directed edges \((u,v)\) with \(c_f(u,v)>0\).
An augmenting path is any \(s\!-\!t\) path in \(G_f\). If an augmenting path exists one may push an additional \(\delta>0\) units of flow along it (where \(\delta\) is the minimum residual capacity on the path), producing a feasible flow of strictly larger value.

Run the Ford–Fulkerson augmenting-path procedure \cite{ahuja1993network}: repeatedly find an augmenting path in \(G_f\) and augment until no augmenting path remains. (For rational/integer capacities the algorithm terminates; for general analysis one may appeal to Edmonds–Karp or preflow-push variants to guarantee termination and polynomial runtime.) Let \(f^*\) be the resulting flow with no augmenting path in \(G_{f^*}\).

\textbf{Constructing a cut from the final residual graph.}
Let \(S\subseteq V\) be the set of vertices reachable from \(s\) in \(G_{f^*}\), and let \(T=V\setminus S\). Since there is no augmenting path, \(t\notin S\). For every edge \((u,v)\) with \(u\in S,\ v\in T\) we have \(c_{f^*}(u,v)=0\), hence \(f^*_{uv}=w_{uv}\) (the edge is saturated). Therefore
\begin{equation}
|f^*|
= \sum_{u\in S,\,v\in T} f^*_{uv}
= \sum_{u\in S,\,v\in T} w_{uv}
= \mathrm{cap}(S,T).
\end{equation}
Combining this equality with the weak upper bound yields the Max-Flow / Min-Cut theorem \cite{ahuja1993network}:
\begin{equation}
\max_{f\ \text{feasible}} |f| \;=\; \min_{(S,T)\ \text{cut}} \mathrm{cap}(S,T).
\end{equation}

\textbf{Corollary: Isolation Duality}. 
Interpret edge capacities as influence strengths. For \(F\subseteq A\) write \(G-F\) for the graph obtained by removing edges in \(F\). Call \(F\) a blocking set for \(s\to t\) iff there is no directed path from \(s\) to \(t\) in \(G-F\). Then
\begin{equation}
\min_{\substack{F\subseteq A\\ F\ \text{blocks }s\to t}} \sum_{e\in F} w_e
\;=\;
\min_{(S,T)\ \text{cut}} \mathrm{cap}(S,T)
\;=\;
\max_{f\ \text{feasible}} |f|.
\end{equation}

Since for any cut \((S,T)\) the set \(F_S=\{(u,v)\in A: u\in S,\ v\in T\}\) blocks all \(s\to t\) paths, so
\(\min_{F\ \text{blocks}}\sum_{e\in F} w_e \le \min_{(S,T)}\mathrm{cap}(S,T)\).
Conversely, if \(F\) blocks all \(s\to t\) paths, let \(S\) be the set of vertices reachable from \(s\) in \(G-F\); then \(t\notin S\) and every edge from \(S\) to \(T:=V\setminus S\) lies in \(F\), yielding
\(\sum_{e\in F} w_e \ge \mathrm{cap}(S,T) \ge \min_{(S',T')}\mathrm{cap}(S',T')\).
Taking infima establishes equality with the minimum cut; the Max-Flow / Min-Cut theorem then gives the stated identity.

\clearpage
\chapter{Benchmark detailed design choices}

\section{Tasks} \label{ap:tasks}
Identifying datasets that satisfy the task-selection criteria outlined in the methodology is non-trivial. Many commonly used datasets fail to meet one or more requirements. For instance, \cite{George2025} generates synthetic data; Unlearn Canvas \cite{Zhang2024UnlearnCanvas} contains only 20 \textit{entities}; Animals-with-Attributes \cite{xian2019animals_with_attributes} is limited to 50 \textit{entities}; CelebA \cite{liu2015caleb_a} and FACET \cite{gustafson2023facet} contain \textit{entities} that are not explicitly named; Objects365 \cite{shao2019objects365} provides only instance-level annotations; DeepFashion \cite{liu2016deepfashion} includes only 50 classes; 
Concepts in the Large-scale Attribute Dataset (LAD) \cite{zhao2018lad} are too coarse-grained, and restricting them to specific subclasses results in fewer than 50 \textit{entities} per category; ImageNet-Attributes \cite{Russakovsky2010} includes 384 classes but is heavily skewed toward animals (only 144 non-animal classes, of which 77 are neither animals nor food).

\section{Attributes} \label{ap:attributes}

The \textit{attributesOfInterest} are chosen such that they yield between 10 and 15 possible \textit{attribute} combinations. For example, six \textit{breed groups} combined with two \textit{grooming-frequency bins} yield 12 combinations. With 100 \textit{entities}, this corresponds to approximately 10 \textit{entities} per group; 
thus, the choice of more granular \textit{attributesOfInterest} would lead to insufficient statistical power.

\subsection{Person unlearning}
For the person unlearning task, we rely on Pantheon \cite{dataset_pantheon}. We select as \textit{attributesOfInterest} the \textit{occupation} and the \textit{HPI}, a continuous variable derived from Wikipedia page access statistics \cite{dataset_pantheon}.

The cardinality of \textit{occupation} is reduced to facilitate dataset balancing by grouping categories such as \textit{musician} and \textit{actor} under a single \textit{artist} value and by removing professions with few individuals. The \textit{HPI} is discretized into four quartile-based groups (\textit{Q0\_25}, \textit{Q25\_50}, \textit{Q50\_75}, \textit{Q75\_100}). Using the resulting 3~$\times$~4 \textit{attribute} grid, the dataset is balanced and 100 \textit{entities} are selected.

\subsection{Dog breed attributes}

For \textit{breeds}, \textit{attributes} are combined from 
\cite{dataset_akc} and
\cite{dataset_pawsomeauthority}. While other authoritative academic sources exist, we did not identify any that provide a sufficiently large number of \textit{breeds} to support balanced selection at the scale required (as recommended by I-CARE, an initial pool of roughly 200 \textit{entities} is needed to select 100 \textit{breeds}).

We select \textit{grooming frequency} (a continuous \textit{attribute} describing how often brushing is required) and \textit{supergroup} (categorical, with values such as \textit{retrievers} and \textit{terriers}) as \textit{attributesOfInterest}. Additionally, more than 40 other \textit{attributes} are available. \textit{Grooming frequency} is discretized into two bins based on quartiles, and \textit{supergroups} with very few \textit{entities} are removed. The resulting 2~$\times$~6 \textit{attribute} grid is balanced, and 100 \textit{breeds} are selected.

\subsection{Scenes}
Dataset balancing is performed automatically by selecting the two \textit{attributes} with the most uniform distributions: \textit{sports} and \textit{natural}. These \textit{attributes} belong to different semantic groups (Functions and Spatial Envelope, respectively) and are binary, indicating whether a scene is suitable for sports activities and whether it corresponds to a natural (non-built) environment.

\section{Metrics} \label{ap:metrics}
\subsection{Similarity between entities.}
We use four \textit{similarity} functions:
\begin{enumerate}
    \item \textbf{Clip}: cosine similarity between the CLIP text embeddings of the names of two \textit{entities}, intended to capture semantic similarity \cite{Radford2021Clip}.
    \item \textbf{Dino}: computes the cosine similarity of DINOv2~\cite{oquab2023dinov2}
ViT-S/14 CLS-token embeddings extracted from the images generated for each \textit{entity}.
Formally, for each \textit{entity} $e$ we generate $4$ images using the original model
and compute the mean CLS-token embedding
$\bar{v}_e = \frac{1}{4}\sum_{k=1}^{4} \phi(x_{e,k}) \in \mathbb{R}^{384}$,
where $\phi$ denotes the DINOv2 encoder.
The similarity between two \textit{entities} is then
$s_\text{dino}(e_1, e_2) = \langle \hat{v}_{e_1}, \hat{v}_{e_2} \rangle$,
where $\hat{v}_e = \bar{v}_e / \|\bar{v}_e\|_2$.
    \item \textbf{Act}: mechanistically grounded metric using the UNet cross-attention activation similarity, by averaging the outputs of all 14 cross-attention layers over spatial positions, denoising steps, and seeds into a single L2-normalized fingerprint, then defining similarity as the cosine between two \textit{entities}' fingerprints. While any UNet could be used, we use the same \textit{Stable Diffusion 1.4} used as generative \textit{model}, which could be considered to break the \textit{SimilarityBetweenEntities} abstraction of depending only on the pair of \textit{entities}.
    \item \textbf{Jacc}: a Jaccard-like similarity computed over the sets of annotated \textit{attributes} associated with each \textit{entity}.
\end{enumerate}

\subsection{Interference per entity pair.}
We use five \textit{InterferencePerEntityPair} metrics. In all cases, comparisons are performed between images generated with the same random seed before and after unlearning and then averaged across all seeds.

\begin{enumerate}
    \item \textbf{$\Delta$Clip}: average (across seeds) CLIP score for image--prompt pairs of $entity_2$ before unlearning, minus the corresponding average after unlearning $entity_1$. This captures changes in semantic alignment with respect to the textual prompt \cite{Radford2021Clip}.
    \item \textbf{$\Delta$Brisque}: average (across seeds) BRISQUE score for images of $entity_2$ before unlearning, minus the corresponding average after unlearning $entity_1$. This captures reference-less structural image quality \cite{mittal2012brisque}.
    \item \textbf{$\Delta$Dino}: how much does the DINOv2~\cite{oquab2023dinov2} representation of $entity_2$ change when $entity_1$ is unlearned. That is, the cosine similarity between the average vectors of ``receiver-images generated by emitter-unlearned model'' and ``receiver-images generated by original model''.
    \item \textbf{RMSE}: root mean squared error between images generated with the same seed, averaged across seeds. This captures pixel-wise changes.
    \item \textbf{SSIM}: structural similarity index measure between images generated with the same seed, averaged across seeds. This captures structural changes in image content \cite{wang2004ssim}.
\end{enumerate}

\subsection{Interference per entity.}
We derive 49 \textit{InterferencePerEntity} metrics using a simple combinatorial aggregation procedure. For each of the five \textit{InterferencePerEntityPair} metrics, results are grouped in three different ways:
\begin{itemize}
    \item \textbf{Emitter}: average impact on all other \textit{entities} when the parametrized \textit{entity} is unlearned.
    \item \textbf{Receiver}: average impact on the parametrized \textit{entity} when each of the other \textit{entities} is unlearned.
    \item \textbf{EmitterMinusReceiver}: difference between emitter and receiver values.
\end{itemize}

This yields 15 ``buckets'' (5 metrics $\times$ 3 groupings), each containing 100 scalar values. We then apply three aggregation functions to each bucket: \textit{WorstInterfered} (interference metric value of the entitiy that suffered the worst interference), \textit{NumberOfInterferedWorseThanTarget} (number of \textit{entities} whose interference metric value was worse than of the target \textit{entity}), and \textit{Average} (mean value of the interference metric). Additionally, for \textit{$\Delta$Clip}-based metrics only, we compute \textit{NumberOfInterferedWorseThanZero}, exploiting the natural boundary between quality degradation and improvement. 

Additionally, we define one further \textit{MetricInterferencePerEntity} that does not
aggregate \textit{MetricInterferencePerEntityPair} values, \textit{EmbeddingSpecificityRatio}, but instead directly compares the
visual representation of the forgotten entity before and after unlearning:
$$m_{e,\text{spec}}(e, u) = \frac{d_\text{self}(e, u)}{d_\text{others}(e, u)}$$
where $d_\text{self}(e, u)$ is the cosine distance between the DINOv2 mean embedding of
\textit{entity} $e$ under the unlearned model and under the original model,
and $d_\text{others}(e, u)$ is the mean such distance computed over all other \textit{entities}
prompted in the same unlearning run.
A ratio $\gg 1$ indicates that the forgotten entity's visual representation shifted
substantially more than the background drift induced by the unlearning procedure,
characteristic of a specific unlearning effect.

This results in a total of $15 \times 3 + 3 + 1 = 49$ \textit{InterferencePerEntity} metrics:
\begin{enumerate}
    \item EmitterWorstInterferedClipDiff
    \item EmitterWorstInterferedBrisqueDiff
    \item EmitterWorstInterferedDinoDiff
    \item EmitterWorstInterferedRMSE
    \item EmitterWorstInterferedSSIM
    \item EmitterNumberOfInterferedWorseThanTargetClipDiff
    \item EmitterNumberOfInterferedWorseThanTargetBrisqueDiff
    \item EmitterNumberOfInterferedWorseThanTargetDinoDiff
    \item EmitterNumberOfInterferedWorseThanTargetRMSE
    \item EmitterNumberOfInterferedWorseThanTargetSSIM
    \item EmitterAverageClipDiff
    \item EmitterAverageBrisqueDiff
    \item EmitterAverageDinoDiff
    \item EmitterAverageRMSE
    \item EmitterAverageSSIM
    \item EmitterNumberOfInterferedWorseThanZeroClipDiff
    \item ReceiverWorstInterferedClipDiff
    \item ReceiverWorstInterferedBrisqueDiff
    \item ReceiverWorstInterferedDinoDiff
    \item ReceiverWorstInterferedRMSE
    \item ReceiverWorstInterferedSSIM
    \item ReceiverNumberOfInterferedWorseThanTargetClipDiff
    \item ReceiverNumberOfInterferedWorseThanTargetBrisqueDiff
    \item ReceiverNumberOfInterferedWorseThanTargetDinoDiff
    \item ReceiverNumberOfInterferedWorseThanTargetRMSE
    \item ReceiverNumberOfInterferedWorseThanTargetSSIM
    \item ReceiverAverageClipDiff
    \item ReceiverAverageBrisqueDiff
    \item ReceiverAverageDinoDiff
    \item ReceiverAverageRMSE
    \item ReceiverAverageSSIM
    \item ReceiverNumberOfInterferedWorseThanZeroClipDiff
    \item EmitterMinusReceiverWorstInterferedClipDiff
    \item EmitterMinusReceiverWorstInterferedBrisqueDiff
    \item EmitterMinusReceiverWorstInterferedDinoDiff
    \item EmitterMinusReceiverWorstInterferedRMSE
    \item EmitterMinusReceiverWorstInterferedSSIM
    \item EmitterMinusReceiverNumberOfInterferedWorseThanTargetClipDiff
    \item EmitterMinusReceiverNumberOfInterferedWorseThanTargetBrisqueDiff
    \item EmitterMinusReceiverNumberOfInterferedWorseThanTargetDinoDiff
    \item EmitterMinusReceiverNumberOfInterferedWorseThanTargetRMSE
    \item EmitterMinusReceiverNumberOfInterferedWorseThanTargetSSIM
    \item EmitterMinusReceiverAverageClipDiff
    \item EmitterMinusReceiverAverageBrisqueDiff
    \item EmitterMinusReceiverAverageDinoDiff
    \item EmitterMinusReceiverAverageRMSE
    \item EmitterMinusReceiverAverageSSIM
    \item EmitterMinusReceiverNumberOfInterferedWorseThanZeroClipDiff
    \item EmbeddingSpecificityRatio
\end{enumerate}




\section{Unlearning methods} \label{ap:unlearning_method}
\subsection{Hyperparameters}
The hyperparameters exposed by each method are as follows:
\begin{itemize}
    \item \textbf{\ac{munba}}: forget set (images), retain set (images), learning rate (float), number of epochs (int) \cite{Wu2024_munba}. It is worth mentioning that the implementation uses the idea of unlearning by subtracting LoRA adapters \cite{zhang2024}.
    \item \textbf{\ac{uce}}: forget concept (str), retain concept (str), erase scale (float), preserve scale (float), lambda (float). We follow the recommended configuration from the original paper, where the guide concepts include both the forget and retain concepts, and the preserve concepts correspond to semantically close retain concepts \cite{Gandikota2023UCE}.
    \item \textbf{\ac{spare}}: forget set (images), retain set (images), forget concept (str), overwrite concept (str), forget--retain balance (float), learning rate (float), number of epochs (int). The overwrite concept is chosen as ``some breed $\rightarrow$ cat'' for dog breeds and ``some person $\rightarrow$ child'' for people. This choice makes the unlearning intentionally more destructive, which facilitates (i) verifying whether unlearning is effective, (ii) manually identifying interference effects, and (iii) amplifying metric variations, making interference easier to observe \cite{kelsch2026fade}.
\end{itemize}

\subsection{Equalization}
 Since \textit{UCE} exposes the fewest hyperparameters, it is used as the reference point. The equalization procedure proceeds as follows:
\begin{enumerate}
    \item Manually select \textit{UCE} hyperparameters by testing multiple configurations on five \textit{entities} from a single \textit{task}.
    \item Run the full interference analysis pipeline for \textit{UCE} (100 \textit{entities}, three \textit{tasks}).
    \item Select \textit{MUNBa} hyperparameters by manual experimentation on five \textit{entities} from one \textit{task}, aiming to match the unlearning quality observed with \textit{UCE}. Higher learning rates and more epochs are expected to increase both interference and runtime, so a trade-off is sought.
    \item Run the full interference analysis pipeline for \textit{MUNBa}.
    \item Select \textit{SPARE} hyperparameters by manual experimentation on five \textit{entities} from one \textit{task}, aiming to match both \textit{UCE} and \textit{MUNBa}. We hypothesize that larger semantic gaps between forget and overwrite concepts increase interference without affecting runtime. Preliminary experiments suggest that \textit{SPARE} induces less interference than \textit{MUNBa}, requiring more aggressive settings.
    \item Run the full interference analysis pipeline for \textit{SPARE}.
\end{enumerate}


\clearpage\chapter{Benchmark selected additional results}
\label{ap:results}

\begin{figure}[H]
  \centering
  \includegraphics[width=0.65\linewidth]{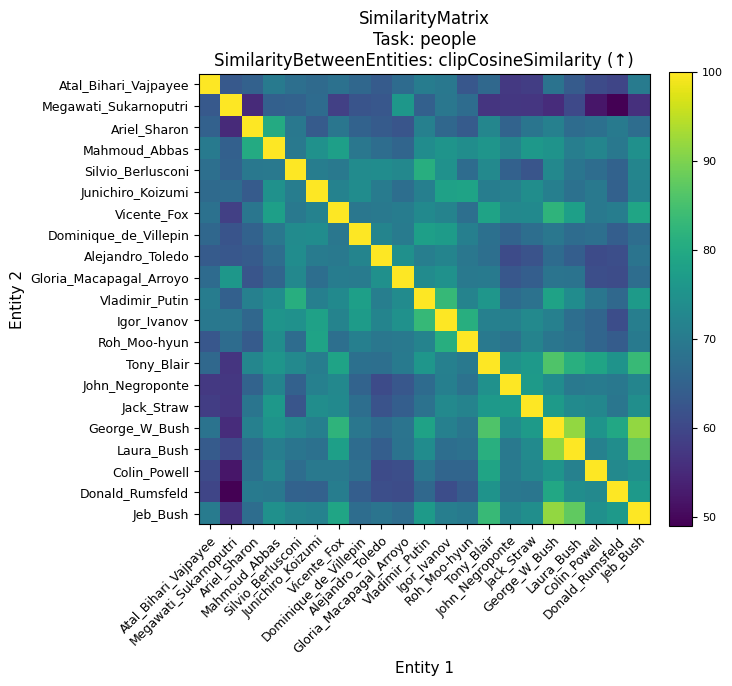}
  \caption{RT \textit{SimilarityMatrix} for the same ranges and sorting as \cref{fig:matrix}. \textit{American politicians} (lower right corner) are noticeably more similar among themselves. As expected, the matrix is symmetric ($s(e_1, e_2)=s(e_2, e_1)$), and the highest similarity is observed for an \textit{entity} with itself ($s(e_1, e_1)=100$). Some identities, such as \textit{Megawati Sukamoputri}, show little similarity to most others; the interferences received and caused by \textit{Megawati Sukamoputri} do not reflect this anomalous pattern.}
  \label{fig:SimilarityMatrix_people_clip}
\end{figure}

\begin{figure}[H]
  \centering
  \includegraphics[width=0.55\linewidth]{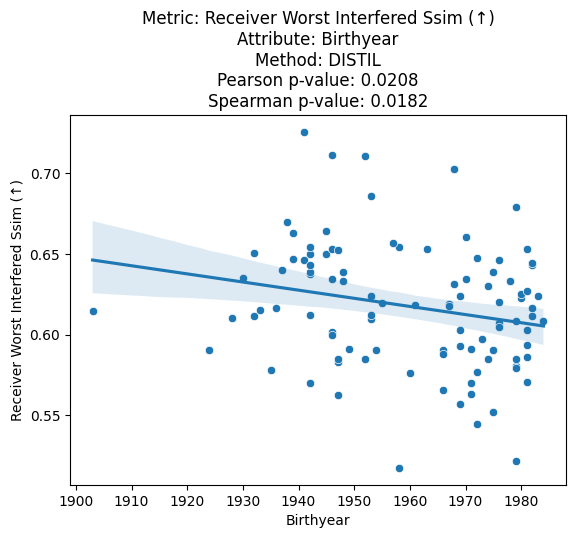}
  \caption{RT \textit{SignificantRelationshipNumerical} analyzing the relationship between the \textit{birthyear} of an \textit{entity} and its $m_e$ \textit{ReceiverWorstInterferedSSIM} (that is, the value of the highest interference that the \textit{entity} received across all 100 unlearning sessions as measured by $m_p=SSIM$). It is important to highlight that \textit{birthyear} is an \textit{attribute} for which the data was not balanced; therefore, this relationship could be due to confounding or other data-related phenomena.}
  \label{fig:SignificantRelationshipNumerical_birthyear_ssim}
\end{figure}

\begin{figure}[H]
  \centering
  \includegraphics[width=0.55\linewidth]{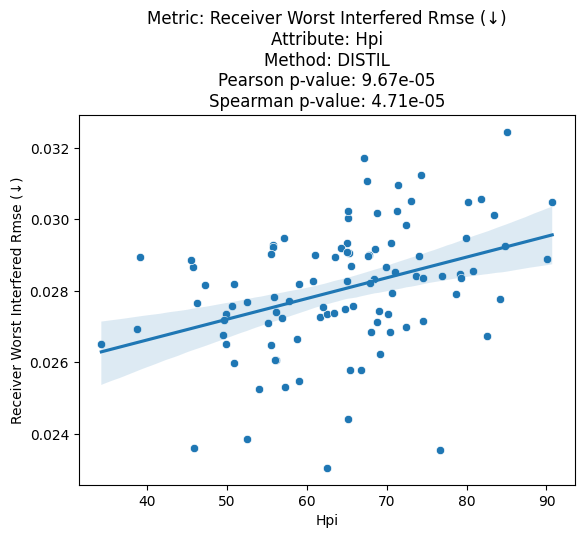}
  \caption{RT \textit{SignificantRelationshipNumerical}. Similar to that observed in \cref{fig:hpi_clip}, HPI (one of the \textit{attributesOfInterest}, based on which the data was balanced) is significantly correlated with the occurrence of interferences. While \cref{fig:hpi_clip} displayed an emitter-grouped $m_e$, here a receiver-grouped $m_e$ is analyzed, demonstrating that more famous \textit{entities} tend to both cause and receive more interferences.}
  \label{fig:SignificantRelationshipNumerical_hpi_rmse}
\end{figure}

\begin{figure}[H]
  \centering
  \includegraphics[width=0.55\linewidth]{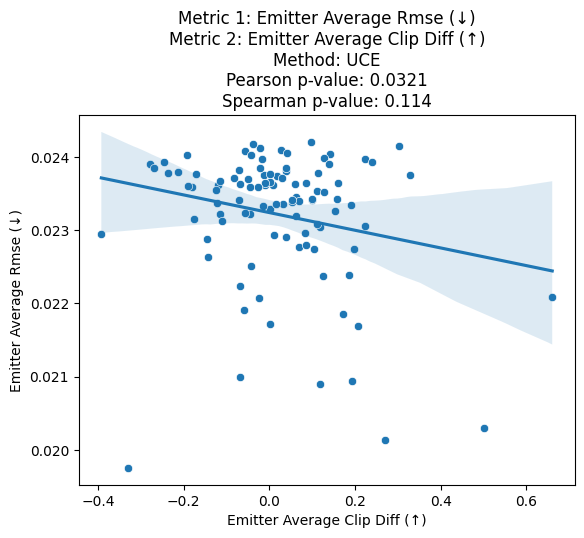}
  \caption{RT \textit{MetricMetricAlignment} displaying the weak relationship between a Clip-based $m_e$ and an RMSE-based $m_e$.}
  \label{fig:MetricMetricAlignment_clip_rmse}
\end{figure}

\begin{figure}[H]
  \centering
  \includegraphics[width=1\linewidth]{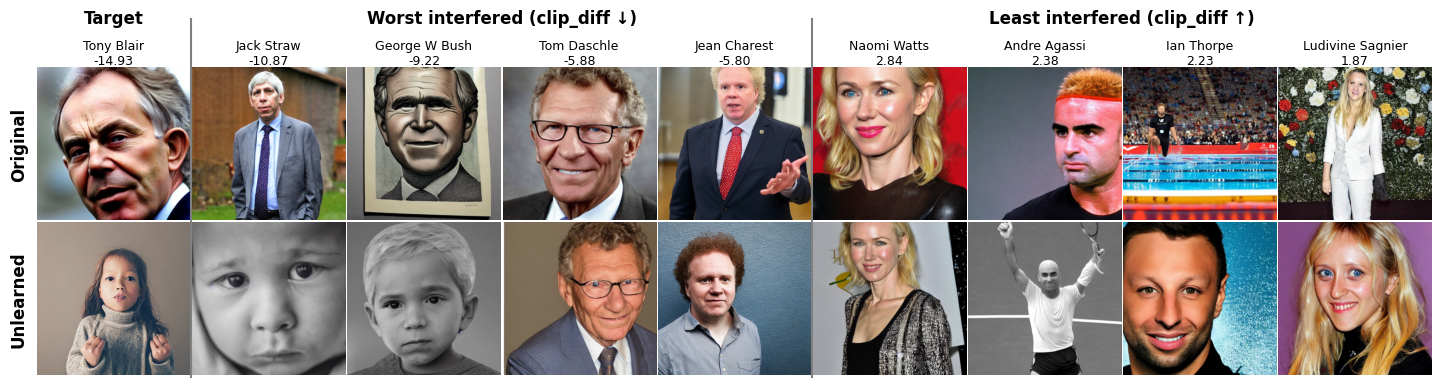}
  \caption{Visual summary of the main effects of overwriting \textit{Tony Blair} by ``kid''. The four \textit{entities} with the worst $\Delta Clip$ are displayed, as well as the four whose generative quality improved the most. We observe that the worst interferences seem to have happened due to meaningful associations between the entities, such as having the same \textit{gender}, \textit{occupation}, and \textit{skin tone}. These three mentioned \textit{attributes} were annotated in the provided data, but only \textit{occupation} was used to balance the dataset. While all images were generated with the same seed, it is observable that the overall layout and image organization changed significantly during unlearning.}
  \label{fig:example_distil_people_blair}
\end{figure}

\begin{figure}[H]
  \centering
  \includegraphics[width=1\linewidth]{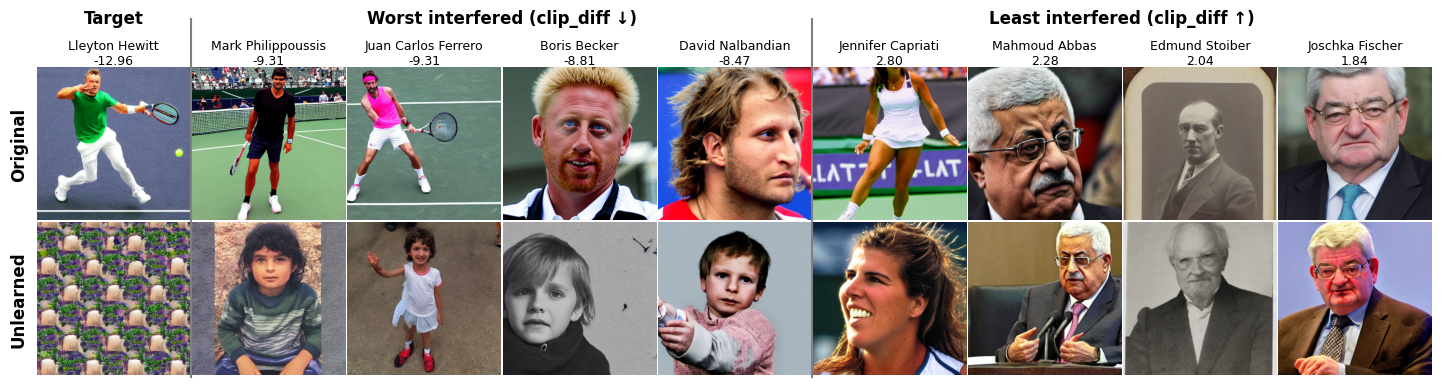}
  \caption{Visual summary of the main effects of overwriting the \textit{athlete} \textit{Lleyton Hewitt} by ``kid'' using \ac{spare}. While the unlearning process was overly destructive to \textit{Lleyton Hewitt} (the generated images are not well-formed), we observe other \textit{athletes} that were generated as children or with child-like features. Curiously, the generation quality for the \textit{female athlete} \textit{Jennifer Capriati} seems to have improved, but this observation was not further analyzed in this work.}
  \label{fig:example_distil_people_hewitt}
\end{figure}

\begin{figure}[H]
  \centering
  \includegraphics[width=1\linewidth]{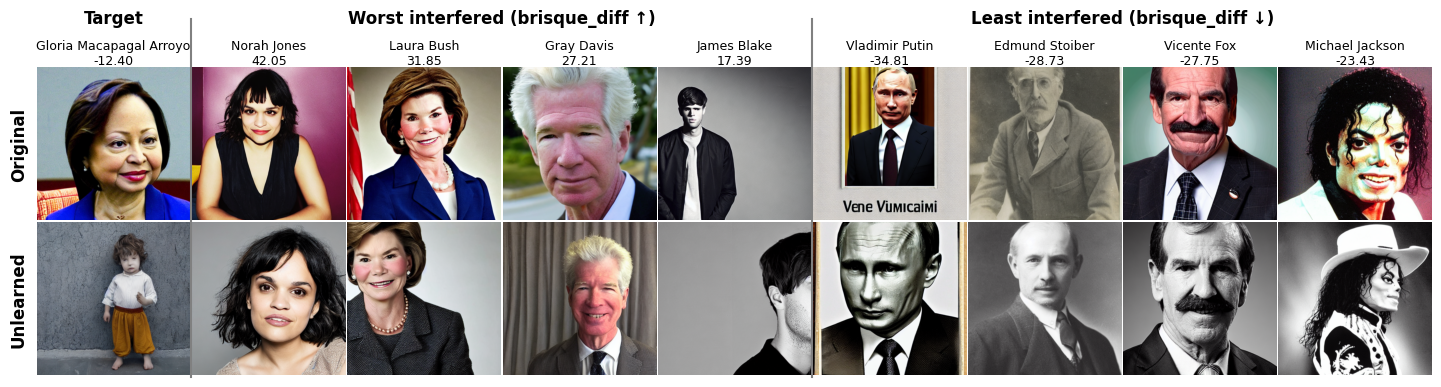}
  \caption{Visual summary of the main effects of overwriting the \textit{politician} \textit{Gloria Macapagal Arroyo} by ``kid''. No significant interference was observed.}
  \label{fig:example_distil_people_gloria}
\end{figure}

\begin{figure}[H]
  \centering
  \includegraphics[width=1\linewidth]{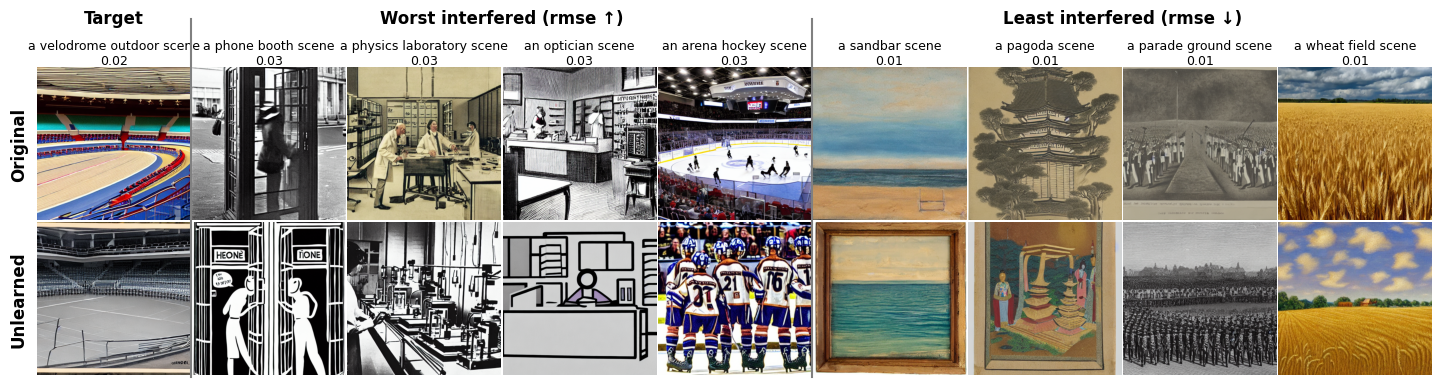}
  \caption{Visual summary of the main effects of unlearning the \textit{scene} \textit{Velodrome Outdoor} using \ac{uce}. We observe that the unlearning process was only partially successful: while the generated scene does not display bike tracks or other velodrome-identifying features, it can still be clearly identified (by the authors; no automated scene classification was performed) as a sports arena. The most-interfered \textit{entities}, as measured by $m_e$ \textit{RMSE}, were generated in a cartoon-like, non-realistic style but preserved a semantic association to their respective \textit{entities}.}
  \label{fig:example_uce_scene_velodrome}
\end{figure}

\begin{figure}[H]
  \centering
  \includegraphics[width=1\linewidth]{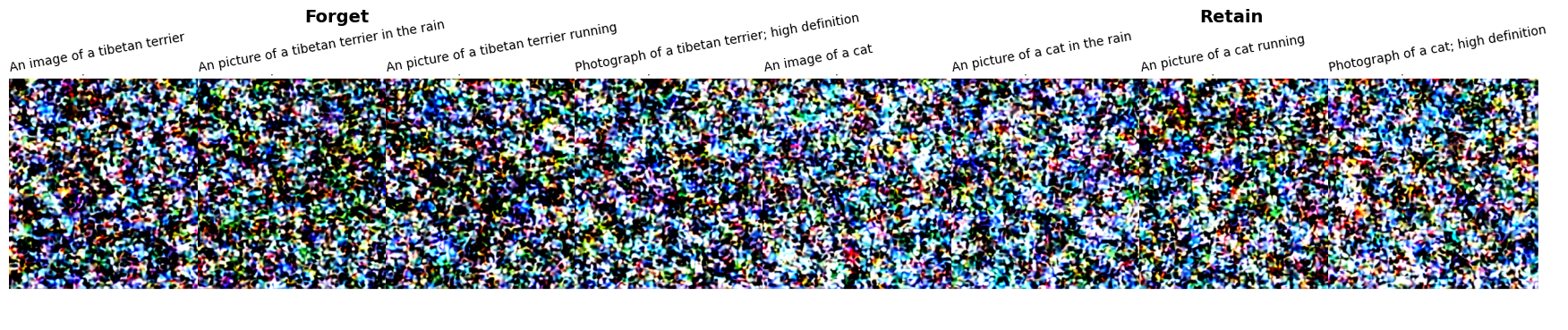}
  \caption{Forget and retain images for the unlearning session of the \textit{breed} \textit{Tibetan Terrier} using \textit{\ac{munba}}. Unfortunately, the \textit{model} collapsed into generating noise, a problem that was observed in a few unlearning sessions, and no apparent cause was identified.}
  \label{fig:example_munba_breeds_fail}
\end{figure}

\begin{figure}[H]
  \centering
  \includegraphics[width=0.55\linewidth]{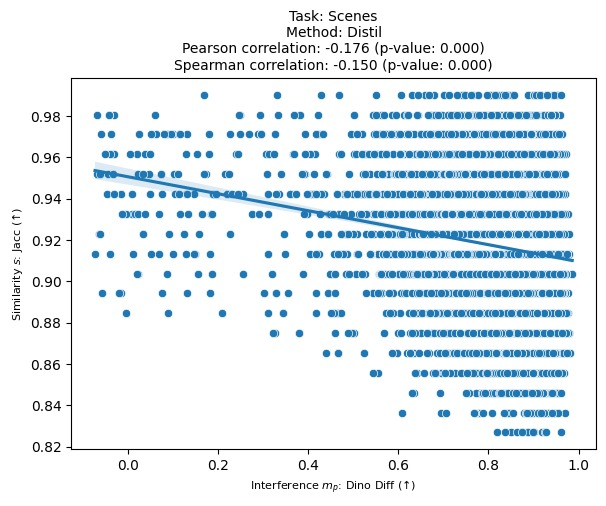}
  \caption{RT \textit{MetricSimilarityAlignment} for \textit{scenes} with \textit{\ac{spare}}, relating the \textit{Jacc} \textit{similarity} to the \textit{$\Delta$Dino} \textit{metricInterferencePerEntityPair}; since \textit{Jacc} is computed over a finite set of annotated \textit{attributes}, it takes discrete values, producing the visible horizontal banding.}
  \label{fig:dinodiff_jacc_scenes_distil}
\end{figure}

\begin{figure}[H]
  \centering
  \includegraphics[width=1\linewidth]{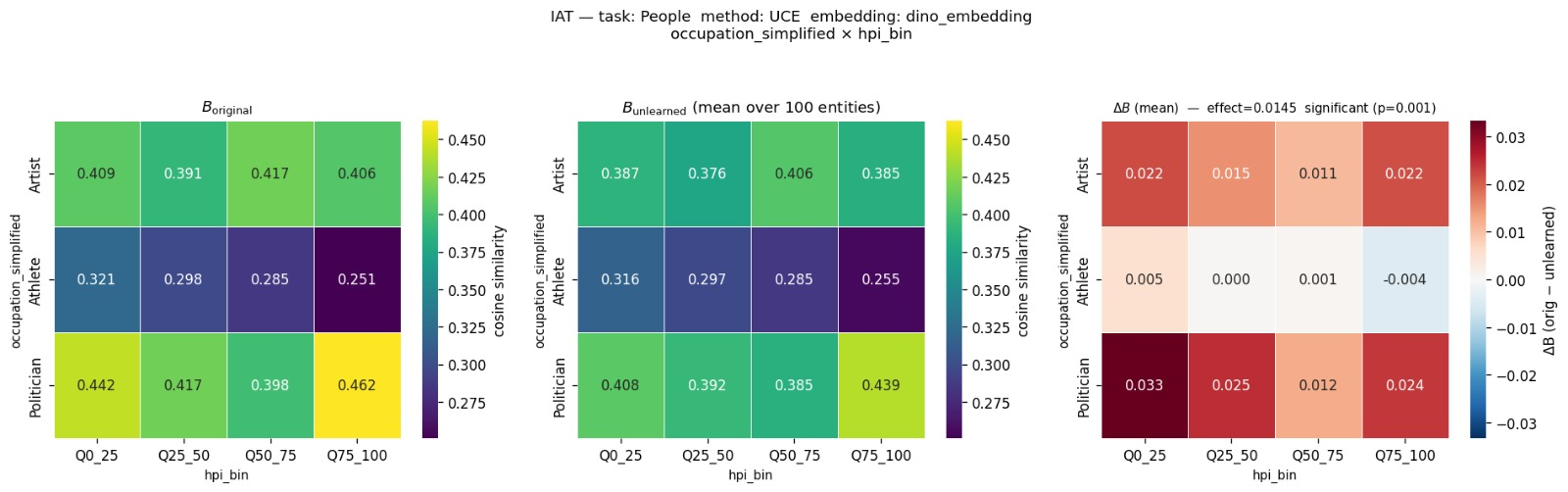}
  \caption{RT \textit{ImplicitAssociationTest} between \textit{occupation} and \textit{HPI} (discretized) for \textit{people} with \ac{uce} (\textit{Dino embeddings}), the \ac{uce} counterpart of \cref{fig:iat_hpi}: associations weaken overall (effect $0.0145$, $p=0.001$), the change being largest for \textit{politicians} and negligible for \textit{athletes}.}
  \label{fig:iat_uce}
\end{figure}

\begin{figure}[H]
  \centering
  \includegraphics[width=0.8\linewidth]{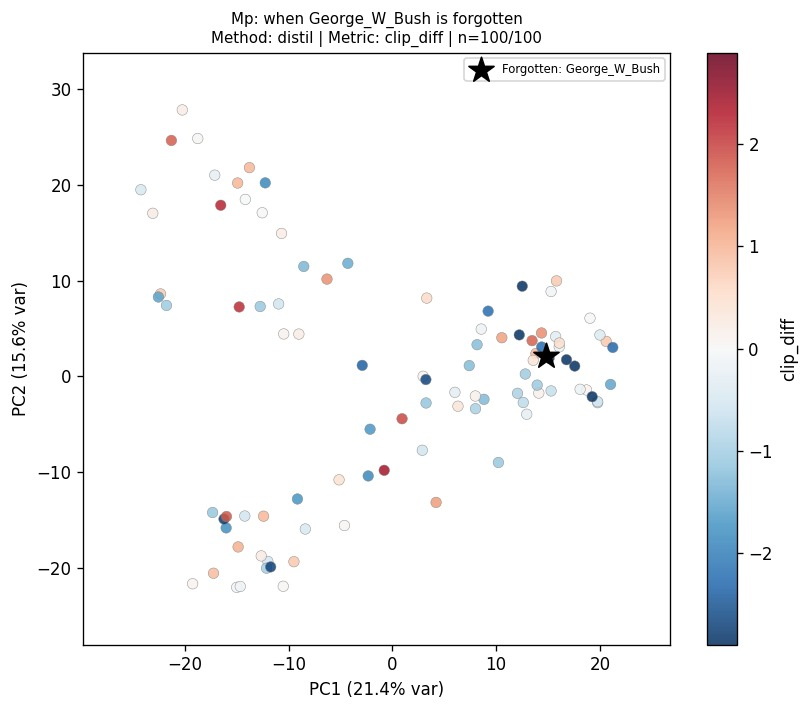}
  \caption{Two-dimensional PCA of the \textit{Dino} \textit{LatentEmbeddings} of the 100 \textit{people} \textit{entities} (the first two principal components capture $21.4\%$ and $15.6\%$ of the variance), with each point colored by the \textit{$\Delta$Clip} interference it receives when \textit{George W. Bush} (star) is unlearned with \ac{spare}. Interference is distributed across the whole embedding space rather than confined to the neighbourhood of the forgotten \textit{entity}, and a few nearby \textit{entities} show positive values (improvement, as described in \cref{sec:results_constructive}); the neighbourhood containing \textit{Bush} is predominantly \textit{politicians}.}
  \label{fig:latent_dino_bush}
\end{figure}

\begin{figure}[H]
  \centering
  \includegraphics[width=0.8\linewidth]{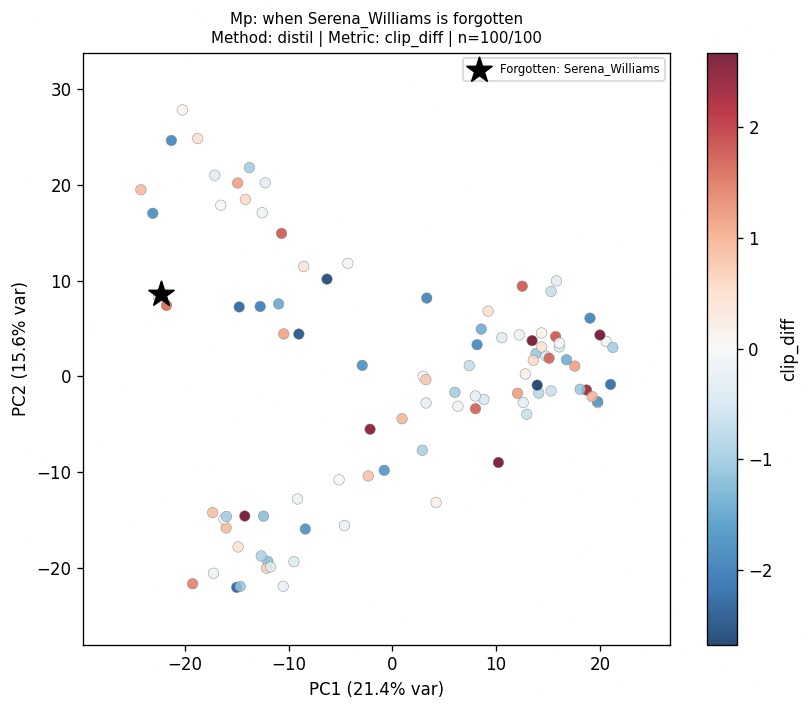}
  \caption{Same PCA projection as \cref{fig:latent_dino_bush}, when \textit{Serena Williams} is unlearned; her group (upper-left) is predominantly \textit{athletes}, and the interference again spreads well beyond her neighbourhood.}
  \label{fig:latent_dino_serena}
\end{figure}

\begin{figure}[H]
  \centering
  \includegraphics[width=0.8\linewidth]{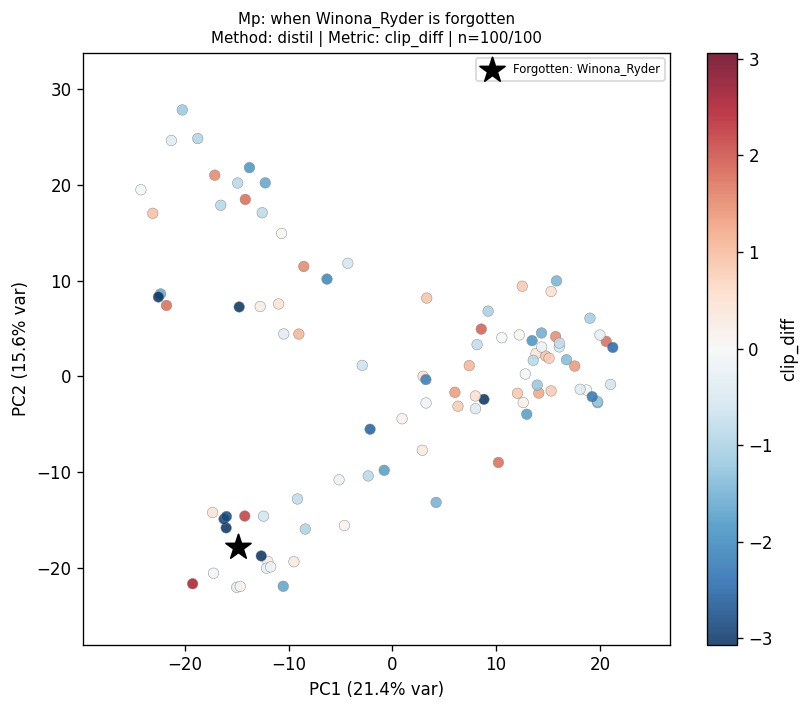}
  \caption{Same PCA projection as \cref{fig:latent_dino_bush}, when \textit{Winona Ryder} is unlearned with \ac{spare}; her neighbourhood (lower-left) is composed predominantly of \textit{artists}.}
  \label{fig:latent_dino_winona}
\end{figure}

\begin{figure}[H]
  \centering
  \includegraphics[width=1\linewidth]{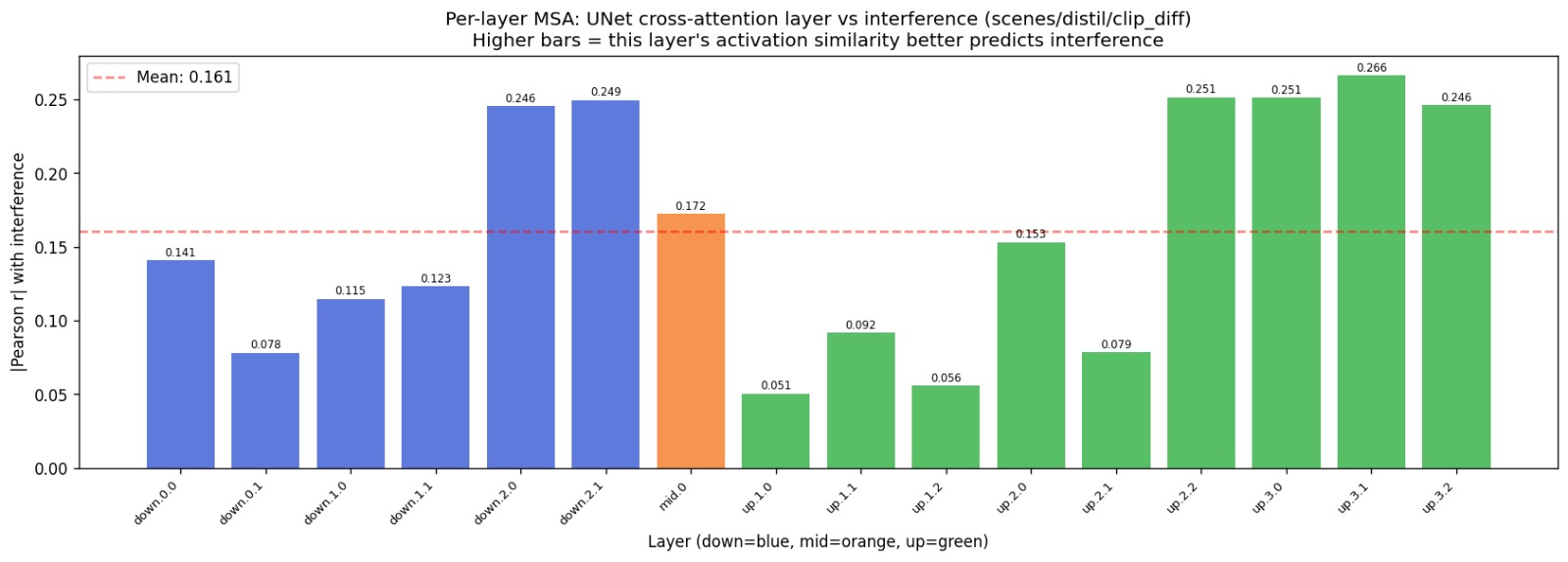}
  \caption{Per-layer \textit{MetricSimilarityAlignment} for \textit{scenes} with \ac{spare}: the absolute Pearson correlation between each UNet cross-attention layer's \textit{activation similarity} and the \textit{$\Delta$Clip} $m_e$. Predictive strength is uneven across the network, highest in the deep down-sampling blocks (\texttt{down.2.x}) and the output-side up-sampling blocks (\texttt{up.3.x}, $\approx 0.25$--$0.27$), and lowest in the shallow up blocks (\texttt{up.1.x}), yet even the best layers remain weakly correlated (mean $0.161$).}
  \label{fig:act_person_per_layer}
\end{figure}

\begin{figure}[H]
  \centering
  \includegraphics[width=0.6\linewidth]{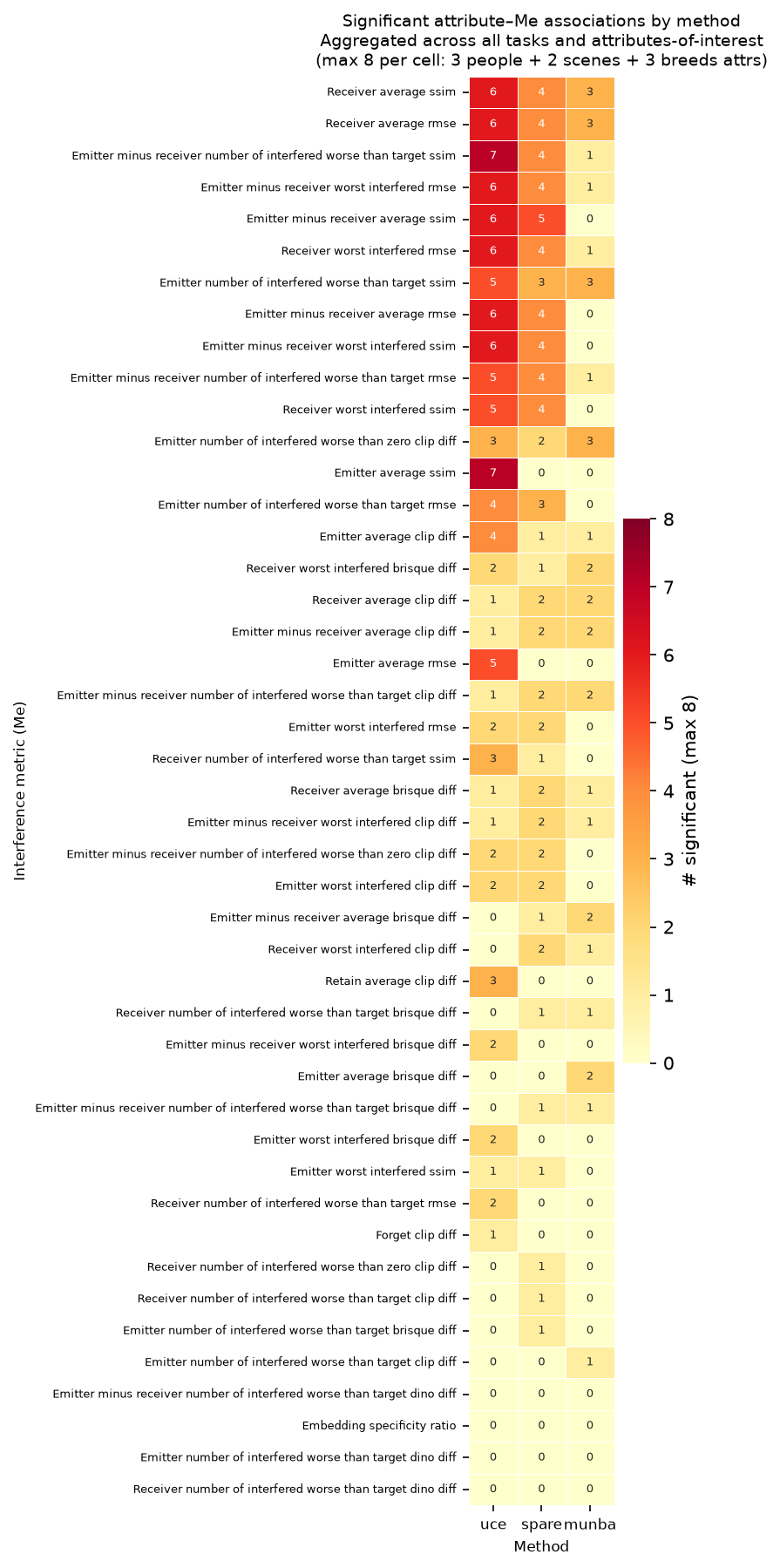}
  \caption{Count of significant relationships found with \textit{SignificantRelationshipCategorical} and \textit{SignificantRelationshipNumerical} grouped by $m_e$. Each cell is aggregated across the 3 \textit{tasks} and all \textit{attributesOfInterest} of each \textit{task}. Studies interested in analyzing relationships between \textit{attributes} and interference could calculate only the most strongly correlated metrics.}
  \label{fig:sig_by_me}
\end{figure}

\begin{figure}[H]
  \centering
  \includegraphics[width=1\linewidth]{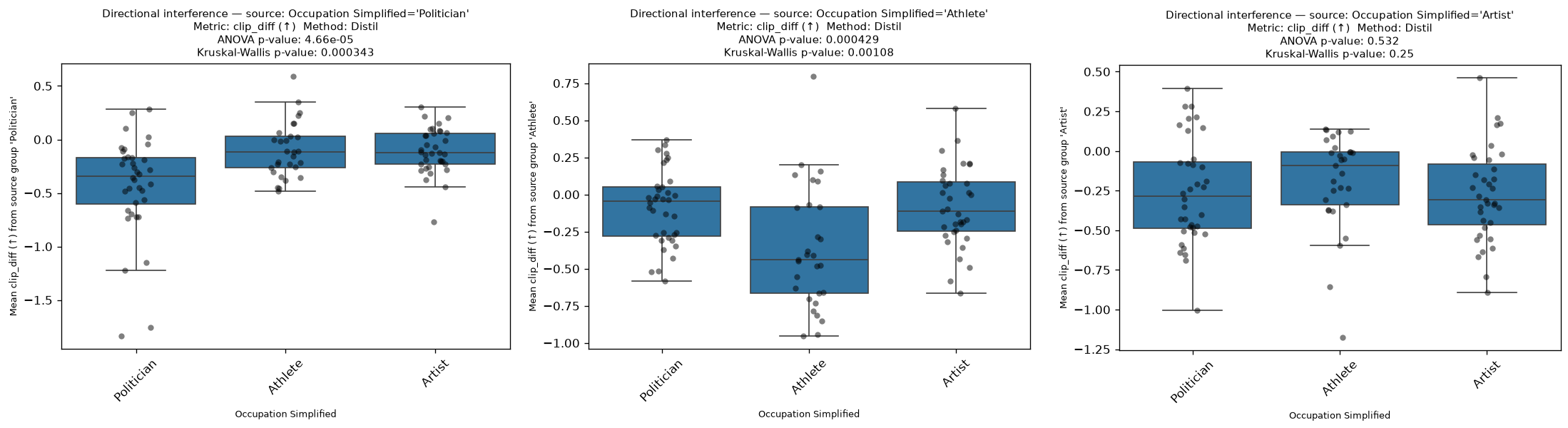}
  \caption{Using the directional variation of \textit{SignificantRelationshipCategorical}, we observe that \textit{politicians} cause more interference to other \textit{politicians}, \textit{athletes} to other \textit{athletes}, but \textit{artists} do not cause significantly more interference to other \textit{artists}.}
  \label{fig:sig_dir_occupation}
\end{figure}

\begin{figure}[H]
  \centering
  \includegraphics[width=0.6\linewidth]{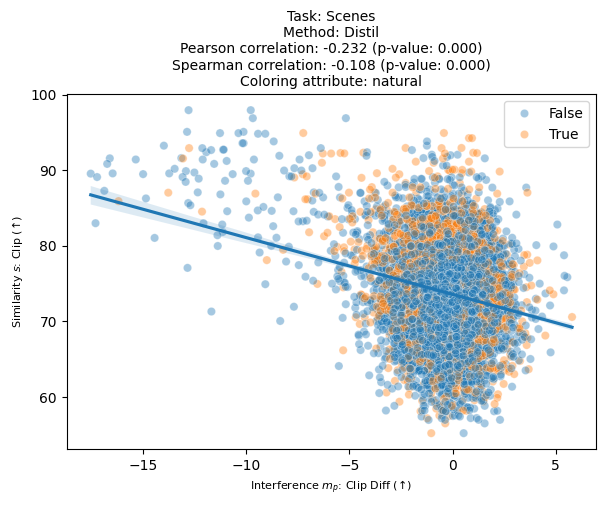}
  \caption{Using the RT \textit{MetricSimilarityAlignment} for 
  \textit{task=scenes} and \textit{method=distil}, with the optional parameter $coloringAttribute$ used for segmenting by unlearned \textit{entities} between natural and artificial scenes. Both groups seem to overlap without distinguishable pattern.}
  \label{fig:msa_coloring}
\end{figure}

\end{appendices}